\documentclass[journal]{IEEEtran}
\usepackage{amsmath}
\ifCLASSOPTIONcompsoc
  \usepackage[caption=false,font=normalsize,labelfont=sf,textfont=sf]{subfig}
\else
  \usepackage[caption=false,font=footnotesize]{subfig}
\fi
\usepackage[utf8]{inputenc}
\usepackage[english]{babel}
\usepackage[dvipsnames]{xcolor}

\usepackage[utf8]{inputenc}
\usepackage[english]{babel}

\usepackage[backend=biber,style=ieee,]{biblatex}
\usepackage{biblatex}
\usepackage{graphicx}
\usepackage{float}
\usepackage{tabto}
\usepackage{csquotes}

\usepackage{soul}
\usepackage[colorinlistoftodos]{todonotes}

\definecolor{navy}{rgb}{0.1, 0.1, 0.8}
\definecolor{gray}{rgb}{0.6, 0.6, 0.6}
\definecolor{myblue}{rgb}{.8, .8, 1}
\definecolor{olive}{rgb}{0.1, 0.5, 0.1}

\usepackage[hidelinks]{hyperref}  
\usepackage{url}
\usepackage[nameinlink,capitalize]{cleveref}
\newcommand{\citep}{\cite}

\usepackage{dirtytalk}

\begin{document}

%
\title{Boosted Genetic Algorithm using Machine Learning for traffic control optimization}
%
%
%

\author{Tuo~Mao,~
        Adriana-Simona~Mih\u {a}it\u {a},~\IEEEmembership{Senior Member,~IEEE,}
        ~Fang~Chen,
        and~Hai~L.~Vu
        
}

%
%

\markboth{IEEE transactions on Intelligent Transport Systems}%
{Shell \MakeLowercase{\textit{et al.}}: IEEE transactions on Intelligent Transport Systems}
%



\maketitle 

\begin{abstract}
Traffic control optimization is a challenging task for various traffic centers around the world and the majority of existing approaches focus only on developing adaptive methods under normal (recurrent) traffic conditions. Optimizing the control plans when severe incidents occur still remains an open problem, especially when a high number of lanes or entire intersections are affected.  

This paper aims at tackling this problem and presents a novel methodology for optimizing the traffic signal timings in signalized urban intersections, under non-recurrent traffic incidents. With the purpose of producing fast and reliable decisions, we combine the fast running Machine Learning (ML) algorithms and the reliable Genetic Algorithms (GA) into a single optimization framework. As a benchmark, we first start with deploying a typical GA algorithm by considering the phase duration as the decision variable and the objective function to minimize the total travel time in the network. We fine tune the GA for crossover, mutation, fitness calculation and obtain the optimal parameters. Secondly, we train various machine learning regression models to predict the total travel time of the studied traffic network, and select the best performing regressor which we further hyper-tune to find the optimal training parameters. Lastly, we propose a new algorithm BGA-ML combining the GA algorithm and the extreme-gradient decision-tree, which is the best performing regressor, together in a single optimization framework. Comparison and results show that the new BGA-ML is much faster than the original GA algorithm and can be successfully applied under non-recurrent incident conditions.
\end{abstract}


%
\IEEEpeerreviewmaketitle

\section{Introduction}
%
%
%
%


 

\IEEEPARstart{T}{raffic} incident management plays an important role for all transportation agencies due to its impact on safety and traffic control operations. To deal with random incidents, various traffic management centres (TMCs) develop policies and response plan strategies in order to minimize the clearance time. Traffic information and control systems are key components in securing an instant response time since they are centralized and can easily alert the incident to TMCs. The typical response plan applied by many TMCs in case of an emergency or an accident is to activate a range of variable message signs, close lanes and force turnings, without having an adaptive control method for signal groups in the affected intersection(s). Most of the time this is a manual process which requires waiting for the incident to be cleared-off until the adaptive control plans are re-activated again. \par

Traffic congestion is generally classified into two types: recurrent congestion (RC) which can appear due to repetitive daily travel patterns and non-recurrent congestion (NRC) which can be caused by unexpected events such as accidents, breakdowns, etc. \cite{Anbaroglu2014,Skabardonis2003,Varaiya2007}. The most problematic incidents can occur at random locations in the city, at various moments in time and do not ever repeat themselves \cite{Anbaroglu2014}. It is a big challenge to model and handle the network optimization under these non-recurrent incidents because of its random occurrence in both time and space. To the best of our knowledge, there are not many research studies which focus on traffic signal control optimization under severe incident conditions due to the high variability of traffic conditions and incident incertitude. \par

This research aims to address this problem and focuses on modelling a new traffic management solution to ease the impact of non-recurrent traffic incidents, by making use of both Genetic Algorithms (GAs) and Machine Learning (ML) models, known for their fast convergence and high accuracy compared to traditional methods. In this paper, we extend the work presented in \citep{Tuo2019} and propose a new double layer algorithm labelled BGA-ML (boosted genetic algorithm using machine learning), which we apply as a a tool for a fast traffic incident response and optimization of the traffic signal control plan during incidents. The method is applied for a case study traffic network and various scenarios are proposed and compared to showcase the benefit of our approach.

Overall, the main contributions of this paper are the following:\par
1.	we propose a new traffic signal control optimization method making use of the integrated power of GA and ML models with the purpose of minimizing the total travel time in urban networks affected by incidents; the approach considers the traffic in all the surrounding area beyond a single individual intersection affected by the disruption and is integrated with traffic simulation to obtain the traffic outcomes;\par
2.  we use ML to replace the traffic simulation modelling and predict directly the total travel time using all previous simulated historical traffic records; findings reveal that Extreme-Gradient Decision-Tree (XGBT) are outperforming other regressors such as Gradient Boosting Decision Tree (GBDT), Random Forest (RF) and Linear Regression (LR); \par
3.  within the optimization framework, we consider the capacity drop caused by the traffic incident and the driver's route diversion;\par 
4.	we showcase the dramatic travel time reduction before and after deploying a regular GA (which we initially proposed in our work published in \citep{Tuo2019}) for signal optimization of an incident affected road network; \par
5.  We then observe low computational time for the new proposed BGA-ML which focuses on a new integrated approach using Machine Learning for speeding the optimisation process. This is mainly due to the fact that we replace the simulation model with the ML model when estimating the fitness value for each new traffic plan that the GA generates. Herein the idea is to learn from all  previous simulation scenarios/runs and choose best traffic signal plan without re-running multiple simulation runs. 

This paper is organised as follows: Section II presents the literature review focusing on existing methods which have applied GAs and ML modelling approaches for traffic signal control so far; Section III introduces the methodology of the paper by presenting the definition, optimization process and the baseline GA modelling followed by the new proposed boosted BGA-ML method; Section IV discusses the case study, the network and optimization construction, hyper-parameter tuning for the BGA-ML approach, followed by the presentation of results through various Scenarios in Section V.

\section{Literature review}

\subsection{Traffic signal control modeling using GA}

Current traffic signal control models are refined to deal with mostly recurrent congestion in the network (daily repetitive travel profiles), but they are not optimized or tuned to the congestion caused by non-recurrent traffic incidents. Severe traffic incidents may strongly influence the overall network performance and should not be neglected. A well-concluded review published in \cite{Papageorgiou2003} presented the traffic control modeling for both arterial roads and motorways. In this review, a ``store-and-forward model” is introduced to simplify the model-based optimization method by enabling the mathematical description of the traffic flow process without discrete variables; as well it uses the Traffic-response Urban Control (TUC)  strategy for calculating the real-time network splits \cite{Diakaki2003}. Ritchie \cite{Ritchie1990} introduced multiple real-time knowledge-based expert systems (KBES) to the advanced traffic management (ATM) system in order to provide suggestions to the control room staff when non-recurrent congestion happened. At that time, the cooperation of artificial intelligence (AI) and ATM were very pioneering and the combination of AI and ATM became a good direction for later research. This conceptual design can be fulfilled now by recent machine learning techniques and more advanced big-data processing. \par
Among various models, the GA is a popular method for optimizing traffic signal controls which was first introduced by Goldberg and Holland \cite{Goldberg1988} in 1988, and later applied to traffic signal timing optimization in 1992 \cite{Foy1992}. In 2004, Ceylan and Bell \cite{Ceylan2004} applied stochastic user equilibrium to model the driver’s route choice under different signal timings while using GAs to optimize the traffic signal timing. It was also concluded that GAs are simpler and more efficient than previous heuristic algorithms. GAs have been successfully used as well for a multi-objective control plan optimizations for choosing the most effective traffic control plan in \cite{Mihaita2018}. Recently, due to an increase in computational power availability, GAs and traffic simulation have started to be combined together in order to optimize the offset, green splits, and cycle time of all intersections in a network \cite{guo2019model}. There is however a gap in terms of delay time needed to finalise the optimization in critical operational times and meet all the needed criteria of traffic centres. \par
   
Over all, most applications are offline, they take a long time to achieve the optimum traffic signal control and there is still a gap in researching the more efficient and fast response in traffic signal control modeling in order to deal with non-recurrent traffic incidents. This is the motivation behind our approach and methodology which try to address these problems by combining GAs and more innovative methods such as machine learning models in order to make use of both the reliability of GA and the fast prediction time of ML. \par

\subsection{Traffic signal control modeling using Machine Learning}
ML modelling, especially reinforced learning (RL) and Q-learning, is normally used for adjusting the real-time adaptive control agents by considering the current state of the network (or sub-network) and by trying different actions with rewards associated to them \cite{sutton1988learning}. 

Since it is impossible to attempt all the actions in the real world, simulation models are used to trial different actions. In the early years, simulation models were fairly simple since the available PC computation power was not very powerful \cite{abdulhai2003reinforcement,thorpe1996tra,wiering2000multi}, such as the cellular automation model \cite{brockfeld2001optimizing}. Later in the years of 2000, traffic simulation software became more complex/realistic and provided APIs for secondary development; therefore most research studies utilized traffic simulators as the base of training ML models and started to be more focused on the structure of the ML framework, including the state space, the action space and the rewards definition. For state space, most researchers use the number of queued vehicles \cite{arel2010reinforcement,balaji2010urban} which are all from the upstream link of an intersection. This set up will ignore the downstream traffic congestion caused by the traffic incidents. The action space is normally defined as all the possible phases for each signal \cite{arel2010reinforcement,el2013multiagent,abdoos2013holonic,balaji2010urban,chin2011q}. The reward definition is normally defined as the delay time (\cite{arel2010reinforcement,el2013multiagent}) and the queue length ( \cite{abdoos2013holonic,chin2011q,balaji2010urban}). Later in 2014 and 2016, two reviews of the traditional reinforcement learning for traffic control research were constructed \cite{el2014design,mannion2016experimental}. In 2015, deep reinforced learning was firstly introduced to traffic signal control optimization in \cite{mnih2015human} and further refined in 2016 by Van der Pol et al. \cite{van2016deep}, while considering the coordination of multiple intersections in a small network. In 2017, a traffic signal control policy has been trained by deep policy gradient and applied to a large traffic network by assuming multiple intersections could be controlled with the same agent \cite{casas2017deep,mousavi2017traffic}. The result showed promising potential for policy-based reinforcement learning for traffic signal control. \par

To summarise, previous traffic signal control using ML barely discussed the capability of solving the sudden capacity drop problem caused by traffic incidents. In all cases, ML models are used for making real-time decisions which may be hard to judge in terms of their reliability and applicability. In this paper, ML modelling is used for performance (fitness value) predicting instead of making decision directly. We use various ML models to process the time series data of the traffic status under all known traffic conditions including the changes in the traffic signal and the capacity drop at the time of the reported accident, and predict the network performance in the near future.

\section{Methodology}
\subsection{Problem formulation}
There are four different steps for creating a traffic incident response: incident identification, verification, response, and clearance. This paper is basically focused on the modeling of traffic management and control after an incident has been confirmed and reported by TMC. The proposed model is applied in the response and clearance phases. To simplify the case study, we assume that the incident was previously detected, verified and the duration of the incident clearance was predicted. In addition, the severity of the incident is also reported as an indication of the number of lanes affected. \par
Last but not least, the incident affected area is determined using previous studies. Recently, Pan et al. \cite{Pan2014} studied the spatial-temporal impact of traffic incidents based on archived data using advanced sensors and came up with the incident impacted area and the delay occurrence prediction in a road network. The affected area normally contains all the surrounding network which experiences the congestion caused by the incident and it is generally time-dependent to the reported location of the incident. The problem we are trying to solve is how to optimize the traffic control plan around the incident location, in order to minimize the impact of the incident in terms of vehicle total travel time. Therefore, we use the road network in the affected area which is pre-determined, and we formulate the problem as following: \par
Given a road network which has been identified as affected by an accident, we define the following:\par
$A$ \tabto{2cm} is the set of links in the network, \par
$W$ \tabto{2cm}	is the set of origin-destination pairs of the network,\par
$R_w$ \tabto{2cm} is the set of routes between origin-destination pair $w \in W$,\par
$d_a$ \tabto{2cm} is the queuing delay at link $a\in A$,\par
$f_r^w$ \tabto{2cm} is the flow on route $r\in R_w$,\par
$v_a$  \tabto{2cm} is the link flow on link $a\in A$,\par
$\lambda_a$ \tabto{2cm} is the ``link green split” $\lambda_a$ which is determined by traffic signals at the end of the link (the definition will be discussed in the next section),\par
$t_a (v_a,\lambda_a )$ \tabto{2cm} is the travel time on link $a \in A$ described as a function of link flow $v_a$ and ``link green split” $\lambda_a$,\par
$S_a$ \tabto{2cm}	is the capacity of link $a \in A$,\par
$\sigma_{ar}^w$ \tabto{2cm} is 1 if route $r$ between O-D pair w uses link $a$, and 0 otherwise, \par
$D_w$ \tabto{2cm}	is the demand between O-D pair $w\in W$,\par
The objective is to minimize the total travel time of the network. The target objective function is as follow:\par
\begin{equation}\label{eq1}
minimize\sum_{a \in A}\int_{0}^{v_a} t_a (v_a,\lambda _a )dx
\end{equation}\par
Subject to\par
\begin{equation}\label{eq2}
\sum_{w \in W} \sum_{r \in R_w} f_r^w  \sigma _{ar}^w=v_a,a \in A
\end{equation}
\begin{equation}\label{eq3}
\sum_{r \in R_w} f_r^w  =D_w,w \in W
\end{equation}
\begin{equation}\label{eq4}
v_a \leq \lambda_a S_a,a \in A
\end{equation}
\begin{equation}\label{eq5}
f_r^w\geq 0,r \in R_w,w\in W
\end{equation}
Equation \ref{eq2} represents the relation between route flows ($f_r^w$) and link flows ($v_a$). Equation \ref{eq3} shows the flow conservation between route flows and O-D demands. Equation \ref{eq4} shows that link flow is limited by the exit capacity, which depends on the link capacity and link green split. Equation \ref{eq5} indicated that link flows must be no less than zero.
\subsection{The definition of link green split $\lambda_a$}
In this paper, the definition of ``link green split” ($\lambda_a$) is the same as the one in the study of  Yang and Yagar \cite{Yang1995}, which is the amount of green time granted for a link (link $a$) in a signalized intersection. As for Smith and Van Vuren \cite{Smith1993}, green time is divided into: phase green time and link green time. A phase is defined as a maximal set of compatible approaches in an intersection. Therefore, the phase green time is the green time of certain phase in a cycle in a signalized intersection. The link green time is the green time granted for a link by all the corresponding phases in a cycle of a signalized intersection.

Let $\Lambda_jk$ be the proportion of green time for which the $k^{th}$ phase at junction j, therefore we can call $\Lambda_jk$ a ``phase green split”. The allocation of green time to all phases at a junction determines the green time of each link entering that junction, therefore for each link $a$, the ``link green split” ($\lambda_a$) is the summation of all those phase green splits ($\Lambda_jk$) for which phase $k$ at junction $j$ contain the movement of link $a$, or:
\begin{equation}\label{eq6}
\lambda_a=\sum_{phases\ S_{jk} \ contain\ link\ a} \Lambda_{jk}
\end{equation}
\par To be clear, for each junction j, the sum (over k) of ``phase green split”  $\Lambda_{jk}$ will be 1:

\begin{equation}\label{eq7}
\sum_k \Lambda_{jk} =1. 
\end{equation}

\subsection{Assumptions}
In this paper, we assume that the O-D demands are predefined and fixed for the duration of our analysis. We use traffic assignment model to get the link traffic flows which depend on link cost functions and O-D demands. Therefore, we can get deterministic link flows. 

In Equation \ref{eq7}, we assume that there is no cycle loss time in each cycle of an intersection. In addition, we assume that the amber (yellow) time for each phase is considered as the green time. In conclusion, the $\lambda_a$ in this paper is the ``link green split” other than the ``phase green split”. In addition, the link travel time function (or cost function) is fixed for all links in the investigated road network which only depends on the link flow and the ``link green split”. Therefore, the only parameter we try to optimize for each link is the ``link green split” $\lambda _a$.\par
For traffic signals in the network, we assume that each phase of a cycle grants green to fixed movements. The cycle length and order of phases in a cycle are fixed. Only the duration of each phase is tunable. The duration of all phases in all signalized intersections are actually the decision variables for the optimization problem. 

In our case study network, all the roads have two lanes and we simulate an incident affecting one of the two lanes at one location. We assume that all similar accidents have the same impact on any two-lane road sections in our network.

\subsection{Optimization process}
The introduction of ``link green split” to our problem leads to an optimization problem for traffic signal timing because of the direct relationship between ``link green split” and ``phase green split” in Equations \ref{eq6} and \ref{eq7}. Now the optimization problem can be transformed into the optimization of the traffic signal timing in a road network.

\subsubsection{Data input}
The specification of the network is required as an input, which consists of:\par
•	O-D configuration: contains the location of origins and destinations,\par
•	O-D demand table: contains the trips between each pair of origin and destination,\par
•	Network configuration: contains all information about links, nodes, speed limits, road capacity, etc. \par
•	Link detail table: contains link free-flow travel time, link speed limit, link capacity, and number of lanes,\par
•	Traffic signal configuration: signalized node indexes, number of phases, cycle time, signal timings, phase green splits, and the links granted green for each phase.\par
\subsubsection{Optimization steps}
\par
We solve the optimization process by following the steps:\par
(1) Import the O-D configuration, O-D demand, network configuration, link detail table, and traffic signal configuration into the traffic simulation model;\par
(2)	Generate all possible fixed traffic signal plans for all nodes in the network. In this paper, the only variable in each traffic signal plan is the phase duration, which means the sequence of the phases and the cycle length are fixed. For example, a network contains $n$ signalized intersections. One intersection (intersection $\#i$) has 4 phases, then we use the phase duration ($ [p_{i1},p_{i2},p_{i3},p_{i4}]$) to represent this intersection and the network traffic signal plan is noted as $[ [p_{11},p_{12},p_{13},p_{14}],[p_{21},p_{22},p_{23},p_{24}],   ...  ,[p_{n1},p_{n2},p_{n3},p_{n4}]]$. The full description of how to generate the fixed traffic signal plans are provided in the ``initialization" module in the section \ref{GA_modelling} (entitled GA modelling);\par
(3)	For each traffic signal control plan, we run the Aimsun simulation model of the network to get the total travel time which is defined in Equation \ref{eq1}; more detailed information of constructing the traffic simulation will be described in the \cref{case_study};\par
(4)	Check all the total travel times for all traffic signal control plans and get the minimal total travel time and the corresponding optimal traffic signal control plan;\par
(5)	 Output the optimal traffic signal control plan.\par

\subsection{GA modelling}
\label{GA_modelling}
In our study, we employ a standard GA algorithm \cite{GA_python} for traffic signal control optimization which we adapt to our network needs and reported traffic incident.
In the following, we detail the parameters and steps we have followed to successfully deploy such model for traffic control plan optimization. \par
•	\textbf{Fitness function}: To adapt our problem to GA, the target function in Equation \ref{eq1} is utilized as the fitness function. As we want to minimize Equation (1) we employ the reverse of Equation \ref{eq1} as our fitness to maximize the fitness value in GA. Then the fitness value is shown in Equation \ref{eq8}.
\begin{equation}\label{eq8}
Fitness=- \sum_{a \in A}\int _0^{v_a}{t_a (v_a,\lambda_a )dx}
\end{equation}
\par
•   \textbf{The decision variable}: The decision variable is a vector of all phase durations for all the signalized intersections within the network. In order to optimize the target function (Equation \ref{eq1}), we need to code the decision variables as the chromosome in GA. The coding process is illustrated as following:\par

Decision variables $\Psi$ (array of arrays) =\\
$$[ [p_{11},p_{12},p_{13},p_{14}],[p_{21},p_{22},p_{23},p_{24}],   ...  ,[p_{n1},p_{n2},p_{n3},p_{n4}]]$$

\par
Chromosome $\psi$ (array) =\\
$$[    p_{11},p_{12},p_{13},p_{14},      p_{21},p_{22},p_{23},p_{24}, ...  ,p_{n1},p_{n2},p_{n3},p_{n4}]$$

Where $p_{uv}$ means the phase duration of intersection $u$ phase $v$ and \textit{n} is the total number of signalized intersections. Observe that the chromosome in GA is the same as the decision variable with less groupings.\par
\begin{figure*}[!t]
\centering
\includegraphics[width=5in]{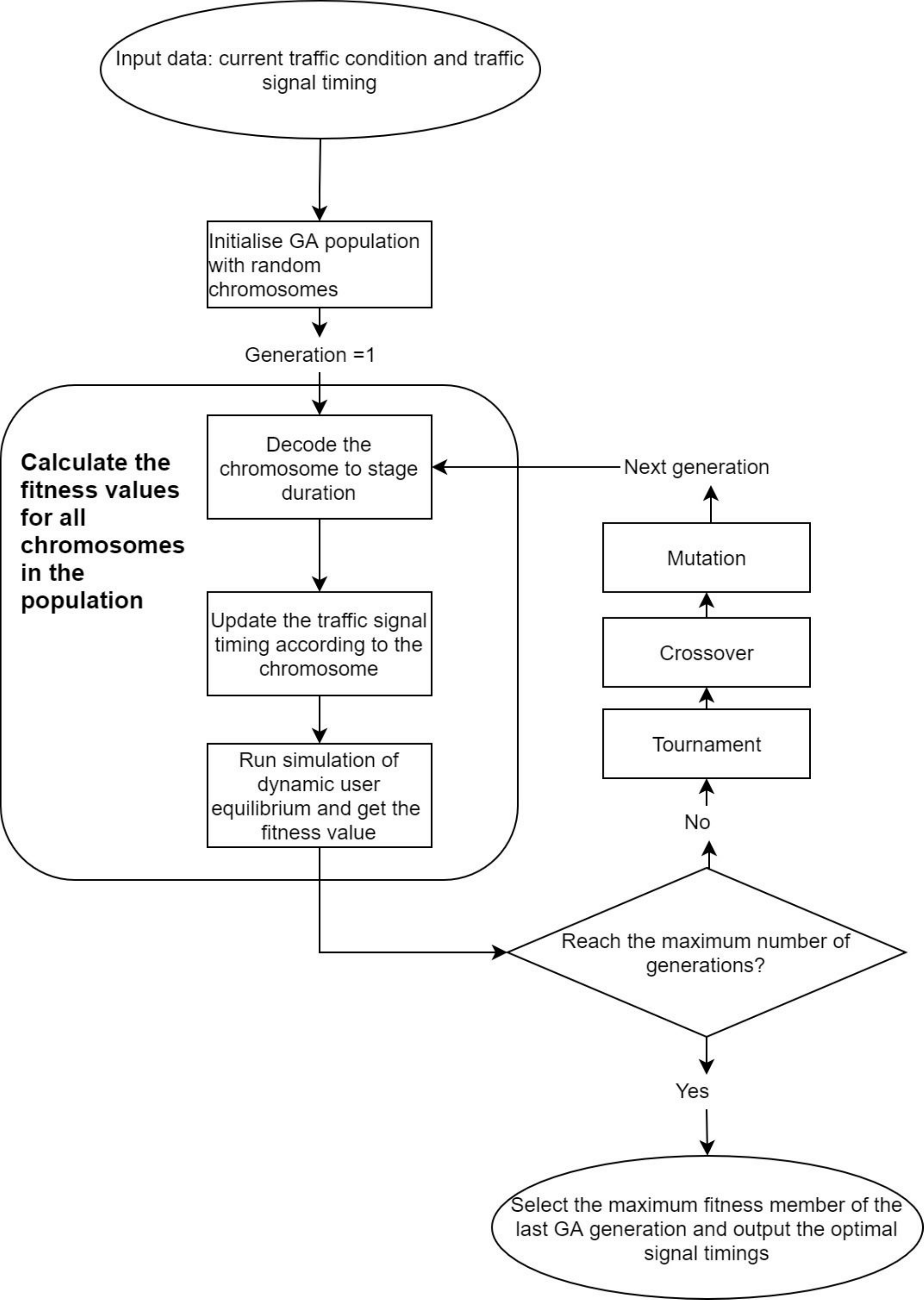}

\caption{GA optimization process.}
\label{fig_GA_flow}
\end{figure*}

•	\textbf{The Genetic Algorithm structure for traffic signal optimization}: is shown in Figure \ref{fig_GA_flow} and contains various modules such as ``check stop”, ``tournament”, ``crossover” and ``mutation” which are also adapted to our application. 
A detailed description of these modules is given in the following:\par
1. \textbf{Prepare input data}: Within GA there are several parameters that need to be determined in order to get a fast convergence and a short computation time. We first use the current traffic condition and traffic signal timing but also fix: the population size, maximum number of generations, probability of crossover, and probability of mutation. \par
2. \textbf{Initialization}: initialize the GA population with random chromosomes of the dataset. As we can see, it is very computational intensive to sample all possible traffic signal control plans with all possible combinations of phases spreading across high phase intervals. Let’s consider, for example, one signalized intersection which has 4 phases. Each phase has a duration ranging between minimum 3 and maximum 90 seconds, which must be an integer. This means a total of $(90-3+1)^4=59,969,536$ possible traffic control plans. The computational times to test all of the phase combinations to find the optimal solution can be quite intensive just for one intersection, not to mention more complicated road networks with various nodes and complicated connections. 
 \par
Therefore, we randomly and uniformly sample the number of individuals in each generation from the total feasibility space of phase combinations as follows: we fix the cycle length to 90 seconds, the number of phases in each signal to 4 and we also establish the sequence of the phases in each traffic signal plan. We also fix the range of each phase to be between $[0,90]$. We allow for a phase to have 0 seconds which means that it can be skipped. \par
For each intersection, we first generate phase 1 duration ($p_1$) by randomly choosing one integer in the range of $[0,90]$ seconds. Then we generate phase 2 duration ($p_2$) by randomly choosing one integer in the range of $[0, 90-p_1]$. Then we generate phase 3 duration ($p_3$) by randomly choosing one integer in the range of $[0,90-p_1-P_2]$. At last we calculate the phase 4 duration ($p_4$) as $90-p_1-p_2-p_3$. \par

3. \textbf{Fitness function calculation}: for each individual we calculate the fitness function by decoding the chromosomes to phase durations, updating the traffic signal timing according to the chromosome and running a simulation model of the network for static user equilibrium. We used Aimsun as our simulation tool to generate the fitness function. Within this function, we first call the Aimsun traffic simulation model to assign the preset OD demand to the network and then run a microscopic stochastic route choice model to obtain the total travel time recorded in the network. At last we use the reverse of the total travel time as the fitness value.\par
4. \textbf{`Reach the maximum number of iterations?”}: this module checks if the maximum number of generations has been reached; if not, it proceeds to the following steps. \par
5. \textbf{``Tournament”}: This module is used in order to obtain two parents from the last generation as a preparation for the next generation. In this module, we randomly select two chromosomes from the population, followed by a tournament between these two chromosomes and comparing their fitness function values. Higher valued chromosome won this tournament and we return the winner as one of the parents.\par
6. \textbf{``Crossover”}: Two chromosomes are selected using the ``tournament” module, and the crossover happens under a preset probability (called probability of crossover. For each child, an inherent index $x_{inherentis}$ randomly selected as a float which is in the range of $(0,1)$. Then the child’s chromosome is calculated as in Equation \ref{eq9}.
\begin{equation}\label{eq9}
    Child = Father \times x_{inherent}+Mother \times (1-x_{inherent})
\end{equation}

7. \textbf{``Mutation”}: Mutation changes the chromosome in children in a preset probability (called probability of mutation). In this application, mutation function only mutates between phases within one intersection. The reason is to maintain the cycle time in each intersection. For example, one child has a chromosome of:
$$[p_{11},p_{12},p_{13},p_{14},     p_{21},p_{22},p_{23},p_{24},…    ,p_{n1},p_{n2},p_{n3},p_{n4} ]$$ \par 
We then randomly select: a) an intersection $u$ b) two phases $v$ and $w$ from this intersection and c) the variation (Var) within the range of $(0, p_{uv})$. The new duration of phases v and w are calculated as: $p'_{uv}=p_{uv}-Var , p_{uw}=p_{uw}+Var$. The rest phase durations of this child remain the same.\par
8. \textbf{``GA optimization”}: continue to the next generation by going to step 2 until the stopping criteria has been reached (in our case the hyper tuned maximal number of generations has been reached).

\subsection{BGA-ML}

\begin{figure*}[!t]
\centering
\includegraphics[width=7in]{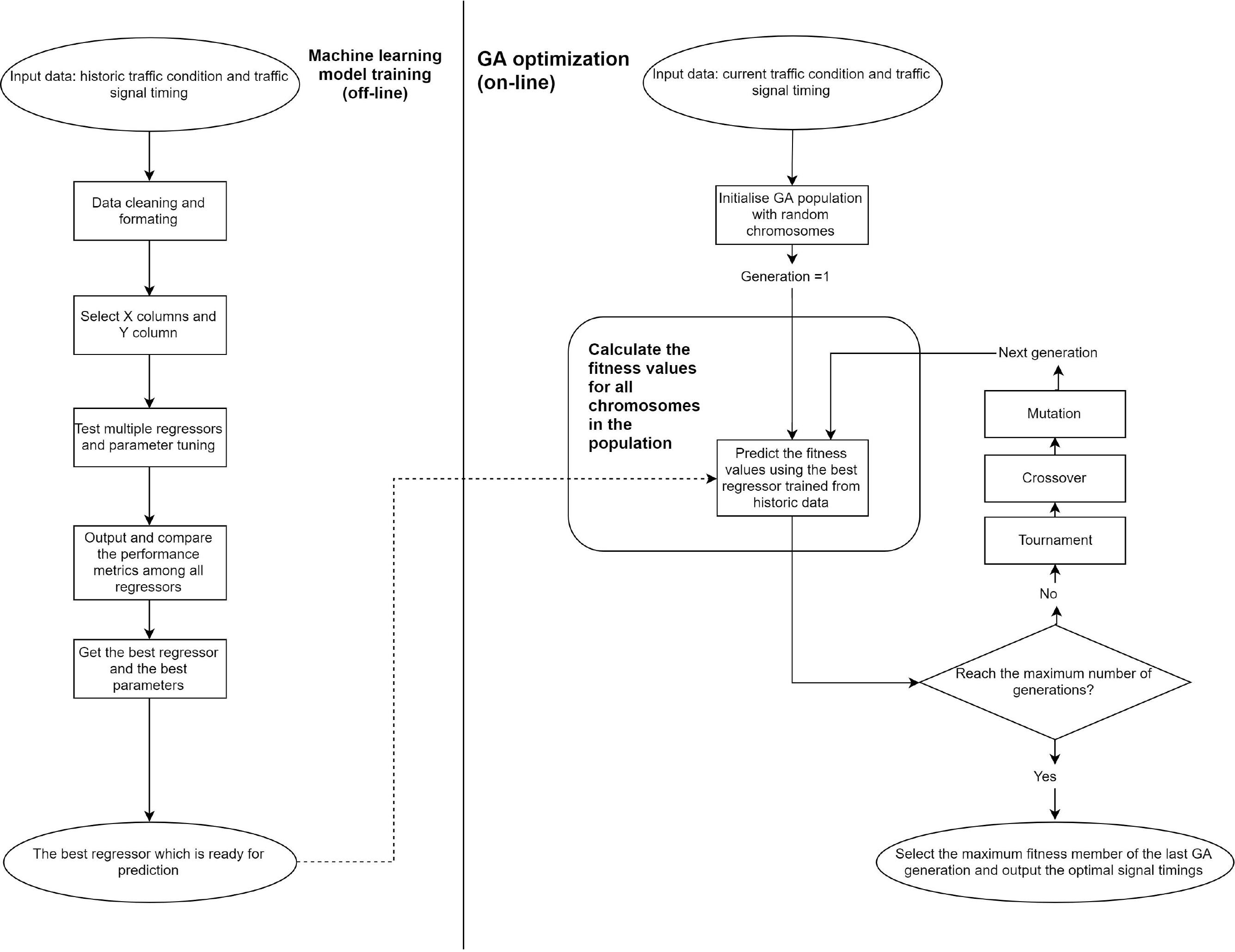}
\caption{BGA-ML optimization process.}
\label{fig_GA_ML}
\end{figure*}

The new proposed boosted genetic algorithm in this paper makes use of the GA structure presented in the previous section and adds the machine learning component in parallel, as presented in Figure \ref{fig_GA_ML}. The machine learning part is trained offline and the BGA-ML process will be launched online whenever there is a reported accident. The following steps describe how the machine learning parts are interconnecting with the GA parts with the purpose of reducing the state space search and predicting the most likely phase duration to be chosen based on previous trained data sets. The biggest advantage of this approach is reducing the time that genetic algorithms spend in creating the initial and subsequent populations, and to learn from previous iterations in the past which were the best choices that meet the optimization criteria, instead of always starting from random and new combinations which need intensive simulations to be run multiple times.

\subsubsection{Optimization process}
Compared to the previous optimization approach in the GA algorithm, here the ML model will replace the traffic simulation; therefore it will have the same role as the traffic simulation which is to produce (more specifically to predict) the total travel time for different scenarios. This helps to reduce the computational time taken by running the traffic simulation for each new phase combination that the GA generates.\par

\subsubsection{BGA-ML Framework}
In this paper, we focus on the proof of concept of the BGA-ML framework applied on a limited data set of a possible traffic accident in our case study network. More specifically, we use the data generated by various simulation runs in the GA experiments in which only the selected incident is introduced and the ML model is trained with the specific capacity drop caused by this incident. The details of the incident will be described in the Section \cref{case_study}. 

Note that the framework can be further extended and trained with a larger data set and random traffic disruptions. One possibility is to divide the big network into small sub-networks with similar characteristics. Other solutions can be clustering the road sections with similar macroscopic fundamental diagram (MFD). Therefore, further research will need to consider clustering of regions/road sections/sub-networks that has the similar characteristics (this is an extension which we aim for in the future).\par 

After the collection of the output data from previous simulation runs, we process and format it into a new data frame in Python, in order to build the feature necessary for training the machine learning models. Furthermore, we select the most important features that will be used for ML training and testing as detailed in the next sub-section ``Feature Generation". Lastly, we apply different regression models to the training data set and validate them using a set of various performance metrics as detailed in \cref{Performance_metrics}. By doing multiple tests over the performance of each ML model under different parameter variations, we will determine the best regression model for our problem with its best hyper-tuned parameters.\par

\subsubsection{Feature generation}\label{feature_generation}
As previously indicated, we record the output data while running the original GA using the traffic simulation with a traffic incident in one of the links of the network and use it as the training data set of our ML models. It's very important to keep the training data set consistent with the GA mechanism to ensure the compatibility between ML and GA integrated optimization framework.\par
The ML models are trained to predict the total network travel time by using the following features:\par
•   \textbf{the traffic state of the network}: In this paper, we simulate the incident for one hour simulation and output the total travel time for the one-hour period, therefore, we recorded profiled traffic status at each 10-minute time interval. Key features recorded are: section capacity (considering the capacity drop after the incident), section flow and section speed. Similarly, we record these features for each link at each ten minute period. To summarize, we collect a total of $72$ links $\times$ $3$ feature/link $= 216$ features. \par
•   \textbf{the traffic signal plan of the signalized intersections:} this is the same as the decision variable in GA which is a vector of all phase durations for all signalized intersections. In this network, there are 4 signalized intersections, and each intersection has 4 phases, therefore we record 16 features which represent the signal plans. \par

Overall, after creating the feature matrix we have obtained a total of 232 columns, and 9743 data records to be used for the model training, validation and testing.

\subsubsection{The regression models}
Choosing the best fitted regression model that can be used for the double-layer optimization framework is not a trivial task and before making any decision, we used four different regression models including: Gradient Boosting Decision Tree (GBDT), Extreme-Gradient Boosted Decision Tree (XGBT), Random Forest (RF) and Linear Regression (LR).
\textbf{GBDT} is an refined machine learning technique based on Decision Trees (DT) \cite{Friedman2001}\cite{Freund1999}\cite{Schapire1999}\cite{Friedman2002}. Boosting can be interpreted as an optimization algorithm on a suitable cost function \cite{Breiman1997} while the DT is a decision support tool which contains a tree-like structure and have been used extensively for various prediction approaches in either classification or regression problems. In a typical DT, each node that is inside the tree represents a decision making procedure, each branch represents the outcome of the decision making, and each leaf node represents a class label. The paths from root to leaf represent the classification rules or the prediction path. 
\textbf{XGBT} is a enhanced version of GBDT \cite{Chen:2016:XST:2939672.2939785} by introducing a regularization parameter in the learning objective function (to control over-fitting); it also introduces a sparsity awareness algorithm for parallel tree learning and has a better support for multi-core processing (this make it very appealing for real-time applications. Only recently they have started to gain more popularity and be applied successfully, for example, for incident duration classification or regression (see \cite{Mihaita2019}).
\textbf{RF} is an ensemble learning method which constructs multitude of DTs at training time and outputs the class that appears most often in classification or the mean prediction of the individual trees in regression \cite{liaw2002classification,breiman2001random}.
 \textbf{LR} is a linear approach to modelling the relationship between an dependent variable and one or more independent variables and is taken here as a baseline of the prediction outcome validation of more advanced machine learning models presented above \cite{weisberg2005applied,montgomery2012introduction}. 

\subsubsection{Performance metrics}\label{Performance_metrics}
In order to compare the performance of each regressor and evaluate their accuracy and performance, we considered several performance metrics such as: the Mean Absolute Error (MAE), the Root Mean Squared Error (RMSE), R Squared ($R^2$), Mean Absolute Percentage Error (MAPE). MAE is a measure of difference between two continuous variables calculated as:\par
\begin{equation}
    \label{eq_mae}
    MAE=\frac{\sum_{i=1}^n |y_i-x_i|}{n}=\frac{\sum_{i=1}^n |e_i|}{n}
\end{equation}
where $x_i$ is the prediction and $y_i$ is the true value, therefore the absolute errors is $|e_i|=|y_i-x_i|$.\par
MSE is an estimator which measures the average of squares of the errors and it's calculated as:
\begin{equation}
    \label{eq_mse}
    RMSE=\sqrt{\frac{1}{n}\sum_{i=1}^n (x_i-y_i)^2}
\end{equation}
where the squared errors is  $(x_i-y_i)^2$.\par
$R^2$ is the proportion of variance in the dependent variable that is predictable from the independent variable(s). The $R^2$ provides a measure of how well observed outcomes are replicated by the model based on the proportion of total variation of outcomes.

MAPE is a measure of prediction accuracy of a forecasting method which usually expresses accuracy as a percentage as indicated below:
\begin{equation}
    \label{eq_mape}
    MAPE=\frac{100\%}{n}\sum_{i=1}^n|\frac{y_i-x_i}{y_i}|
\end{equation}

\subsubsection{Hyper-parameter tuning}
The chosen machine learning algorithms have a set of hyperparameters – parameters related to the internal design of the algorithm that cannot be fit from the training data. In order to fine tune the dozens of parameters for each regressor that we have been using in our optimization framework, we perform a five-fold cross-validation (5CV) method when deciding the training and testing data sets. First, we randomly divide our whole data set into five folds which have the same size. Then we choose 4 folds as the training data set and use the remaining 1 fold as the testing data set. We will shuffle the folds five times and each fold serves as a test data set once. For each regression, we tune the hyperparameters on each training data set, at each learning fold using various random combinations, evaluated using the 5CV. A detailed discussion on all settings is further provided in Section IV-C of the Case Study analysis. When training the regression models, the average values of all performance metrics are recorded for further comparison.

\section{CASE STUDY}\label{case_study}
For showcasing the benefits of the proposed approach, a four-intersection network was designed in Aimsun \cite{Aimsun2012} and three scenarios are constructed in order to optimize the traffic signal timings under normal conditions and under traffic incident conditions. The GA model and BGA-ML model are then tuned by running multiple times using different parameter settings before converging towards the optimal parameters to be used in the case study. 
\subsection{Network Configuration}
\begin{figure*}[!t]
\centering
\subfloat[]{\includegraphics[width=3 in]{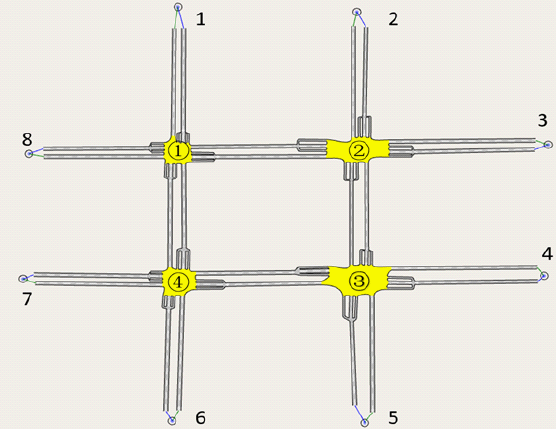}%
\label{fig_network}}
\hfil
\subfloat[]{\includegraphics[width=3in]{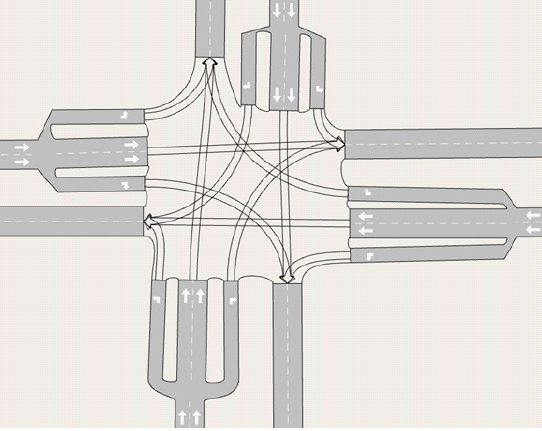}%
\label{fig_intersection}}
\caption{(a) Network layout and (b) intersection 1 layout.}
\label{fig_sim}
\end{figure*}

This network layout of the simulation model is shown in Figure \ref{fig_network} and is a left-hand drive model to accommodate the Australian road environment. The simulation duration is one hour and each intersection is a typical four-branch signalized intersection with dedicated right turning lane and dedicated left turn lane. The detailed layout of intersection $1$ is shown in Figure \ref{fig_intersection} as an example, and all the other intersections are configured in the same way.

\subsubsection{Configuration and traffic signals}
Each intersection has the same cycle time duration (which is 90 seconds) and the same number of phases (which is 4). The order of phases are fixed. Within each phase, the green granted movements are the same and fixed for all intersections. The only variable in signal configuration is the phase green times. The configuration of traffic signals for each intersection is shown in Table \ref{tb1}.

\begin{table}[!t]
\renewcommand{\arraystretch}{1.3}
\caption{Configuration of traffic signals for eahc intersection}
\label{tb1}
\centering
\begin{tabular}{|c|c|}
\hline
Phase ID & Traffic signal configuration (green movement highlighted)\\
\hline
1 & \includegraphics[width=2.5 in]{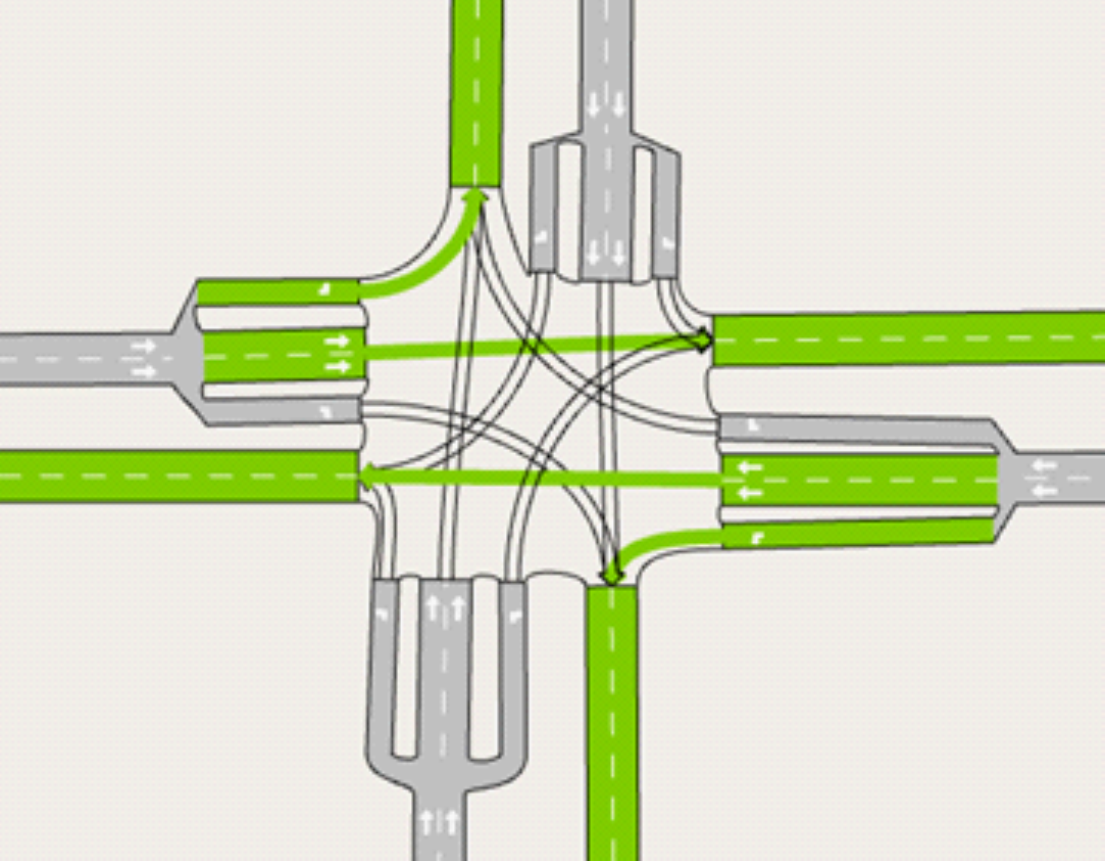}\\
\hline
2 & \includegraphics[width=2.5 in]{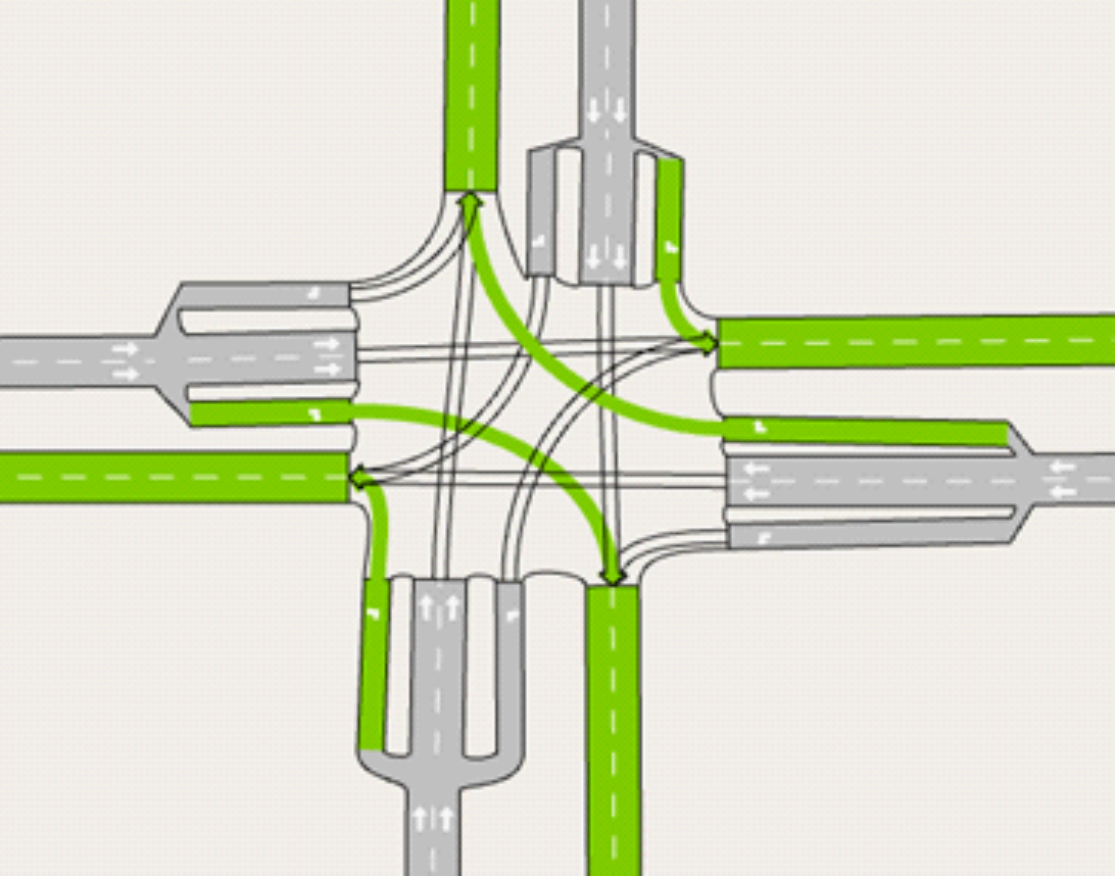}\\
\hline
3 & \includegraphics[width=2.5 in]{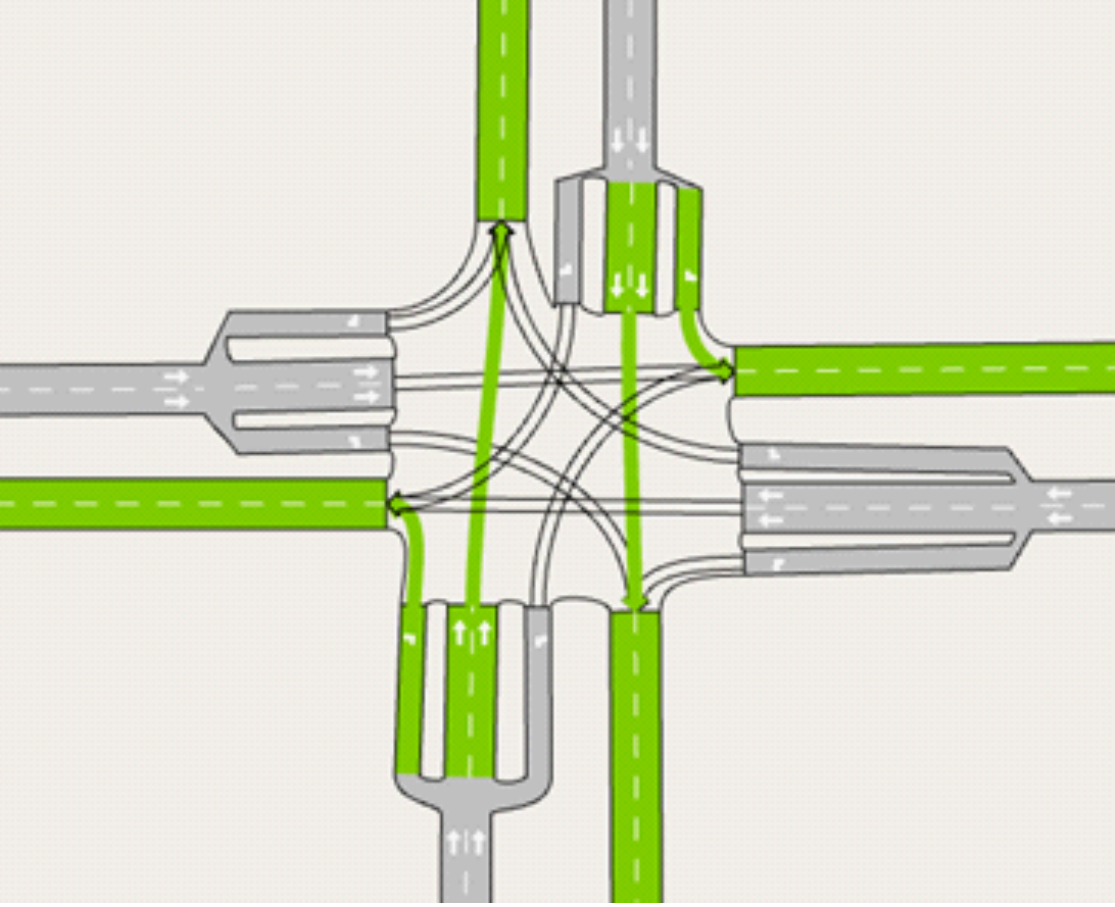}\\
\hline
4 & \includegraphics[width=2.5 in]{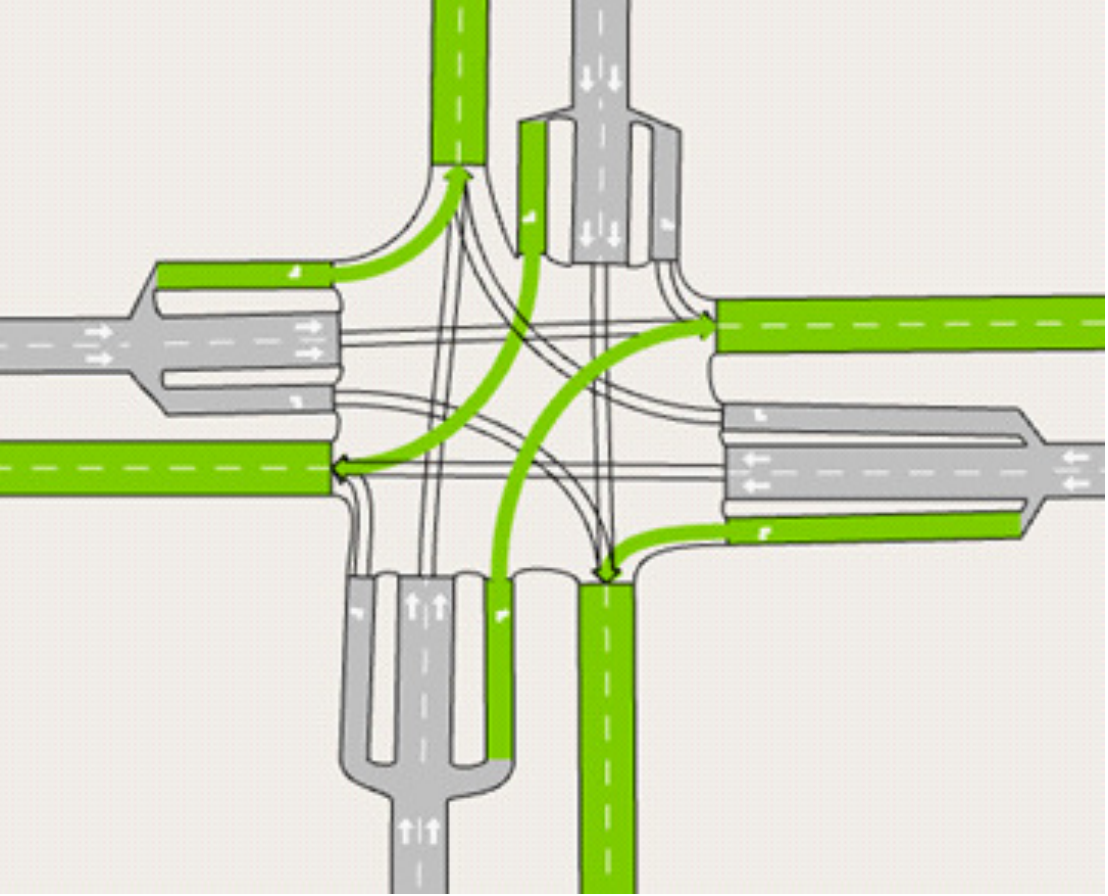}\\
\hline
\end{tabular}
\end{table}
\subsubsection{Traffic demand}
\begin{figure*}
    \centering
    \includegraphics[width =5 in]{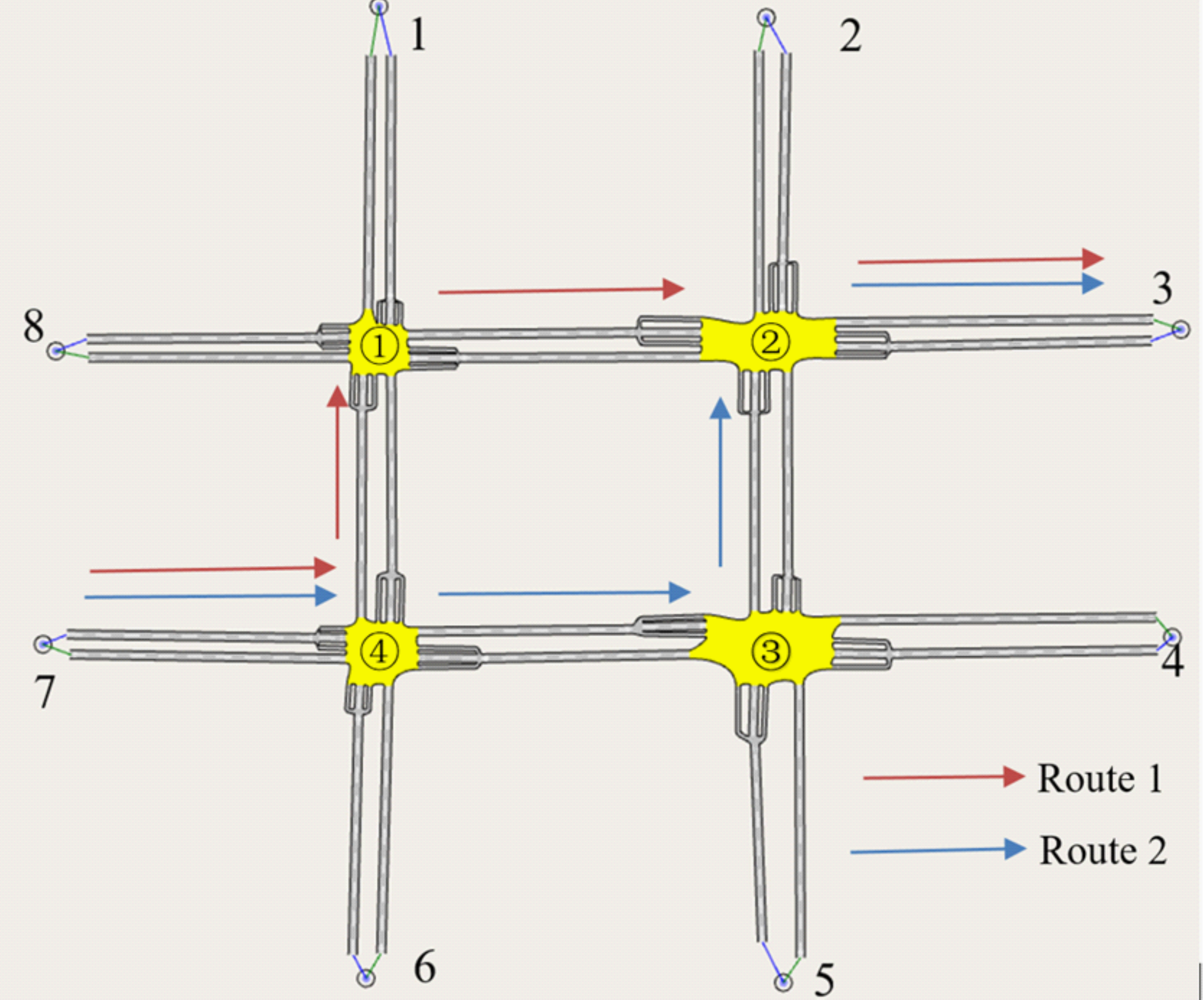}
    \caption{O-D index configuration}
    \label{fig4}
\end{figure*}

The O-D indexes are shown in Figure \ref{fig4} and the O-D trips for one-hour simulation are shown in Table \ref{tb2}. As highlighted in Table \ref{tb2}, a higher flow is set from centroid 7 to centroid 3. This O-D pair contains 2 routes, which are shown in Figure \ref{fig4}. Special attention will be paid in observing the flows on these two routes and how they are impact by traffic accidents as well as the optimization methods proposed in this paper.

\begin{table*}[!t]
\renewcommand{\arraystretch}{1.3}
\caption{Traffic demand}
\label{tb2}
\centering
\begin{tabular}{|c|c|c|c|c|c|c|c|c|c|}
\hline
From \addslash To &	1 &	2 &	3 &	4 &	5 &	6 &	7 &	8 &	Total\\
\hline
1 &	0 &	150 &	150 &	150 &	150  &	100 &	100 &	150 &	950\\
\hline
2 &	150 &	0 &	100 &	100 &	100 &	150 &	150 &	100 &	850\\
\hline
3 &	150	& 100 &	0 &	150 &	100 &	100 &	100 &	150 &	850\\
\hline
4 &	100 &	150 &	100 &	0 &	150 &	100 &	150 &	150 &	900\\
\hline
5 &	150	 &100 &	100 &	150 &	0 &	150	 &150 &	100	 &900\\
\hline
6 &	100 & 100 &	100	& 100 &	0 &	0 &	150 &	100 &	650\\
\hline
7 &	100 &	150 & {\colorbox{Yellow}{750}} &	150 &	150 &	100 &	0 &	150 &	{\colorbox{Yellow}{1550}}\\
\hline
8 &	100 &	150 &	150 &	100 &	150 &	100 &	100 &	0 &	850\\
\hline
Total &	850 &	900 &	{\colorbox{Yellow}{1450}} &	900	 & 800 &	800 &	900 &	900 &	7500\\
\hline
\end{tabular}
\end{table*}

\subsection{Aimsun simulation setups}
We follow the standard process of Aimsun simulation to generate the total travel time for each traffic signal plan,
by first updating the ``link green split” for each link connected to any signalized nodes using Equation \ref{eq6}. For those links which are not connected to any signalized nodes, the ``link green split" will be set to 1. Aimsun uses the ``link green split" to calculate the travel time for each link as it affects the flow exiting multiple interconnected links. For example, if one link has the ``link green split" of 0.4, this means that only 40\% of the time this link will be granted a green light. \par
Next we apply a static traffic assignment modelling scenario to obtain the initial link flows assigned to each of the road sections during a one-hour simulation set up for morning peak.\par 
The initial OD demand of our simulation will be further split and profiled into 6 time intervals of 10 minute each by running microscopic dynamic user equilibrium (DUE) scenario. This ensures a dynamic behavior of our traffic simulation modelling alimented by time-dependent OD matrices with adaptive traffic signal plans. Finally, the simulation output consists in the total travel time obtained at each each 10-minute time interval while  running microscopic DUE simulation. \par

\subsection{GA parameter tuning}
There are several parameters that need to be set up for the initial genetic algorithm creation, which are: the population size, the maximum number of generations, the crossover probability, and the mutation probability. These have been tuned with the computational time in mind as well and are detailed as follows:\par
\subsubsection{Population size and maximum number of generations}
Population size is the number of individuals in one population in one generation. In our experiment by individual we refer to a traffic signal plan which is represented by a chromosome noted as $[p_{11},p_{12},p_{13},p_{14},     p_{21},p_{22},p_{23},p_{24},…    ,p_{n1},p_{n2},p_{n3},p_{n4} ]$.
Maximum number of generations is the maximum number of how many evolutionary generations we will run in one optimization cycle. The max number of generations is determined by the performance of the fitness function and is set at the step after which the fitness function doesn’t improve anymore. 

In order to set these two parameters, we choose four possible combinations for our pilot experiment, which are shown in Table \ref{tb3}.

\begin{figure*}
    \centering
    \includegraphics[width= 6in]{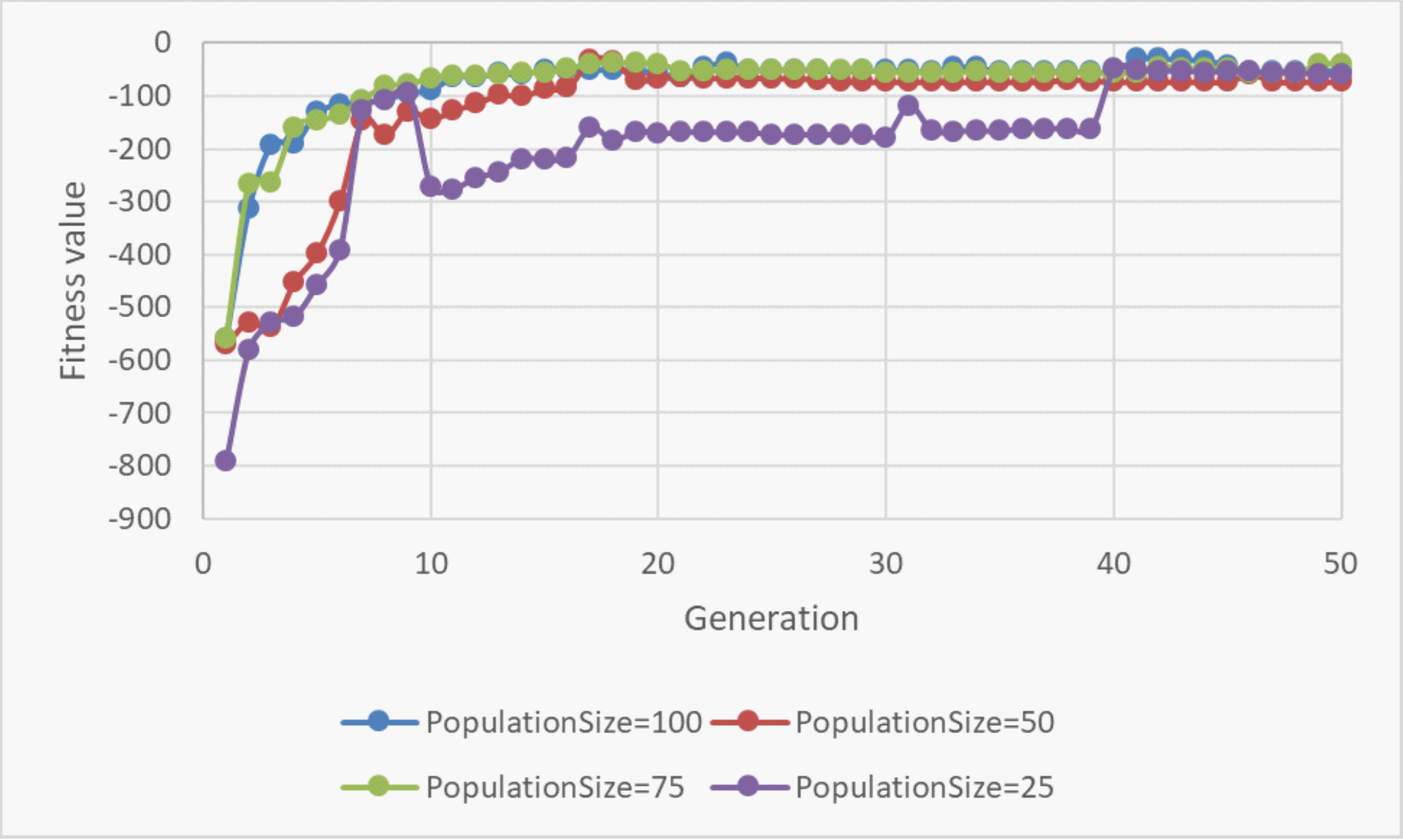}
    \caption{Fitness values for different combinations of population size and maximum number of generations}
    \label{fig_converge}
\end{figure*}

\begin{table*}[!t]
\renewcommand{\arraystretch}{1.3}
\caption{Experiment specifications}
\label{tb3}
\centering
\begin{tabular}{|c|c|c|c|c|c|c|c|c|c|}
\hline
Run ID &	Population size &	Maximum number of generations &	Tested generation size\\
\hline
1 &	25&	50&	[1,2,..50]\\
\hline
2&	50&	50&	[1,2,..50]\\
\hline
3&	75&	50&	[1,2,..50]\\
\hline
4&	100&	50&	[1,2,..50]\\
\hline

\end{tabular}
\end{table*}
As shown in Figure \ref{fig_converge}, the fitness values of final optimal traffic signal solutions are plotted for each generation and for four different population sizes (25, 50, 75 or 100 individuals). We have also tested larger population sizes and results indicated that the algorithm converges very fast after 75 - 100 individuals in a population without any further improvement; therefore we only show these 4 difference convergence rates for the above four population sizes in this subsection. The unit of the fitness value is ($vehicle \cdot hour$).  As we can see, the cases when the ``population size = 100” and ``population size = 75” present the same converging trends and they both converge simultaneously after the $20^{th}$ generation. The ``Population size = 50” case also converge at the $20^{th}$ generation, but shows a different converging trend compared to the ``100” and ``75” cases, which is slower in the initial stages of less than 10 generations. The ``population size = 25” case converges at the $40^{th}$ generation, which is the slowest case to convergence, therefore we exclude it as a possibility for the best optimization setup.

As a plus, the optimal traffic signal settings are recorded and the phase durations in intersection 1 and intersection 3 are shown in Figure \ref{fig_conv2} and \ref{fig_conv3} respectively. The phase durations for other intersections can be found in Appendix A.
Once again, the fitness value converges at the $20^{th}$ generation for all population sizes except the ``population size = 25” case, where the phase durations still not converge towards the optimal values and reiterating once again this would be a unwise setting. 

\begin{figure*}
    \centering
    \includegraphics[width =6 in]{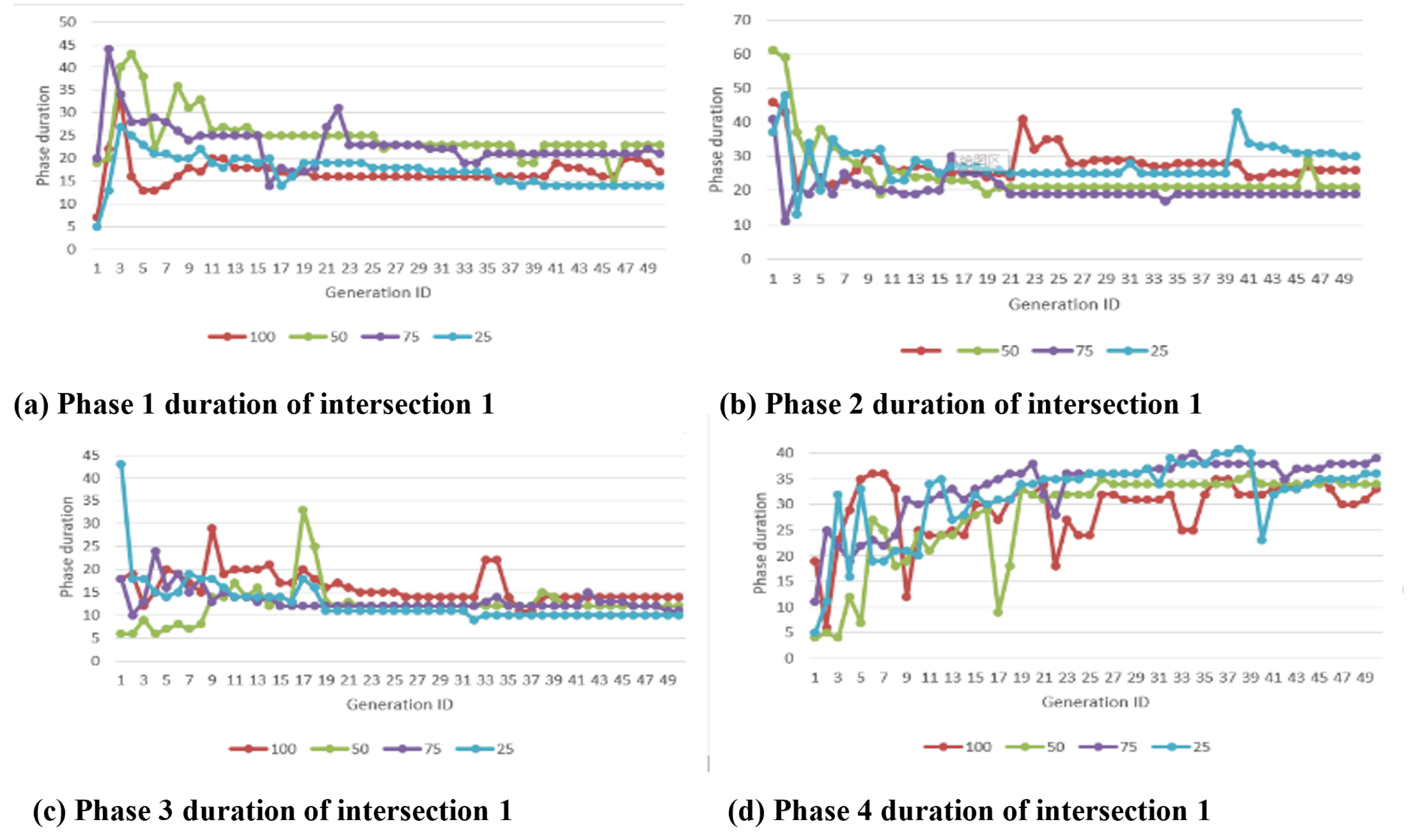}
    \caption{Phase durations of intersection 1}
    \label{fig_conv2}
\end{figure*}

\begin{figure*}
    \centering
    \includegraphics[width =6 in]{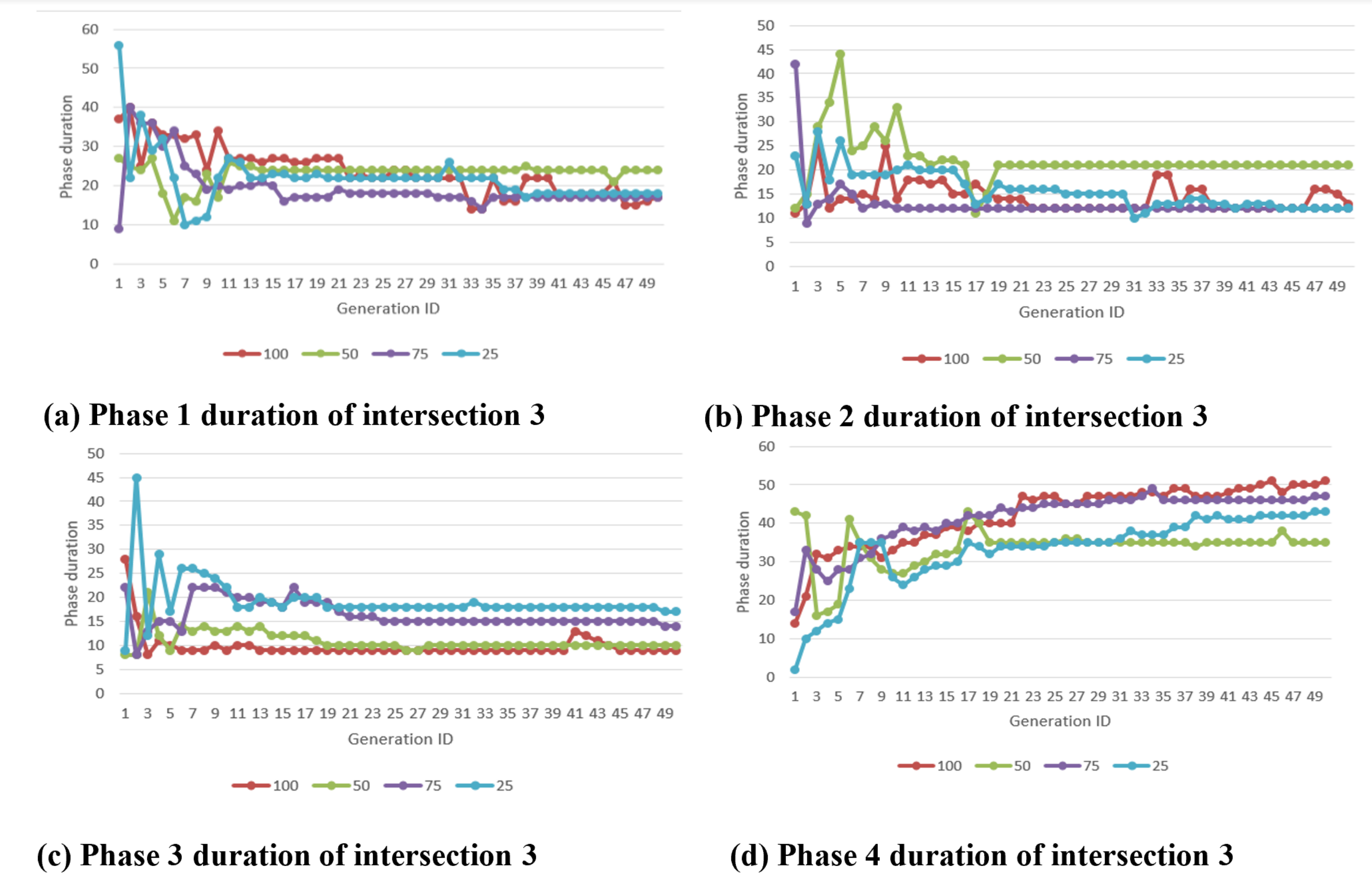}
    \caption{Phase durations of intersection 3}
    \label{fig_conv3}
\end{figure*}

\subsubsection{Probability of crossover}
This parameter enables to inherit a good fitness from the last generation to a new generation; a higher value guarantee a fast convergence, so the probability of crossover is set to 0.8 in all experiments in this paper. Because mutation is applied to all individuals independently after crossover for each generation, setting up a high crossover probability wouldn't affect the mutations.  

\subsubsection{Probability of mutation}
Mutation generates new chromosomes which enrich the gene library and is a double-edge sword. On one hand, the mutation may happen to a chromosome with bad fitness values and transform it into a chromosome with better fitness values. On the other hand, mutation creates noise to the convergence of GA and diversity so the algorithm can jump out of local optima. In order to avoid noise in convergence, the mutation probability is set to 0.1 in all experiments of this paper.

\subsubsection{Computational time}
\begin{figure*}
    \centering
    \includegraphics[width = 6in]{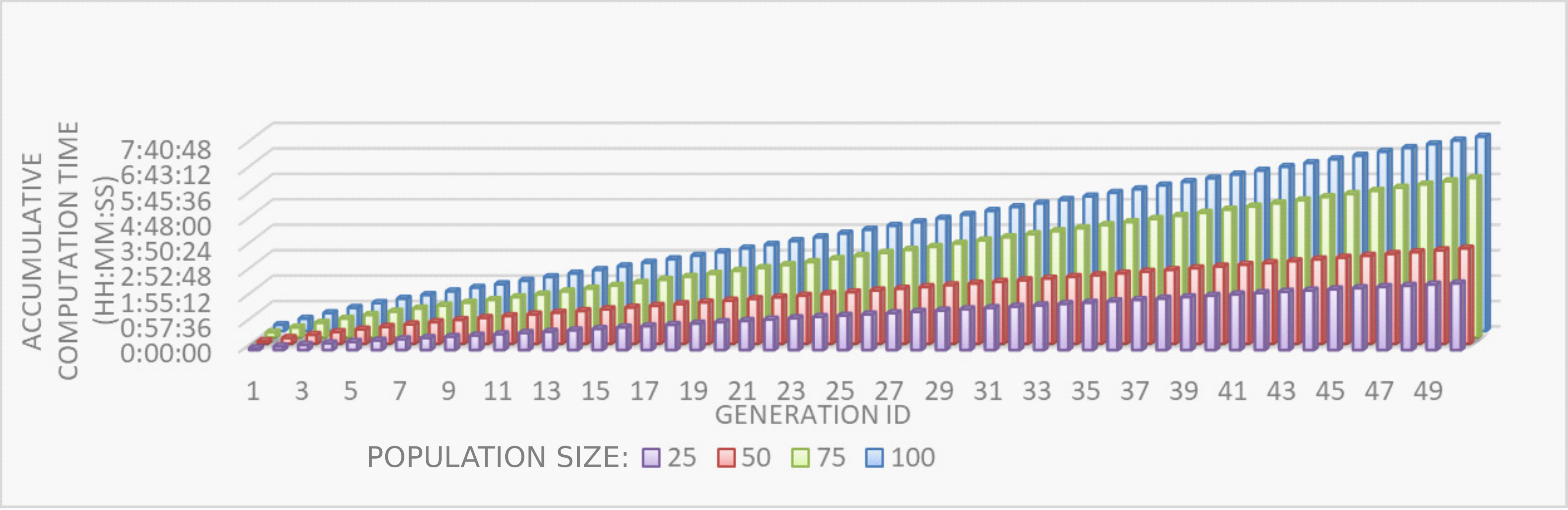}
    \caption{Accumulative computational time}
    \label{fig_compute1}
\end{figure*}

\begin{figure*}
    \centering
    \includegraphics[width = 6in]{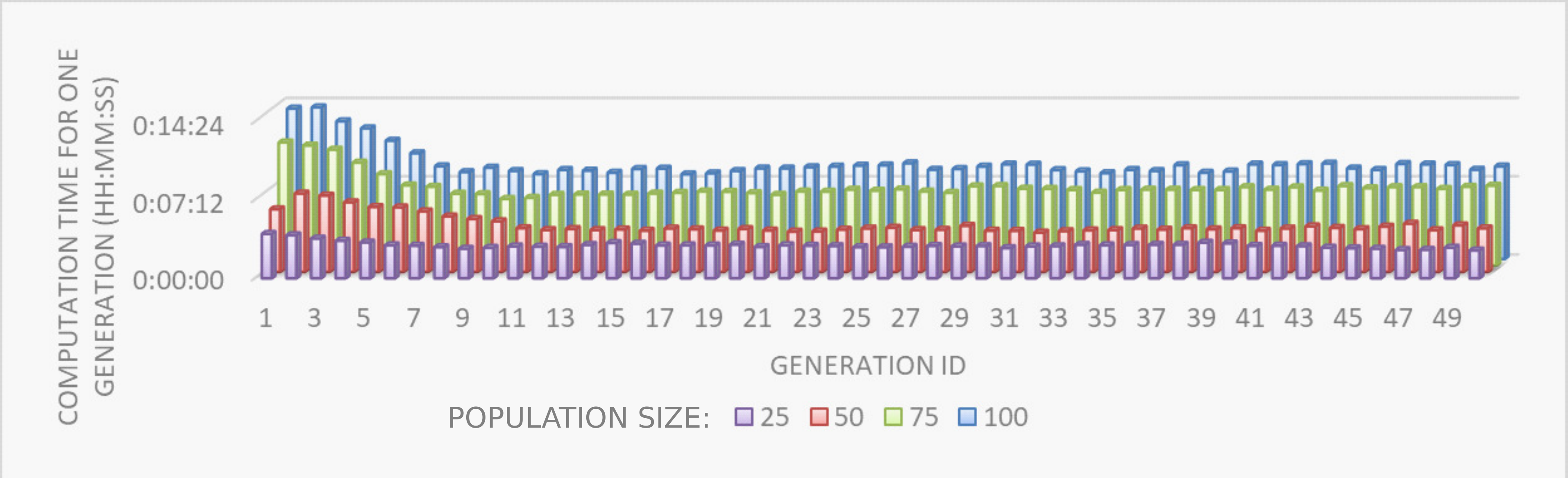}
    \caption{Computational time for one generation}
    \label{fig_compute2}
\end{figure*}

Computational times are recorded at the beginning and the end of a generation. Figure \ref{fig_compute1} shows the accumulative computational time of each generation with different population sizes. It shows the linear relationship between accumulative computational time and generation ID. Figure \ref{fig_compute2} shows that the first 10 generations always consume more time than the rest of generations and after 10 generations, each generation takes the same time to be finalised. Besides, we notice a linear relationship between the computation time and the population size for the same generation ID which indicates that the computational times of GA do not increase exponentially with the size of the population.

\subsubsection{Optimal parameter choice}
From previous combinations of GA parameters and observations mentioned above, we determine a set of parameters with fast and stable convergence and relatively short computation time. For our study, the maximum number of iterations is set to 20, the population size is set to 75, the crossover probability is set to 0.8, the mutation probability is set to be 0.1, which guarantees an average computational time of about 7 minutes per complete generation run.

\subsection{BGA-ML Parameter tuning}
As previously mentioned, because we use the different kinds of regression models there are a lot of hyper-parameters within the models. The data used to train the ML models are the ``initial state of the network" and the ``traffic signal setting of the signalized intersections". In our network, there are 72 sections and each section we recorded the speed, flow and density over each of the ten-minute period of the simulation. In addition, we have 4 intersections which contain 4 phases, therefore, in total we had $3*72+16=232$ features considered in the ML training. We ran 10,000 simulations using randomized traffic signal control plans and saved the 232 features in a database external tot he traffic simulation model. After cleaning the runs with repeated traffic signal control plans, we have 9,743 good runs in our database. 

\subsubsection{Hyper parameter tuning}
For all of our regression models, we use the randomized search as the searching method \cite{bergstra2012random} and 5CV as the cross validation method, so there are several important parameters which are tuned within the random search algorithm such as: \textit{ n\_iter, scoring, n\_jobs}. In general, \textit{n\_iter} is the number of random search iterations, \textit{scoring} is the defined model evaluation rules which follows the conventional scheme: higher return values are better than lower return values \cite{pedregosa2011scikit}. For example, the ``accuracy" scoring means that the higher accuracy values are better than lower accuracy values. At last, \textit{n\_jobs} is the number of processors used for parallel computing.   \par
In our GBDT and XGBT regression models, we considered \textit{ max\_depth, learning\_rate, n\_estimators, subsample} \cite{pedregosa2011scikit} as the main parameters to be hyper-tuned, where: \textit{ max\_depth} represents the maximum depth of the individual regression estimators (each estimator is a decision tree (DT)), \textit{learning\_rate} is the contribution of each tree to the overall outcome, \textit{n\_estimators} is the number of boosting stages to perform, and \textit{subsample} is the fraction of samples to be used for fitting the individual base learners (if smaller than 1.0 this results in Stochastic Gradient Boosting).  \textit{subsample} parameter interacts with the  \textit{n\_estimators} parameter. Choosing $\textit{subsample} < 1.0$ leads to a reduction of variance and an increase in bias.\par

Table \ref{tb_ml} shows all the parameters we have tested and their ranges.\par  
\begin{table}[!t]

\renewcommand{\arraystretch}{1.3}
\caption{Parameter range specifications for ML models}
\label{tb_ml}
\centering
\begin{tabular}{|c|c|c|c|c|c|c|c|c|c|}
\hline
Parameter name & Range\\
\hline
\textit{ n\_iter} & $\{50,100,150,200\}$\\
\hline
\textit{scoring} & $\{MAE,RMSE,MAPE,R^2\}$\\
\hline
\textit{n\_jobs} & 12\\
\hline
\textit{ max\_depth} & $\{3,5,7,9,11,13,15\}$\\
\hline
\textit{ learning\_rate}&$\{0.0001,0.001,0.1\}$\\
\hline
\textit{ n\_extimators} & $\{20,21,22,23,...,198,199,200\}$\\
\hline
\textit{ subsample} & $\{0.6,0.7,0.75,0.8,0.85,0.9,0.95,1.0\}$\\
\hline
\end{tabular}
\end{table}

\subsubsection{Optimal parameter selection for BGA-ML}

\begin{figure*}
    \centering
    \includegraphics[width=6in]{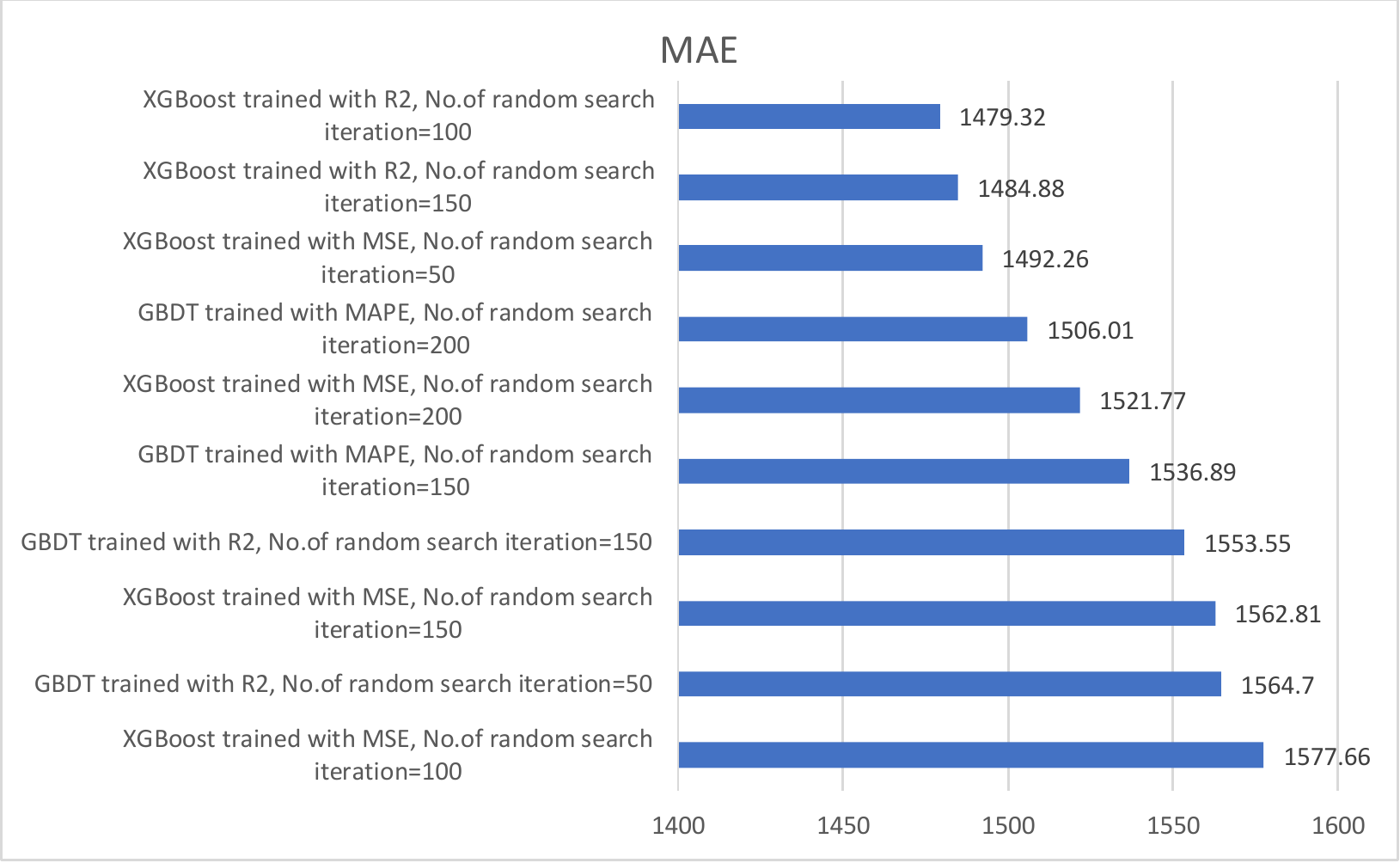}
    \caption{Top 10 regression models achieving the lowest MAE}
    \label{fig_top10MAE}
\end{figure*}

\begin{figure*}
    \centering
    \includegraphics[width=6in]{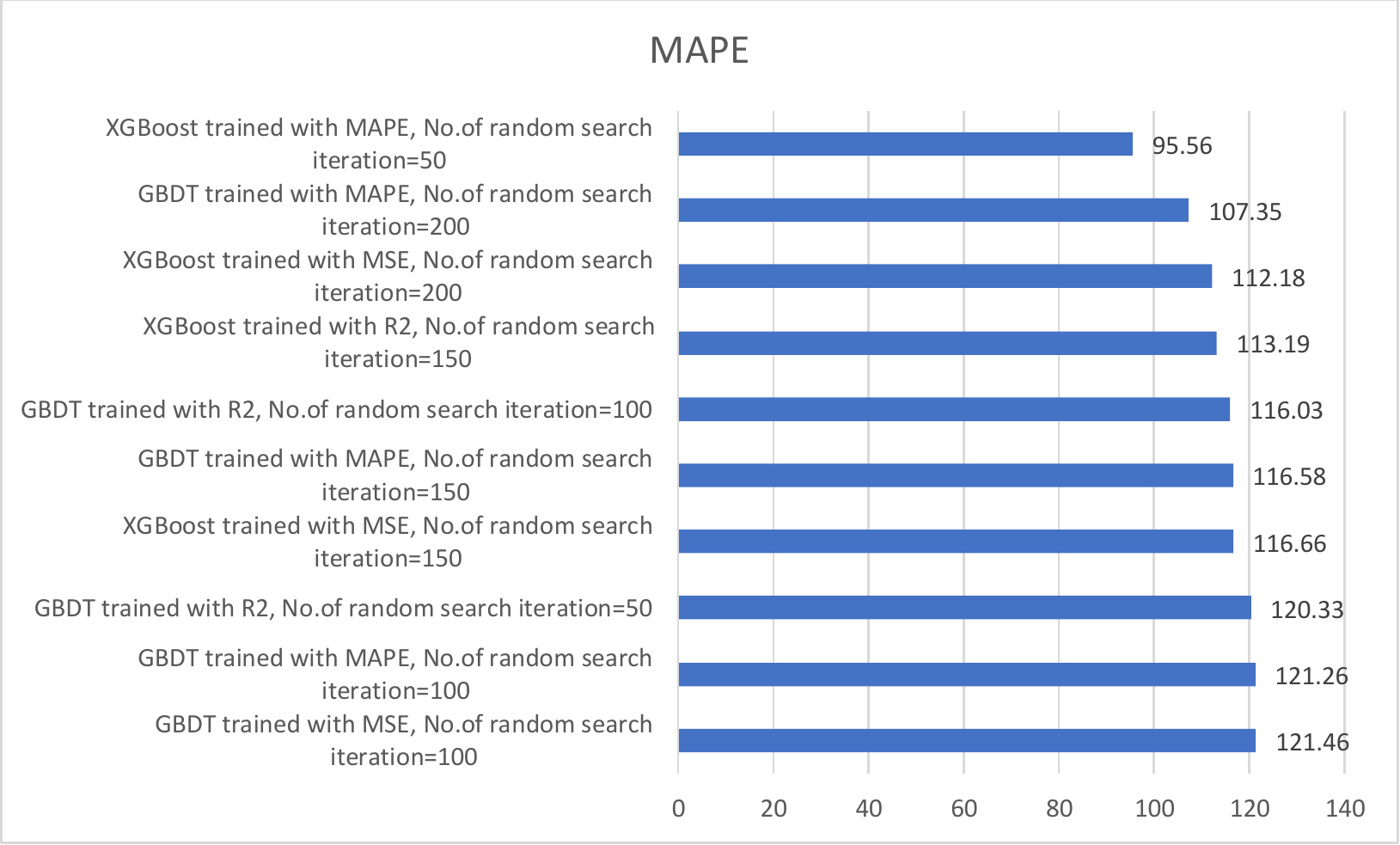}
    \caption{Top 10 regression models achieving the lowest MAPE}
    \label{fig_top10MAPE}
\end{figure*}

\begin{figure*}
    \centering
    \includegraphics[width=6in]{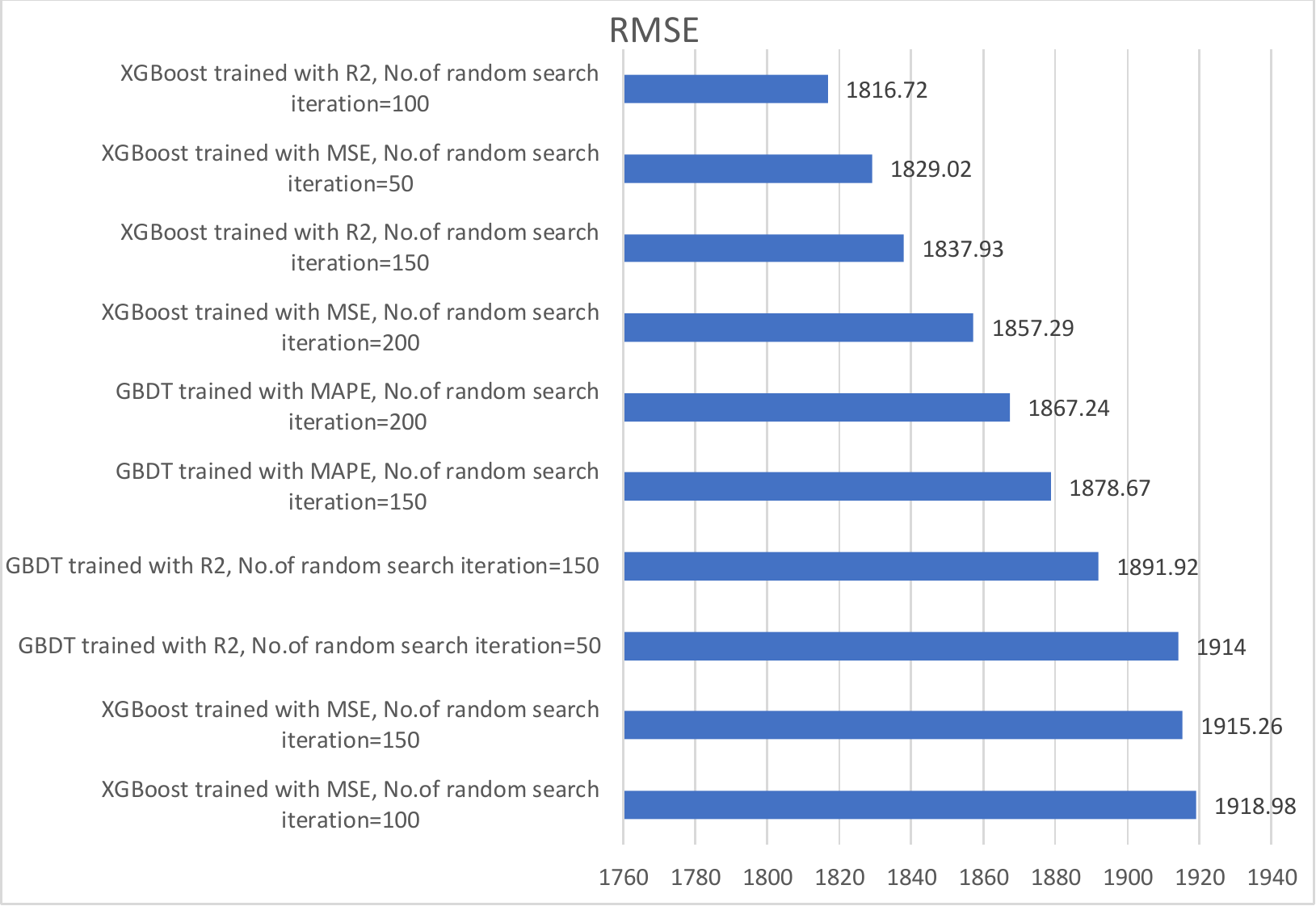}
    \caption{Top 10 regression models achieving the lowest RMSE}
    \label{fig_top10RMSE}
\end{figure*}

\begin{figure*}
    \centering
    \includegraphics[width=6in]{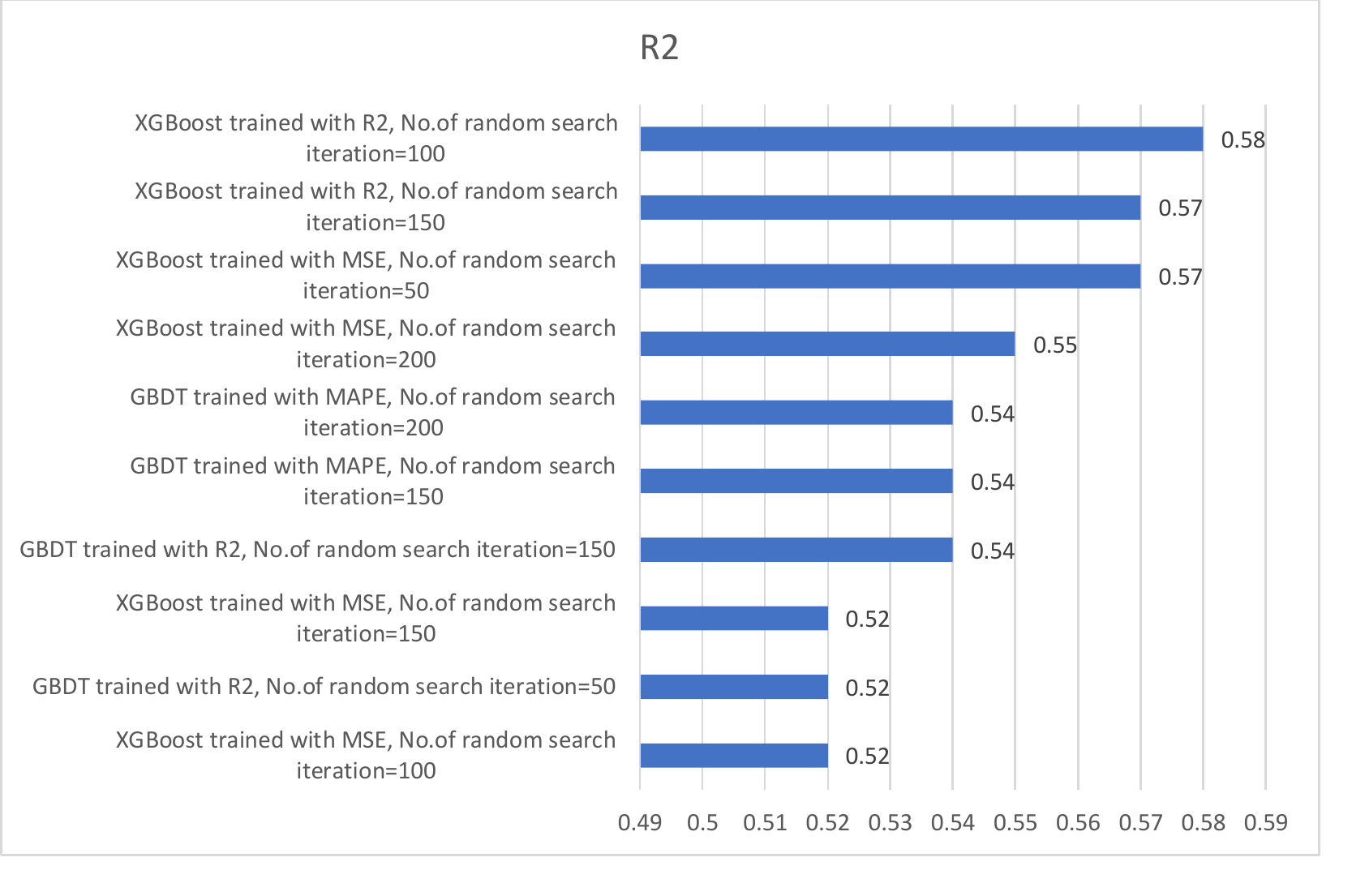}
    \caption{Top 10 regression models achieving the highest R2}
    \label{fig_top10R2}
\end{figure*}

We first test the parameters used in the randomized search and 5CV which are \textit{ n\_iter} and \textit{ n\_iter}. Then we record all the performance metrics for all the models in all parameters. Then we rank all the regression models according to each performance measure presented previously. Figures \ref{fig_top10MAE},\ref{fig_top10MAPE},\ref{fig_top10RMSE} and \ref{fig_top10R2} show the top 10 regression models which have been been winning across all combinations, and evaluated for each performance measure. We make the observations that the number of combinations to be plotted is very large and these figures have been selected to represent the best performing models under the best parameter setting.

After observing the top 10 lists shown in these figures, we can conclude that:\par
1. \textbf{XGBT trained with $R^2$} over $n\_iter = 100$, is no doubt the best regression model because it's the top 1 against three of the performance metrics (MAE, RMSE, and $R^2$) and is on the $14^{th}$ place when evaluated with MAPE = 126.91.\par
2. \textbf{XGBT trained with MSE} over $n\_iter = 50$ is also a good regression model because it's always in the top 3 for three of the performance metrics (MAE, RMSE, and $R^2$) and is on the $14^{th}$ place with MAPE = 124.17. \par 
3. \textbf{XGBT trained with MSE} over $n\_iter = 200$ is the all-rounder with good reliability because it's in top 5 for all four metrics. \par
4. \textbf{GBDT trained with MAPE} over $n\_iter = 200$ is also a reliable regression model because it's in top 5 for all four metrics; however it slightly underperforms compared to the third case presented above. \par

Overall from all results obtained we observed that XGBT and GBDT outperforms RF and LR in all measures in the performance metrics. We choose \textbf{XGBT trained with $R^2$} over $n\_iter = 100$ as our best regression model and therefore, we further run XGBT multiple times to obtain the best sets of values for \textit{ max\_depth},\textit{ learning\_rate}, \textit{ n\_extimators}, \textit{ subsample}. Finally, we conclude on the best parameters for the chosen model: \\
``\textbf{XGBRegressor (base\_score=0.5, booster=`gbtree',\\
\tabto{2.4cm} colsample\_bylevel=1,colsample\_bynode=1,\\
\tabto{2.4cm} colsample\_bytree=1, gamma=0,\\
\tabto{2.4cm} importance\_type=`gain', \\
\tabto{2.4cm} learning\_rate=0.1,\\
\tabto{2.4cm} max\_delta\_step=0, max\_depth=7,\\
\tabto{2.4cm} min\_child\_weight=1, missing=None, \\
\tabto{2.4cm} n\_estimators=190, nthread=None,\\
\tabto{2.4cm} objective=`reg:linear', random\_state=0,\\
\tabto{2.4cm} reg\_alpha=0, reg\_lambda=1,\\
\tabto{2.4cm} scale\_pos\_weight=1, seed=None,\\
\tabto{2.4cm} silent=None, subsample=0.6, verbosity=1)}"\par

\begin{figure*}
    \centering
    \includegraphics[width=5 in] {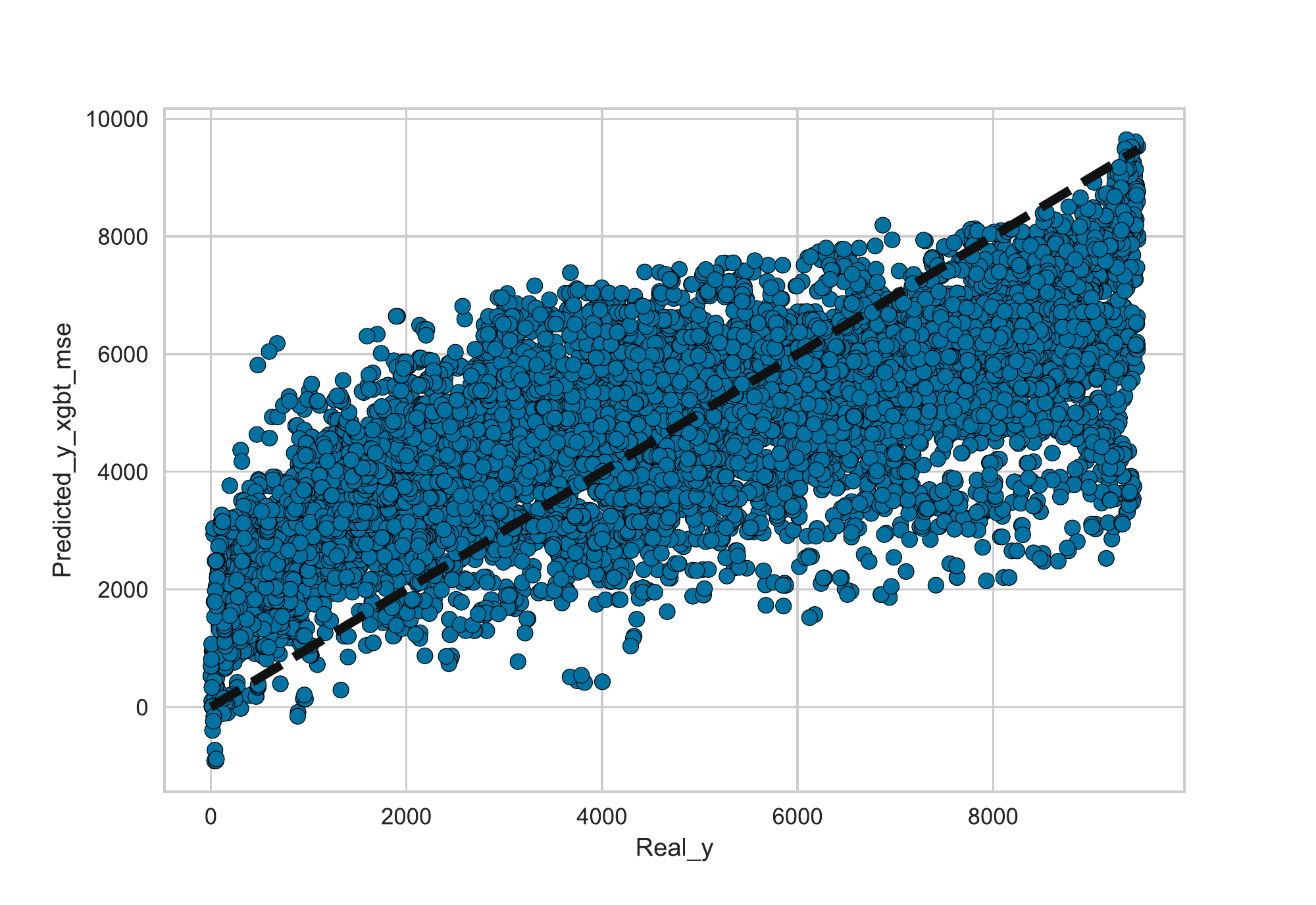}
    \caption{Predicted vs. real data using the best XGBT regression model after the optimal hyper-parameter tuning.}
    \label{fig_y_xgbt_mse}
\end{figure*}

Figure \ref{fig_y_xgbt_mse} presents the predicted travel time (y axis) [in seconds] and the real travel time (x axis) [in seconds] using the best performance regression model chosen above. It shows that although the model is the best regressor over all models, there is room for improvement due to high noise and large variation in the training data sets which have been obtained from traffic simulation modelling, not from real-world set-up; an ideal extension of our work is to use the training data set as from real intersection set-up; this would require however extensive information to be provided by management centers which are not always stored for optimization purposes. Finally, we will use this model as ready-to-use prediction model in the BGA-ML optimization. \par

\subsection{Scenarios}
Using the above GA and ML parameters previously fine-tuned, four scenarios are designed for our case study which are detailed below:\par
1.	\textbf{Regular traffic scenario which is using GA for traffic control optimization}: the proposed GA model will be applied to the ``no-incident network” and a simulation applying the optimal signal control (we can call it ``no-incident optimal signal control”) is recorded.\par
\begin{figure*}
    \centering
    \includegraphics[width=5 in]{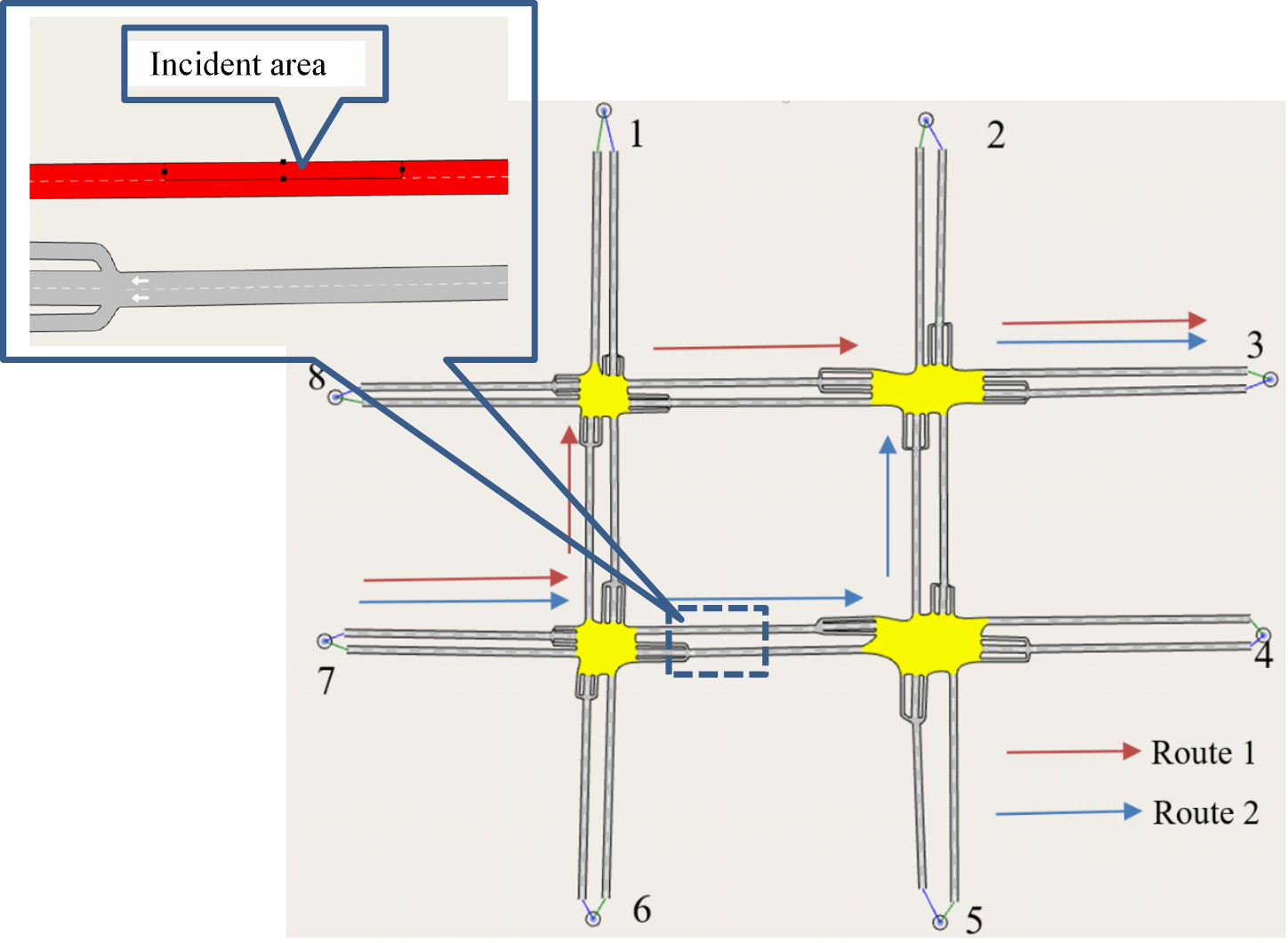}
    \caption{Traffic incident configuration}
    \label{fig_incident}
\end{figure*}

2.	\textbf{Traffic incident scenario without GA traffic control optimization}: an incident is created in the network at the location shown in Figure \ref{fig_incident} which will last for one hour. The incident blocks one lane of a two-lane link in route 2 from centroid 7 to centroid 3. The traffic flows on both route 1 and route 2 will be affected by this incident. The traffic signal plan in scenario 2 is the same as scenario 1. \par
3.	\textbf{Traffic incident scenario with the GA traffic control optimization}: the proposed GA model will be applied to the network and a simulation using the new optimal signal control will be recorded. \par
4.	\textbf{Traffic incident scenario with the BGA-ML traffic control optimization}: the proposed BGA-ML optimization framework will be applied to the network and a simulation using the new optimal signal control will be recorded. \par

\section{Results}
Tests on all four scenarios using corresponding optimization models are performed in our experiments. The following sections display the results of all the scenarios as well as their performance.
\subsection{Scenario 1: No incident scenario with GA }
Let’s denote ${a_i,b_i,c_i,d_i,i=1,..4}$  as the phases of each intersection, where $a_1$ is the first phase of intersection 1, $b_1$ is the second phase of intersection 1, etc. The outcome of proposed GA model returned the following optimal phase values [in seconds] of the whole network under no incident conditions: 
$$Optimal\ phase\ setting\  scenario\ 1=$$
$$\{18,22,12,38,20,19,15,36,17,12,17,44,30,22,9,29\}$$
The corresponding optimal fitness value is -22.41, which corresponds to a total travel time of 22.41 $vehicle\cdot hour$. \par
\begin{figure*}[!t]
    \centering
    \subfloat[phase 1]{\includegraphics[width=3 in]{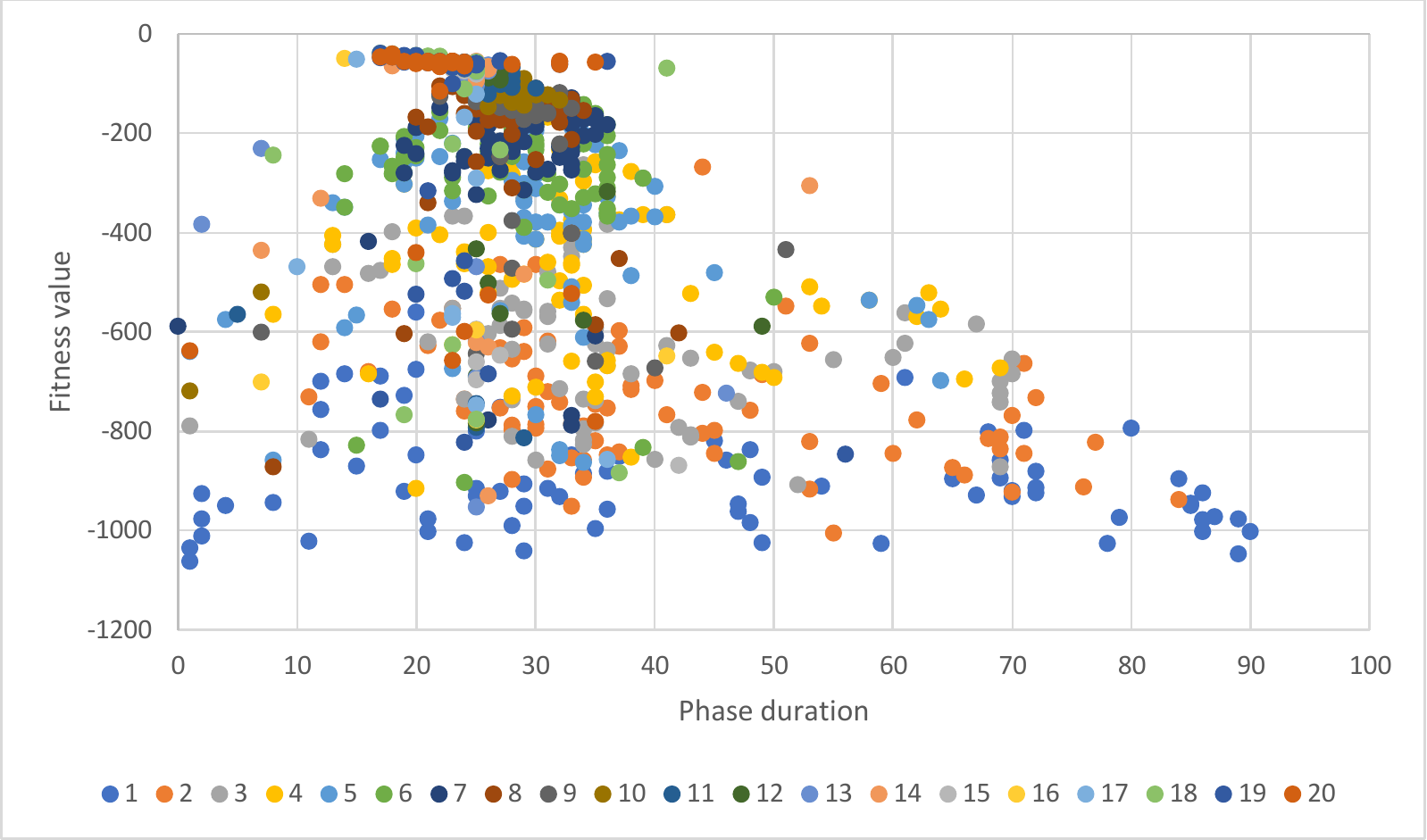}
    \label{fig_p11}}
    \subfloat[phase 2]{\includegraphics[width=3 in]{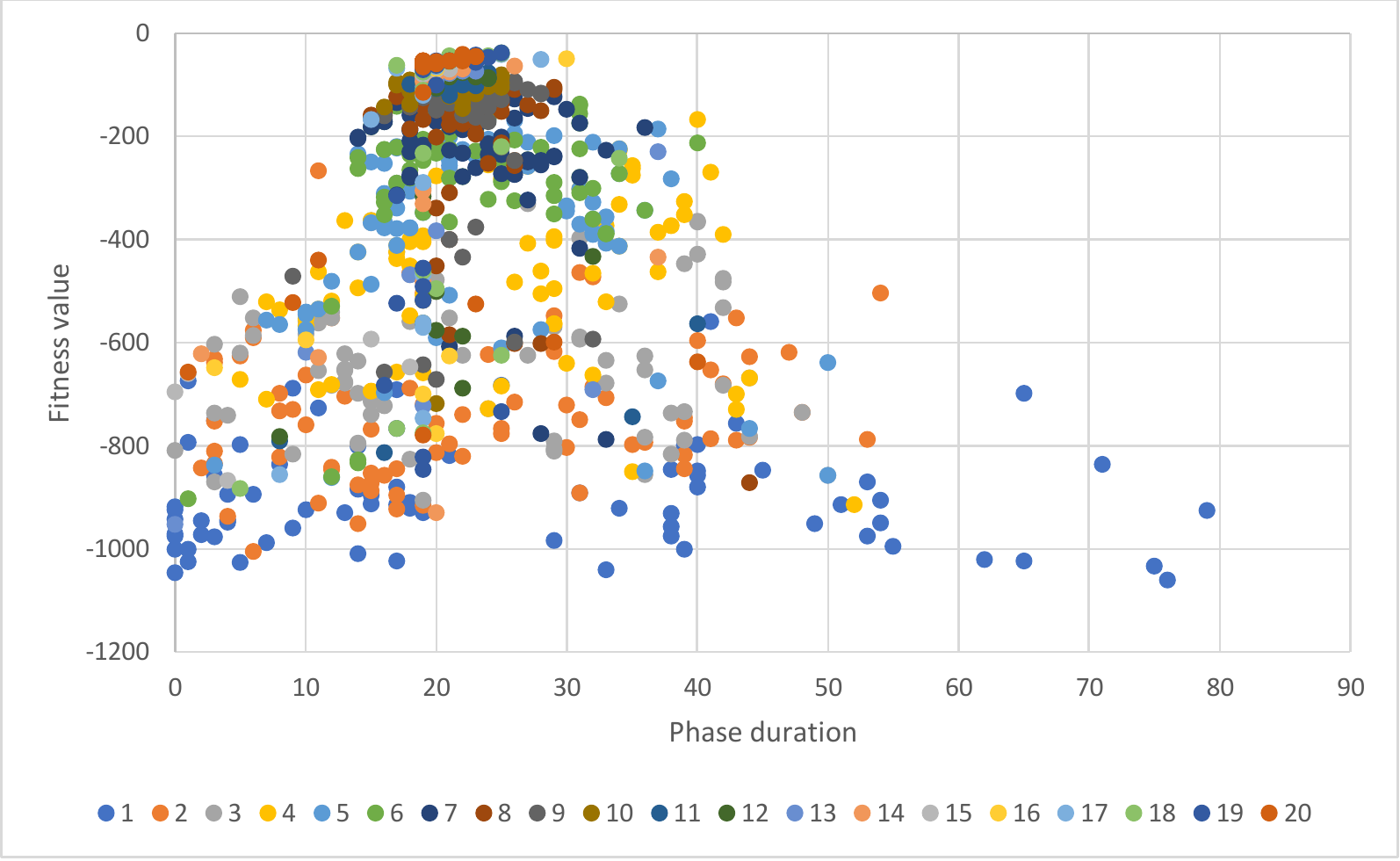}
    \label{fig_p12}}
    \\
    \subfloat[phase 3]{\includegraphics[width=3 in]{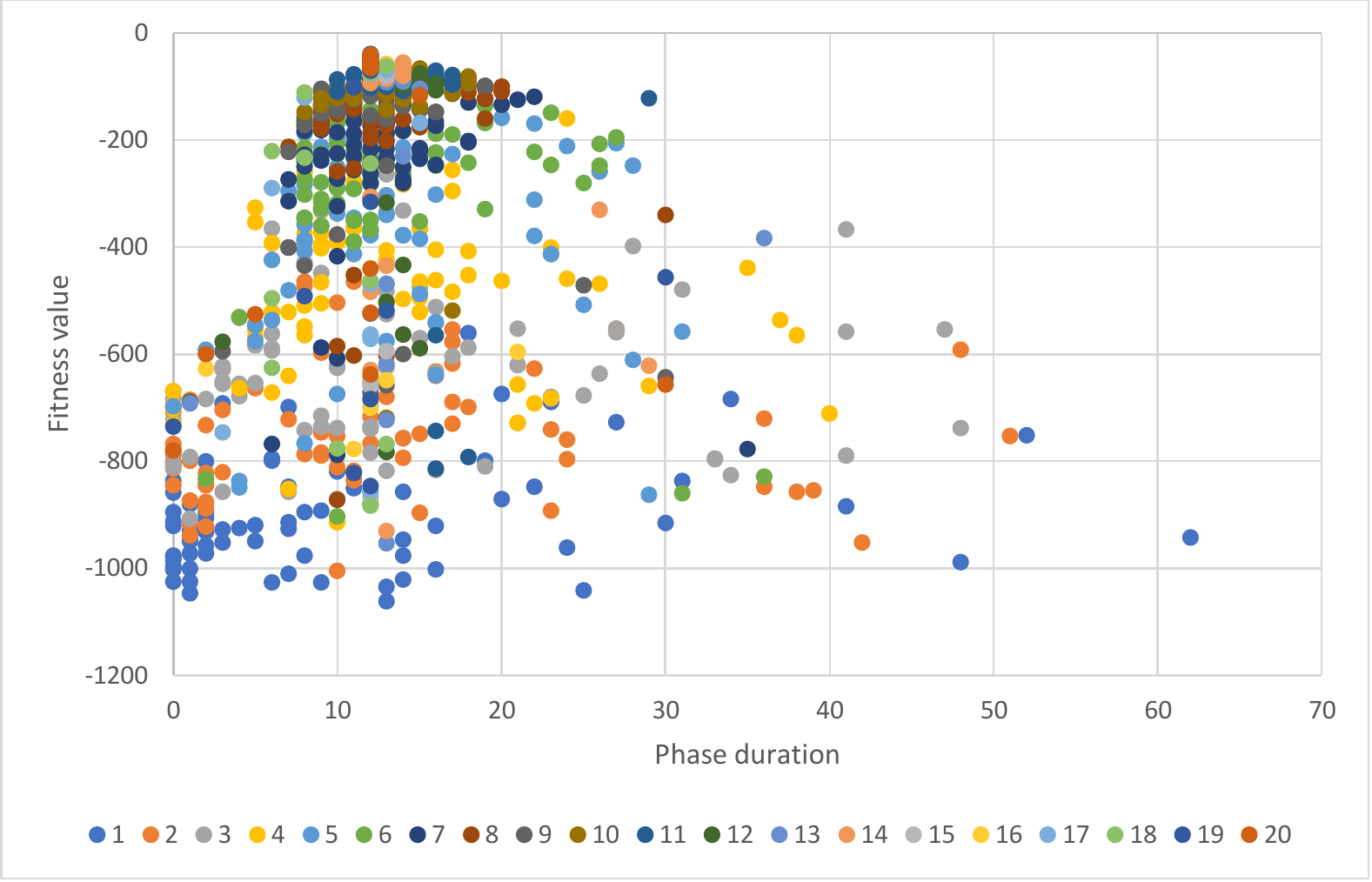}
    \label{fig_p13}}
    \subfloat[phase 4]{\includegraphics[width=3 in]{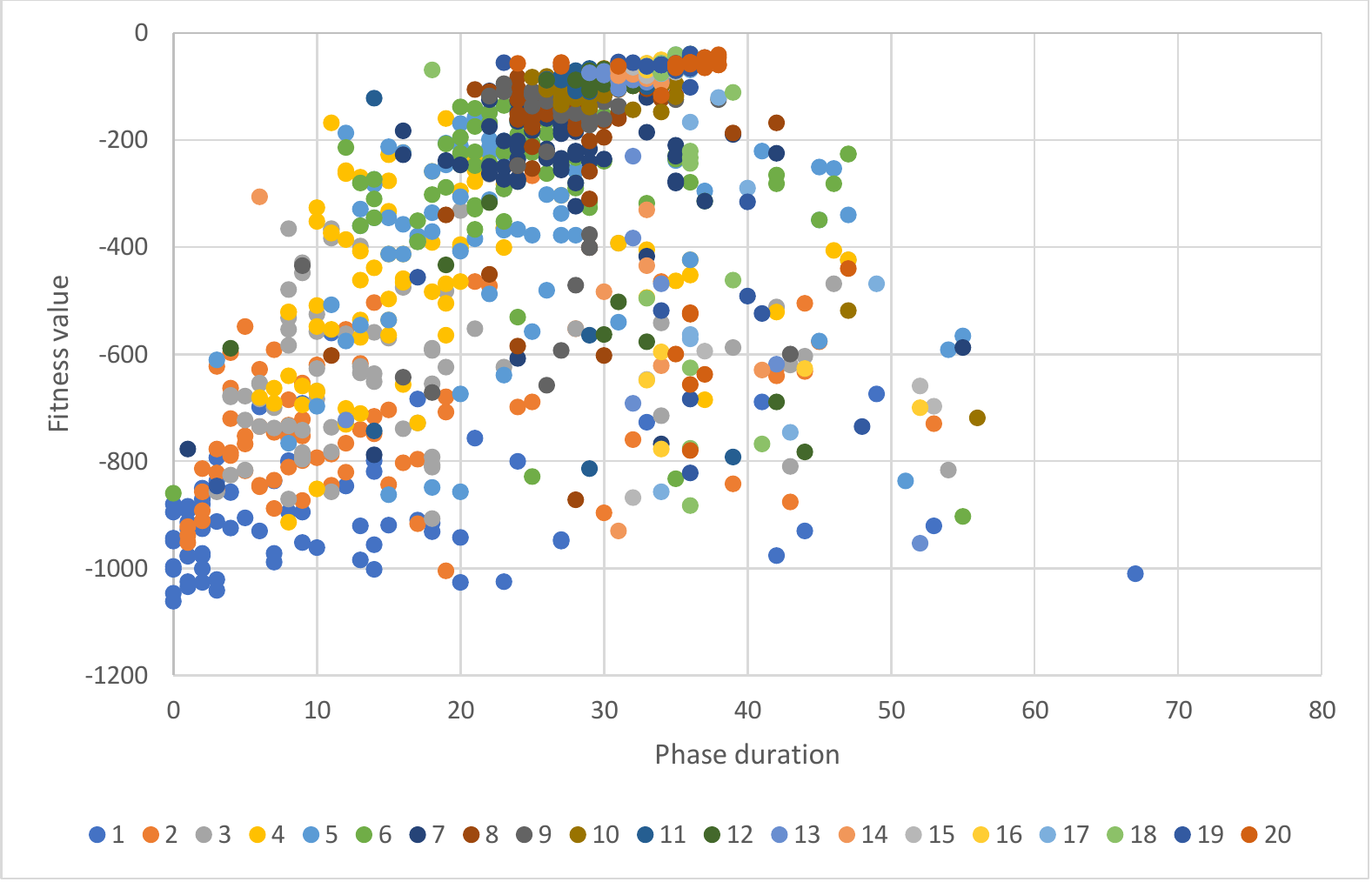}
    \label{fig_p14}}
    \caption{Phase duration convergence in intersection 1}
    \label{fig_result1}
\end{figure*}
\begin{figure*}
    \centering
    \subfloat[phase 1]{\includegraphics[width=3 in]{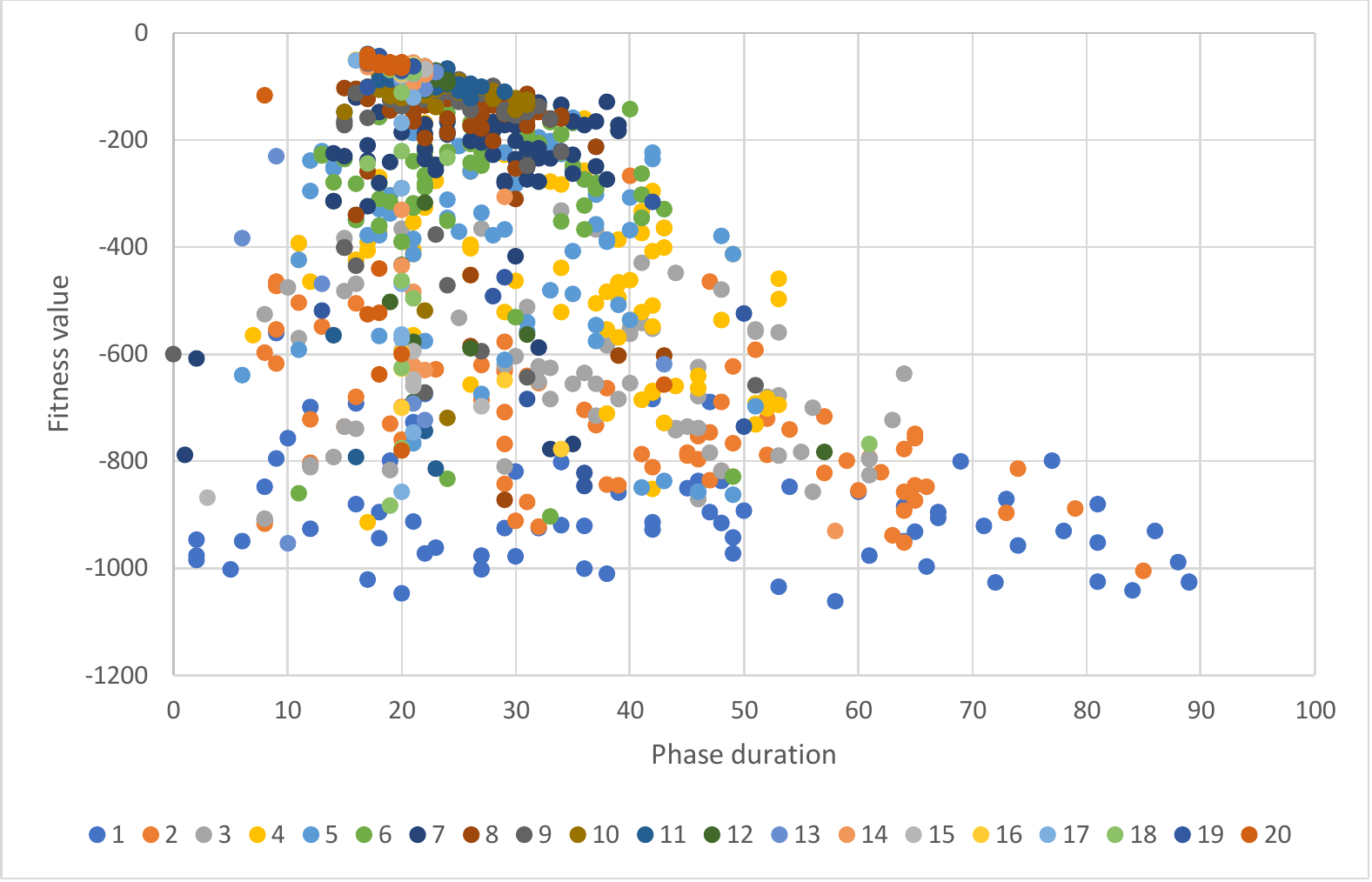}%
    \label{fig_p31}}
    \subfloat[phase 2]{\includegraphics[width=3 in]{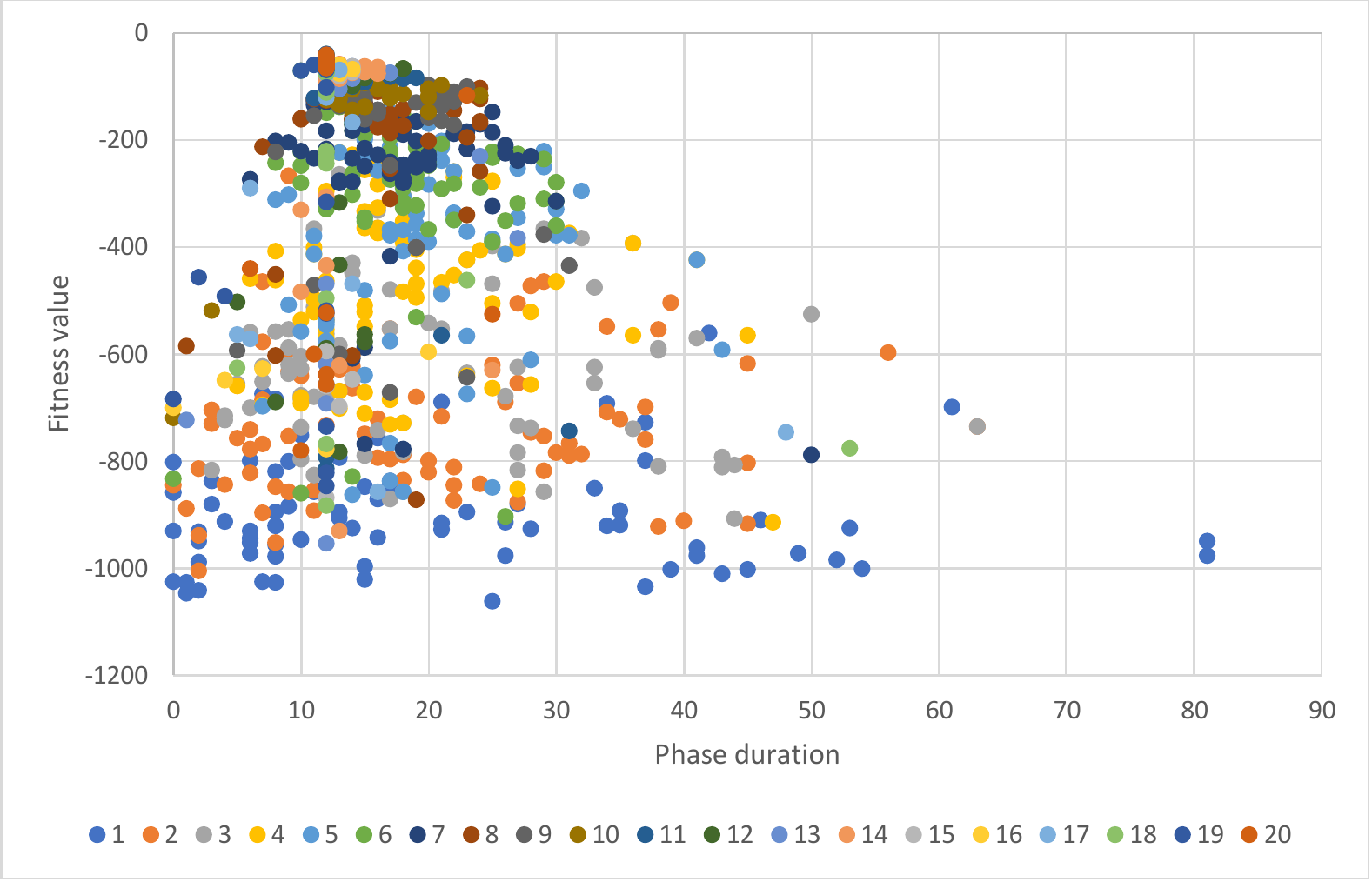}%
    \label{fig_p32}}
    \\
    \subfloat[phase 3]{\includegraphics[width=3 in]{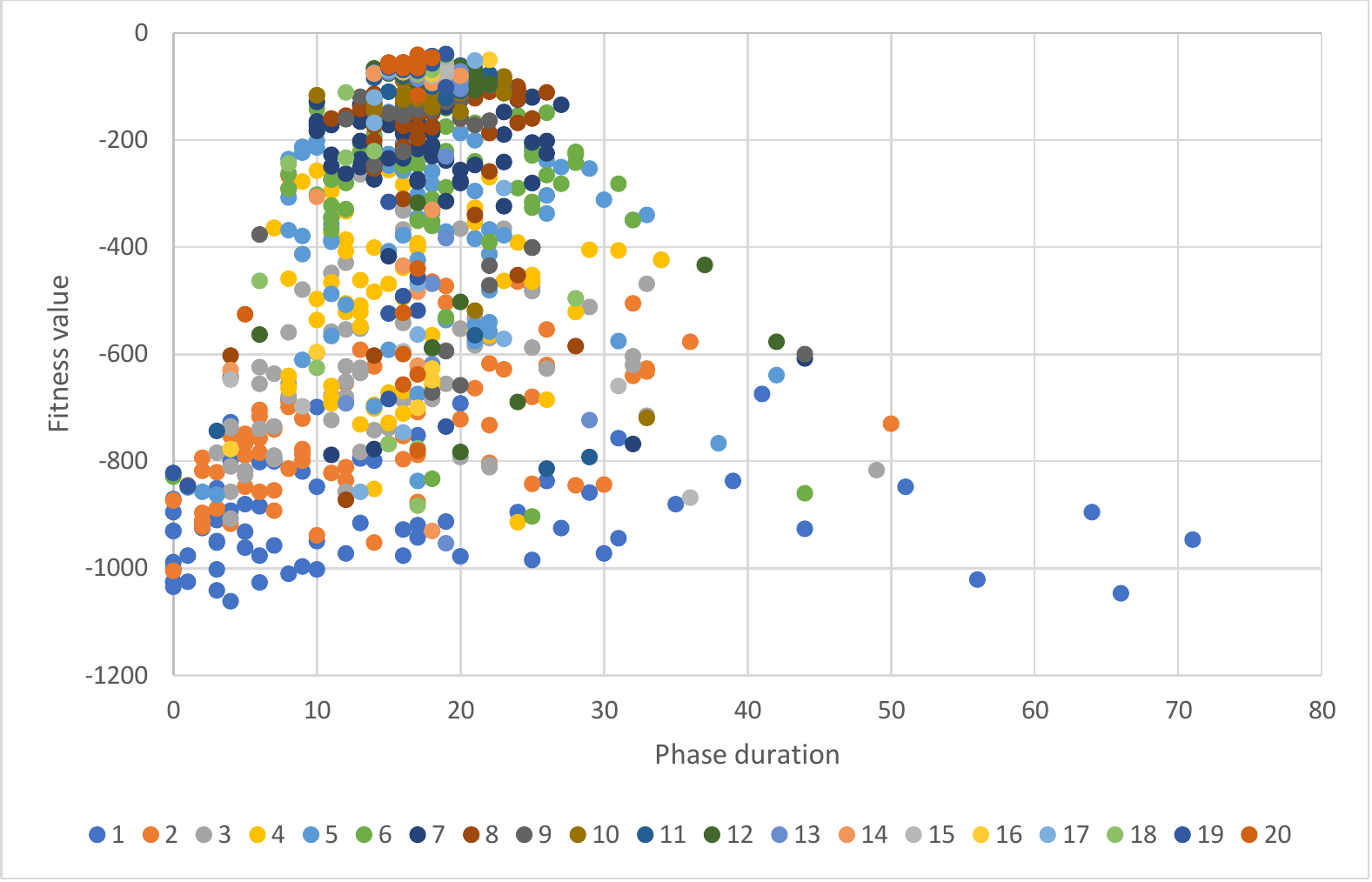}%
    \label{fig_p33}}
    \subfloat[phase 4]{\includegraphics[width=3 in]{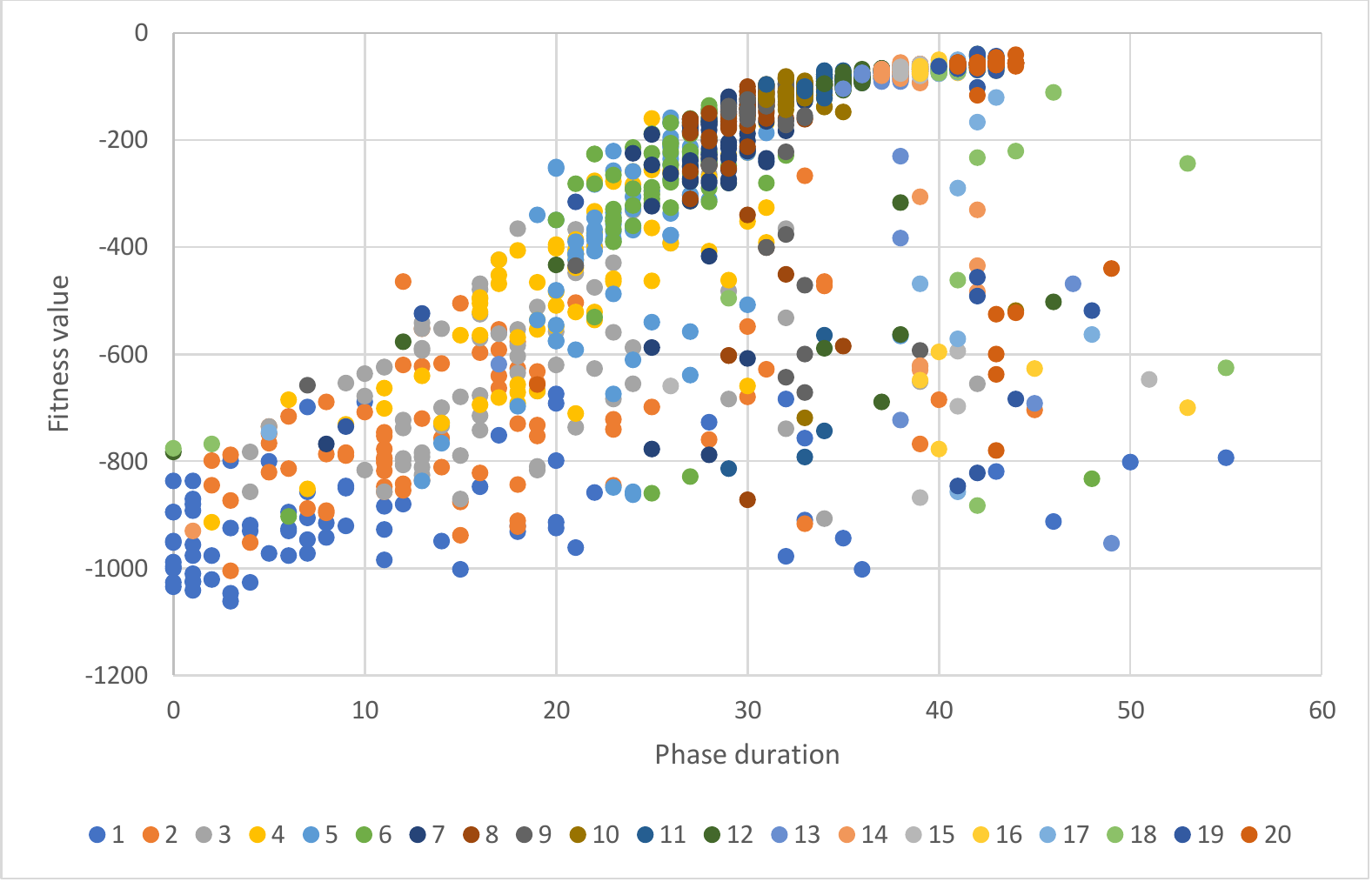}%
    \label{fig_p34}}
    \caption{Phase duration convergence in intersection 3}
    \label{fig_result2}
\end{figure*}

The convergence of each phase duration in intersection 1 and intersection 3 are presented in Figure \ref{fig_result1} and \ref{fig_result2} respectively. The convergence of each phase in other intersections have the same pattern as intersection 1, which can be found in Appendix B. These are the outcome of the GA optimization which starts from an initial population and converge towards the optimal values of each phase duration.

In each sub-figure (such as (a), (b), (c) and (d)), GA started with a big range of phase duration with scattered corresponding fitness values in generation 1. Then after various generations of evolution, the fitness values increase gradually and all phases have reached convergence at the end of GA process in generation 20. \par
There is a significant trend for intersection 3 where the duration of phase 4 is getting longer as the number of generations increases. As shown in Table 1, phase 4 contains the right-turn movement of north and south bound traffic and left-turn movement from east and west bound traffic. The reason for this trend is the high demand from centroid 7 to centroid 3 shown in Table 2 and Figure 3, which leads to high flows using route 1 and route 2 shown in Figure 3. The increasing trend in phase 4 duration in intersection 3 provides more green time to accommodate the traffic flows using route 2.\par
\begin{figure*}
	\centering
	\subfloat[Simulated flow under optimal traffic signal settings without any incident]{\includegraphics[width=3in]{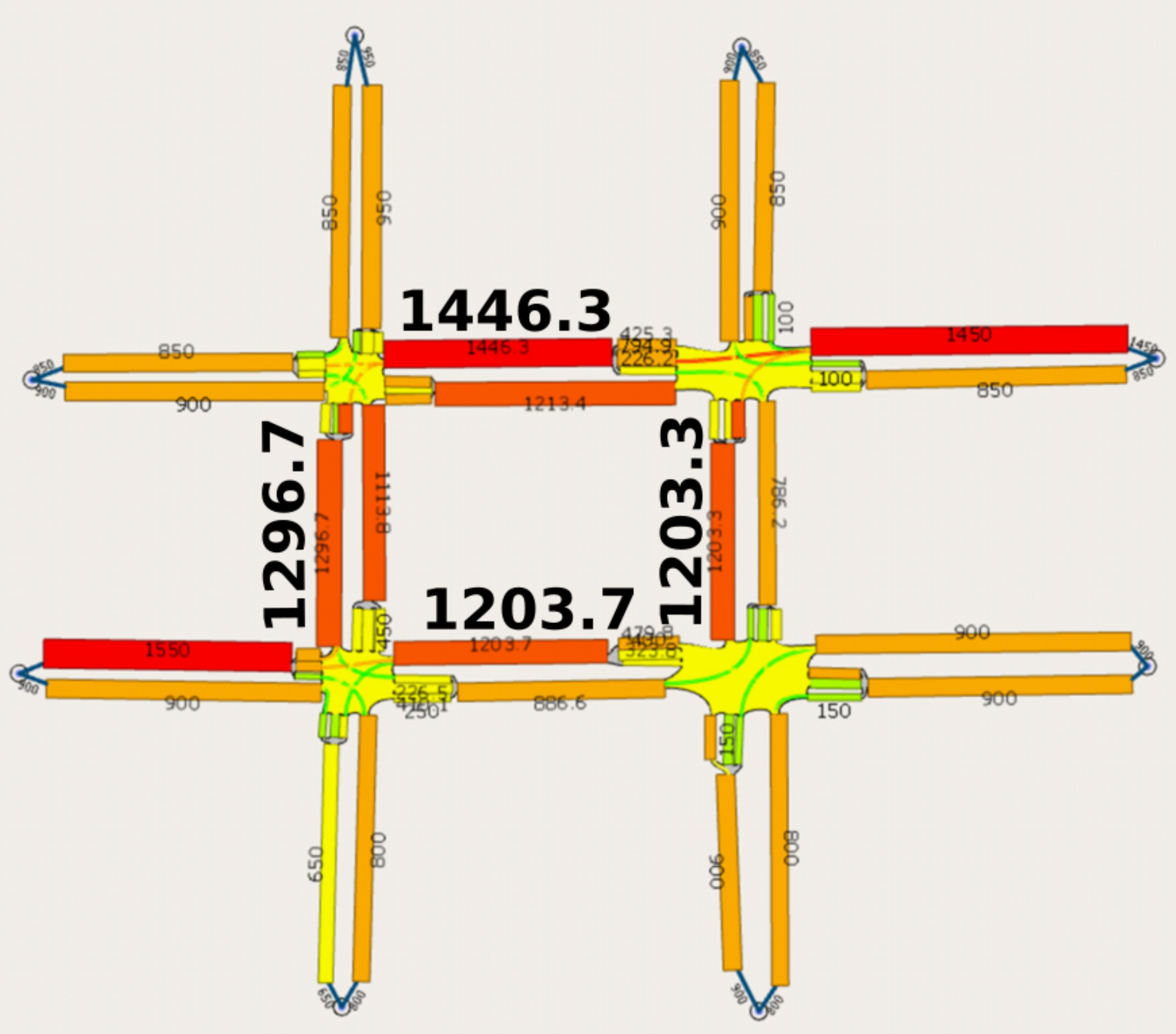}
		\label{fig_result3}}
	\subfloat[Simulated flow with incident]{\includegraphics[width=3in]{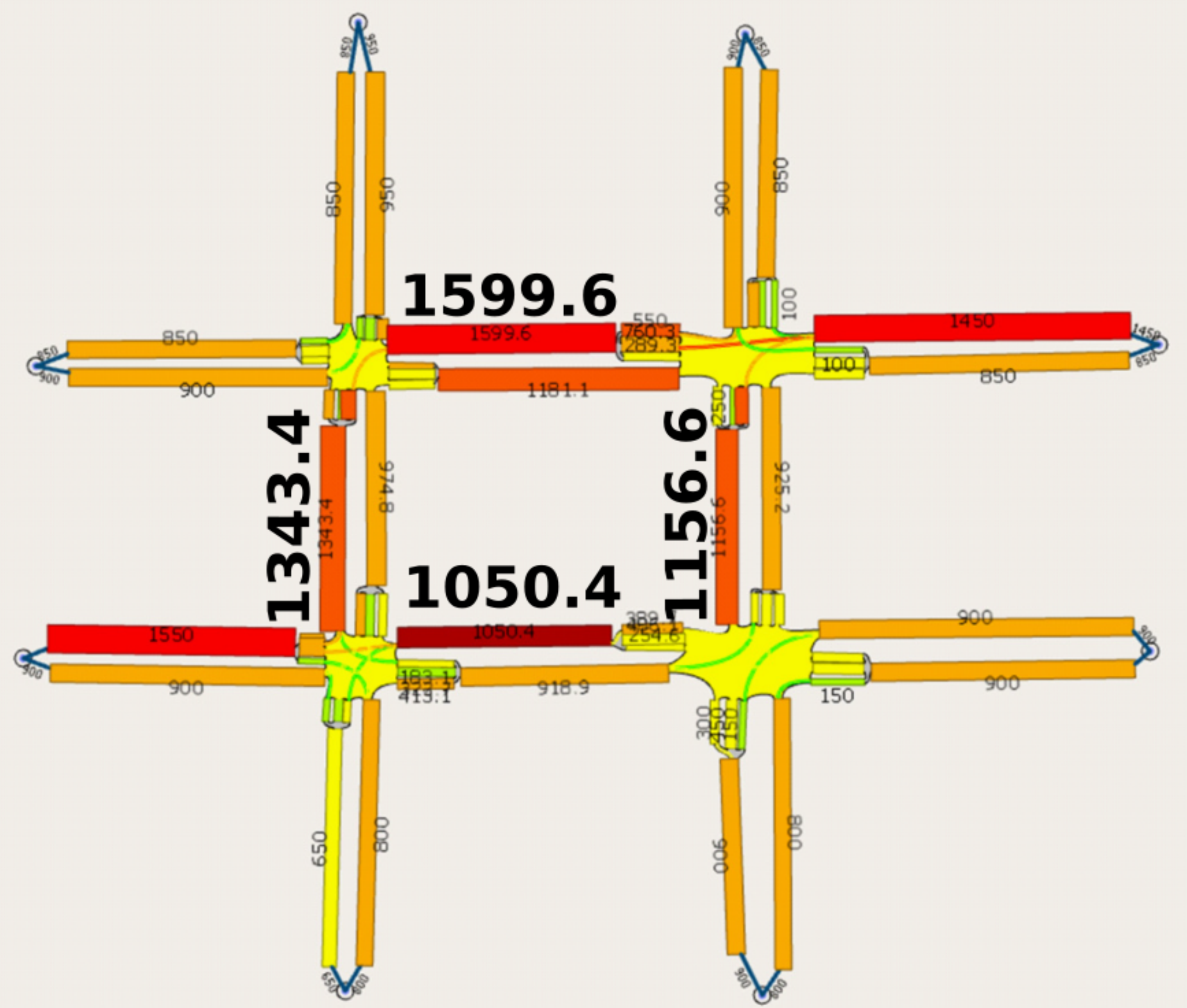}
		\label{fig_result4}}
	\\
	\subfloat[Simulated flow under incident with GA optimized signal control]{\includegraphics[width=3in]{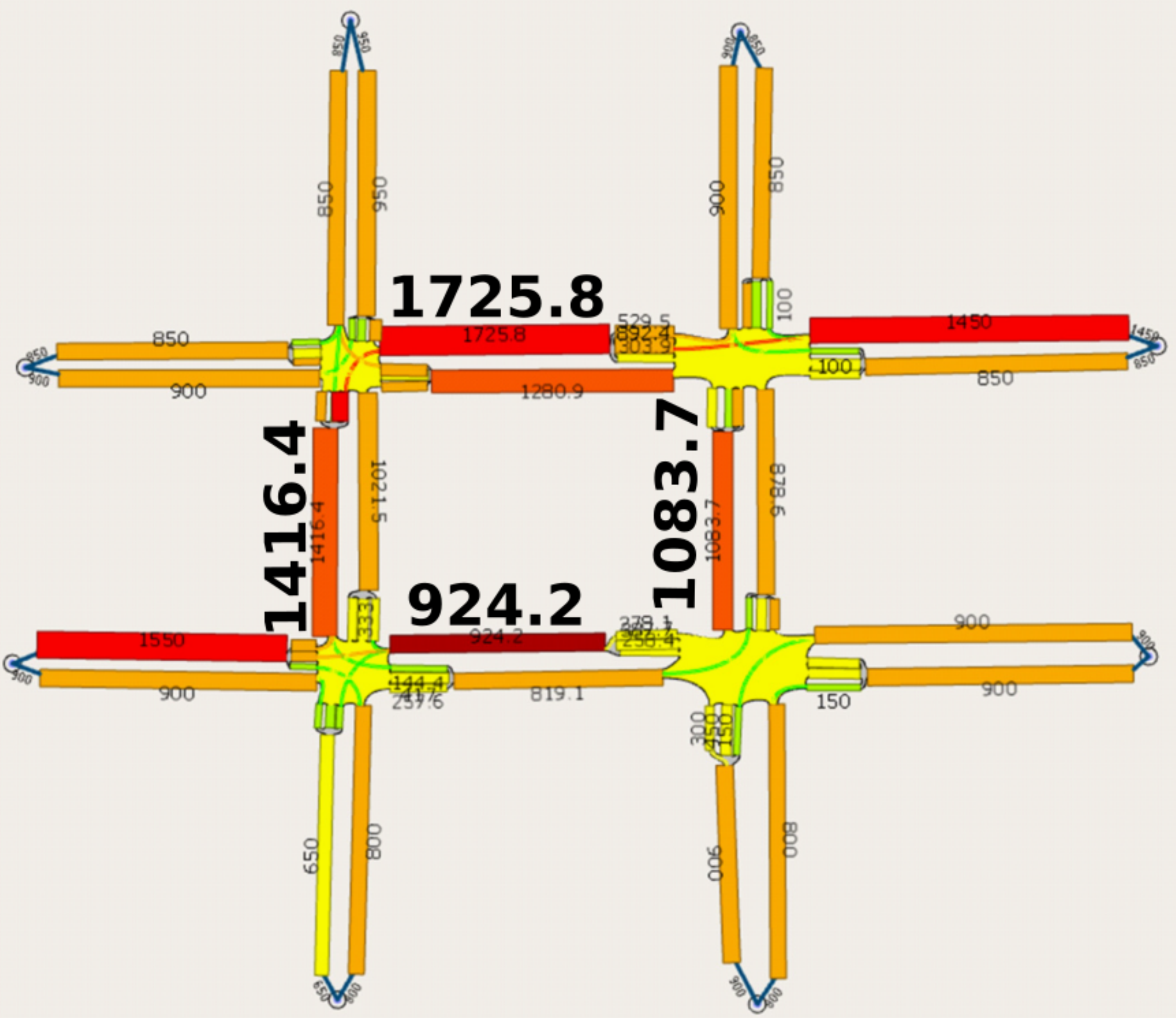}
		\label{fig_result5}}
	\caption{Flow comparison for Scenario 1,2, and 3}
	\label{fig_flowcomparison}
\end{figure*}

 
In addition, the simulated flow using the optimal traffic signal timings generated from GA model is presented in Figure \ref{fig_result3}. The simulated flows prove that the optimal signal timings generated by GA model are aware of the high demand and diverge the flows for two routes. The flows along route 1 and route 2 are around 1200 to 1300 vehicle/hr which are quite even. The reason of evenly split between route 1 and route 2 flows is that both route 1 and route 2 has similar lengths, capacities and turnings in our network.\par

\subsection{Scenario 2: Traffic incident scenario without GA/BGA-ML}
In this scenario, the same signal control plan as scenario 1 is used and the simulated flow are presented in Figure \ref{fig_result4}. The total travel time obtained in this case is 47.37 $vehicle \cdot hour$, which is $111.38\%$ more than the travel time experienced under no incident conditions (22.41 $vehicle \cdot hour$). \par
By comparing Figure \ref{fig_result4} and Figure \ref{fig_result3}, the traffic flow on route 1 increased while the traffic flow on route 2 decreased. This is reasonable, because there is an incident happening during the simulation on route 2. \par

\subsection{Scenario 3: Traffic incident scenario optimization using GA}  
The outcome of proposed GA model is recorded in this case of optimization after the accident has happened. The convergence of each phase in each intersection have the same pattern as in Figure \ref{fig_result1} and can be found in Appendix C.
The final outcome of the GA otpimisation for this scenario is: 
$$Optimal\ phase\ setting\ scenario\ 3=$$
$$\{31,22,13,24,29,23,17,21,30,21,18,21,29,38,9,14\},  $$ 
and the corresponding optimal fitness value is -28.24, which means total travel time is 28.24  $vehicle\cdot hour$ which is $26.02\% $ more than the travel time experienced under no incident condition in scenario 1 (22.41  $vehicle\cdot hour$) and $40.76\%$ lower than scenario 2 (47.37  $vehicle\cdot hour$). This means the GA model is capable of minimizing the total travel time under non-recurrent incidents by almost $40\%$. \par
In addition, the simulated flow using the optimal traffic signal timings generated from GA model is presented in Figure \ref{fig_result5}. The flow on the incident located section dropped comparing to Figure \ref{fig_result3}. On the other hand, by comparing Figure \ref{fig_result4} and Figure \ref{fig_result5}, the allocation of trips along route 1 and route 2 are almost the same, which means that the GA optimized signals maintain the traffic flow and do minor adjustment of signal timing to minimize the total travel time in the network.

\subsection{Scenario 4: Traffic incident scenario optimization using BGA-ML}
\begin{figure*}
	\centering
	\subfloat[Generation 1]{\includegraphics[width=3 in]{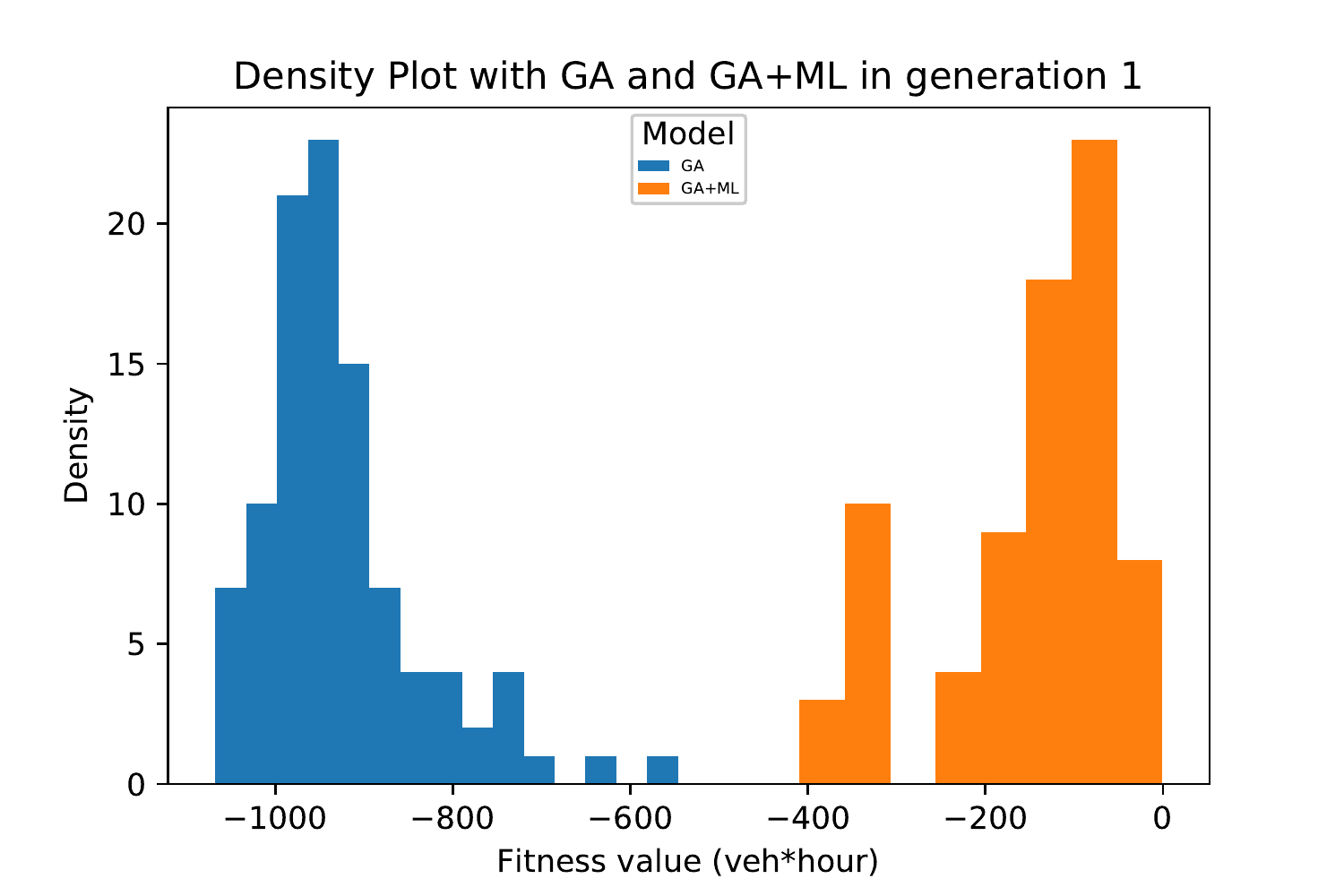}
		\label{fig_gen1}}
	\subfloat[Generation 5]{\includegraphics[width=3 in]{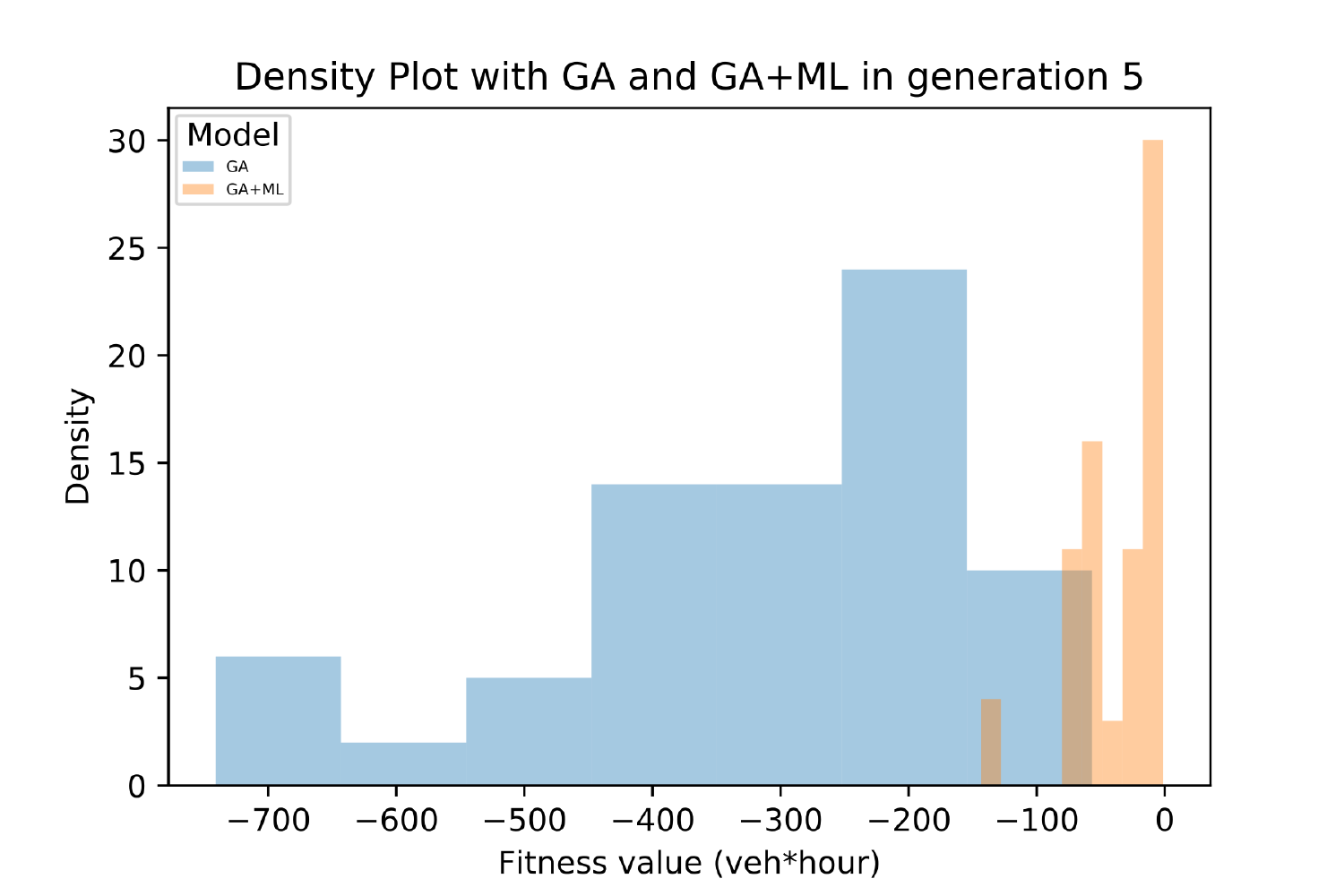}
		\label{fig_gen5}}
	\\
	\subfloat[Generation 10]{\includegraphics[width=3 in]{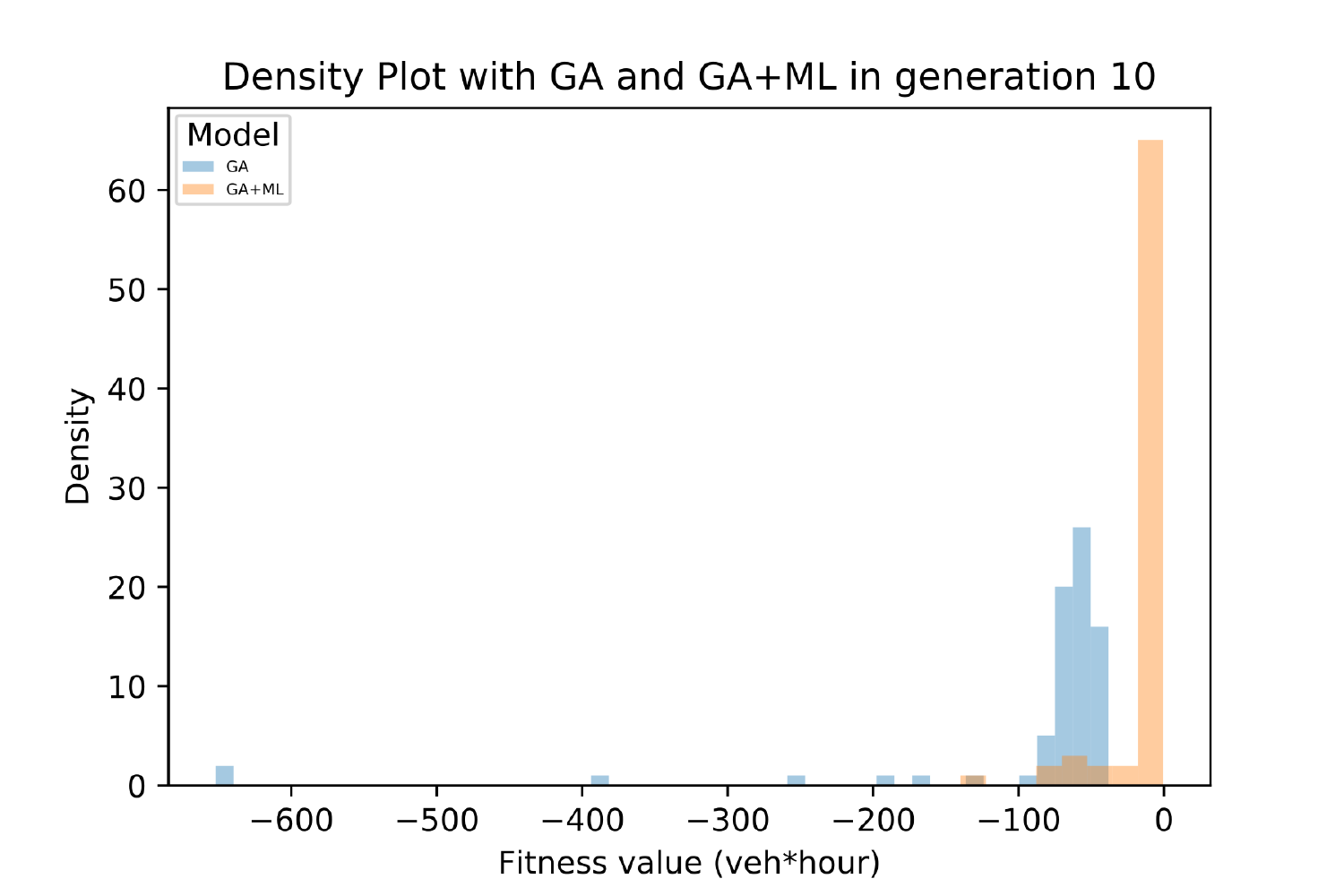}
		\label{fig_gen10}}
	\subfloat[Generation 15]{\includegraphics[width=3 in]{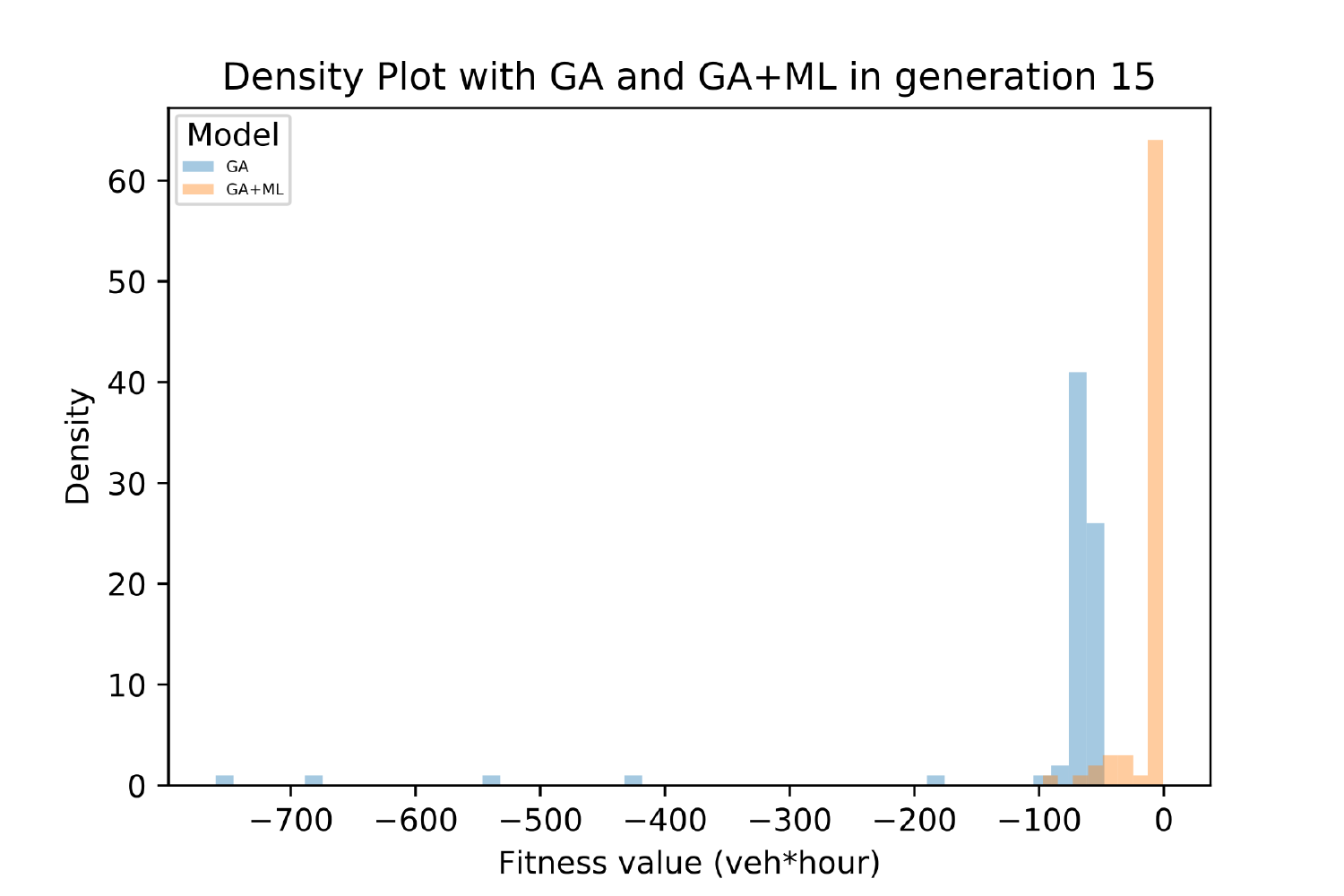}
		\label{fig_gen15}}
	\\
	\subfloat[Generation 20]{\includegraphics[width=3 in]{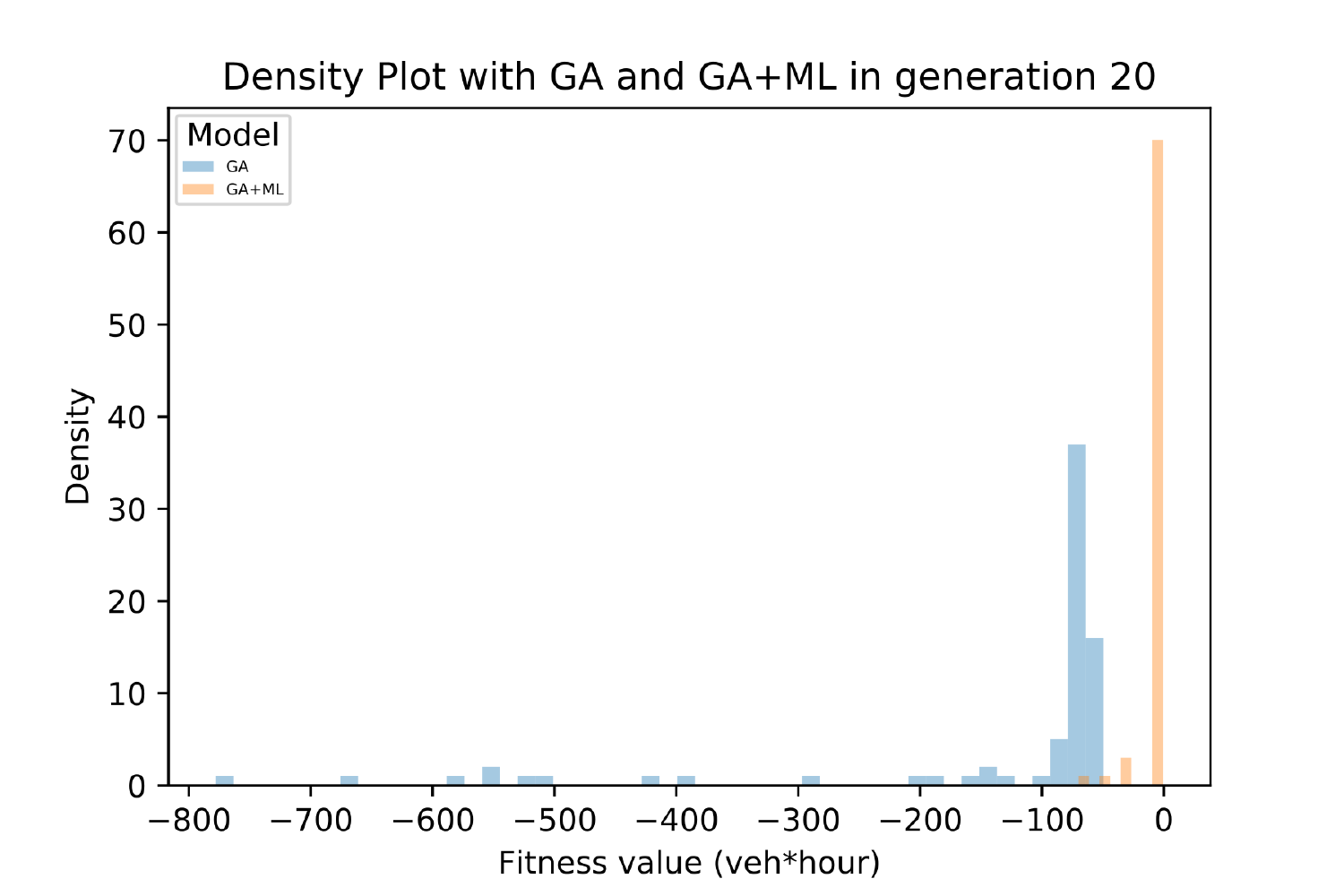}
		\label{fig_gen20}}
	\caption{Fitness value density plot with GA and BGA-ML}
	\label{fig_fitness}
\end{figure*}

In this scenario, we use the best regressor model previously adopted to replace the simulation and to predict the total travel time. The outcome of proposed BGA-ML model is recorded and discussed here.
The BGA-ML model has an unique behavior comparing to GA model, as showcased from the density plots of the fitness values of each model after the first generation and represented in Figure \ref{fig_gen1}. From the beginning, the very first generation of BGA-ML already contains some very good fitness values and has a better and less sparse population coverage of its fitness value than GA.  

As shown in Figure \ref{fig_gen5}, GA's fitness values are slowly increasing while BGA-ML's fitness values are centralizing faster towards a higher and more compact fitness value; even from generation 5, this value is highly possible to be the best fitness value we can get in Generation 20. This indicates a fast convergence of the BGA-ML towards the optimal solution and its efficiency.  

\begin{figure*}
    \centering
    \includegraphics[width=7 in]{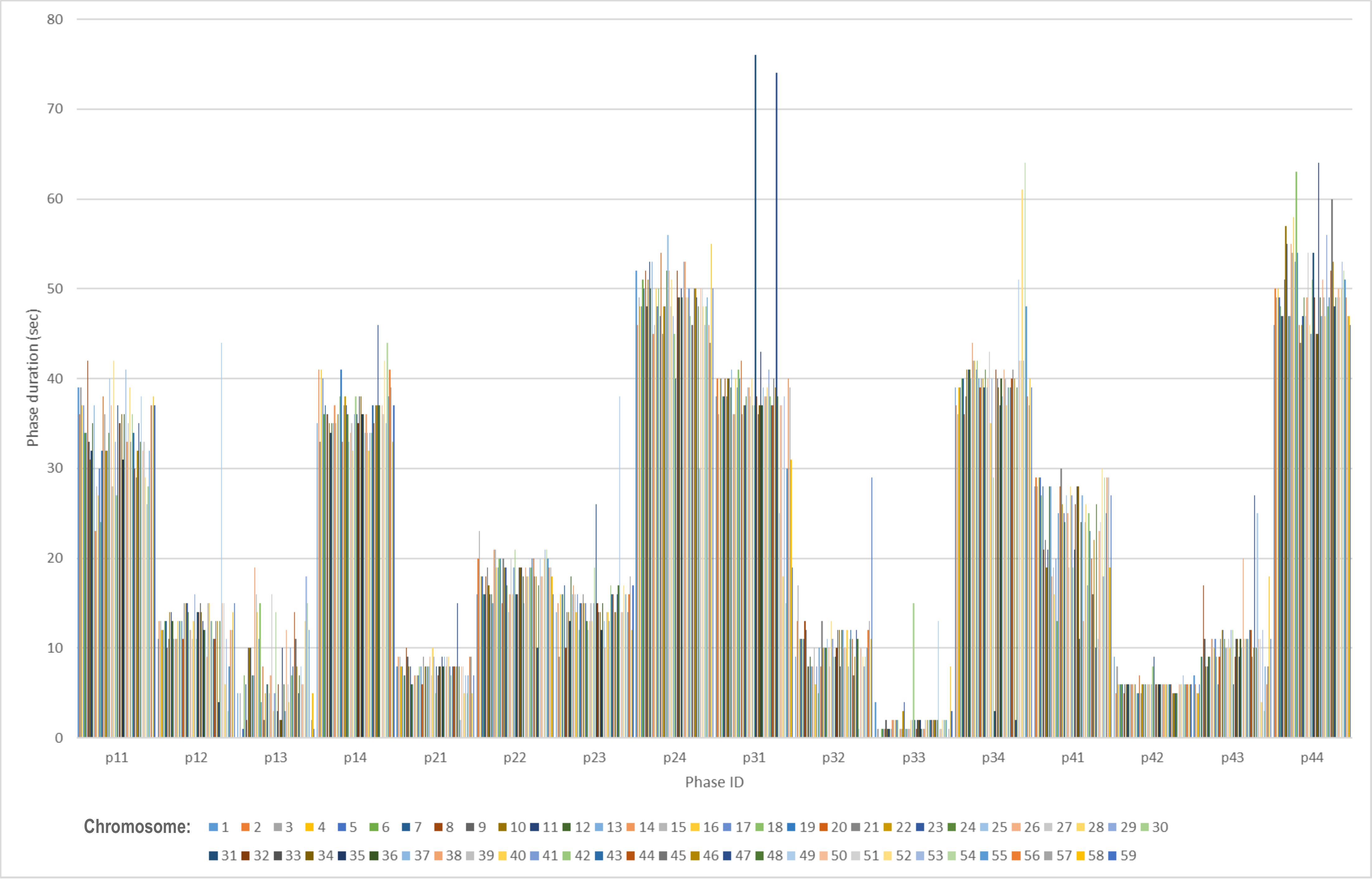}
    \caption{The phase durations in generation 20 of BGA-ML}
    \label{fig_phase_gen20_GAML}
\end{figure*}

As further shown in Figures \ref{fig_gen10} and \ref{fig_gen15}, GA is catching up slowly to its optimal fitness value while BGA-ML stays almost the same, meaning it converged earlier than the simple GA. Finally we can take a closer look at the last generation 20 when BGA-ML has a lot of chromosomes with the same greatest fitness value (very tight density plot). This is a pure indication of a fast converge of the BGA-ML compared to regular GA. However, it is hard to decide which one is the best chromosome. Therefore, we plot all the chromosomes which contains the phase durations for all signalized intersections in Figure \ref{fig_phase_gen20_GAML}. 

 As we can observe from this figure \ref{fig_phase_gen20_GAML}, 59 out of 75 chromosomes have the best fitness value. The BGA-ML can capture the best phase durations but with a lot of prediction noise. Therefore, we treat the mean values of all phase durations as the optimal phase duration to remove the noise. 

The final outcome of BGA-ML model for this scenario is: 
$$Optimal\ phase\ setting\ scenario\ 4=$$
$$\{33,	13,	8,	36,	9,	18,	15,	48,	38,	9,	4,	39,	23,	7,	10,	50\},  $$ 
and the corresponding optimal fitness value is -16.06, which means total travel time is 16.06  $vehicle\cdot hour$. This is even lower than the original traffic condition without any accident (by $25\%$) which indicates that the BGA-ML not only is better than regular GA optimizer (by $43\%$), but has a higher capacity of optimizing better the traffic phase durations under incidents when compared even to the \say{no-accident} conditions. However, we would like to mention that the accuracy of the total travel time predicted by BGA-ML is highly affected by the embedded ML model. In this case, our ML model was trained within 9,743 runs mentioned previously in the \cref{feature_generation}, under one-incident conditions blocking a lane out of two in the road section. \par
Nevertheless with limited number of runs (close to 10,000), our BGA-ML model still shows remarkable good performance as we will discuss in the next section (Sec. \ref{sec_discussion}). Its performance can be further improved with several hundred thousands or more runs to better train the ML regressor; however this will require extensive computational power due to the large number of possibilities and hyper parameters to be tuned.\par
In terms of computational time, BGA-ML takes only 11 minutes to complete while GA needs about 8 hours. BGA-ML is much faster for any real-time applications in term of computational time. We still believe there is room for improvement in both running time and accuracy by expanding the modelling spaces and find faster convergence methods.

\section{Discussion}\label{sec_discussion}
In scenario 1, we simulated the daily normal traffic under normal traffic control plan. The GA model was applied to get the optimal traffic control plan. Then in scenario 2, a traffic incident was created in the network, and no more action was taken to respond to the traffic incident. The total travel time in scenario 2 increased by 111.38\% comparing to the total travel time in scenario 1. We then simulated the case that we took instant response to the traffic incident and applied the GA model to re-estimate the optimal traffic control plan. The total travel time in scenario 3 only increased by 26.02\% 1comparing to the total travel time in scenario 1. 
By comparing scenario 2 and 3, the proposed GA model is able to adjust the signal timings to minimize the total travel time. In our case study, a 40.76\% of total travel time saving is achieved in the network.
Lastly we evaluated the BGA-ML model in the scenario 4 and revealed is lower by $43\%$ than the regular GA and a further $25\%$ compared to the no accident conditions, revealing his power of best optimizing the phase durations under any type of conditions. After a series of hyper-parameter tuning, we were able to shorten the computational time from 8 hours to 11 minutes, which proved to be a significant time saving procedure.\par
Observe that the behavior of BGA-ML differs from the original GA. First, the BGA-ML model converges much faster than GA. Although this might be caused by the training process and the range of training data which is limited; therefore the prediction by the ML may not cover the whole space of possible fitness values. It's safe to infer that ML boosted the BGA-ML's converging process considerably as it converges at about $10^{th}$ generation while GA needs at least 15-20 generations to converge.\par
Secondly, also as a reason of the fast convergence, the prediction of the ML on the input data is very robust and fuzzy. In other words, a lot of chromosomes have the same fitness value in our trial. Although this is highly dependent on the training of the ML model, we believe is unavoidable due to the lack of real-life training data.\par
Thirdly, some ML predictions are out of the feasible range. In our case, the prediction is the total travel time, but we observe some negative values predicted by the ML as shown in Figure \ref{fig_y_xgbt_mse}. It's unrealistic to have the travel time to be negative, so we treat this as an over-fitting and we replace the prediction with a very large travel time by applying a heuristic rule.\par
Lastly, the determination of the final output in BGA-ML is different than GA. Because of the nature of the ML we discovered, is more difficult to determine the optimal chromosome. After we observed all the chromosomes from the BGA-ML (in Figure \ref{fig_phase_gen20_GAML}), most of the chromosomes contains similar values except some noise which means we can use some statistical skills such as averaging all the chromosome to determine the optimum. We calculate the average and the standard deviation of the phase duration for each phase which are shown in Table\ref{tab_statistics_20GEN} with a relatively small standard deviation. We use the mean value as our optimum and we make a final adjustment by making all the phase duration into integers. 
\begin{table}[!t]
    \renewcommand{\arraystretch}{1.3}
    \centering
    \caption{The statistics of the phase duration in the last generation}
    \begin{tabular}{|c|c|c|}
    \hline
    PhaseID     & Average phase duration & Standard deviation \\
    \hline
    p11&	33.830&	4.231\\
    \hline
    p12&	12.85&	4.77\\
    \hline
    p13&	7.39&	4.70\\
    \hline
    p14&	35.93&	5.46\\
    \hline
    p21&	7.78&	1.66\\
    \hline
    p22&	18.17&	2.12\\
    \hline
    p23&	15.27&	3.85\\
    \hline
    p24&	48.78&	3.75\\
    \hline
    p31&	38.20&	8.69\\
    \hline
    p32&	10.35&	3.37\\
    \hline
    p33&	2.08&	2.53\\
    \hline
    p34&	39.35&	8.52\\
    \hline
    p41&	23.34&	5.25\\
    \hline
    p42&	6.03&	0.80\\
    \hline
    p43&	10.42&	4.14\\
    \hline
    p44&	50.20&	4.26\\
    \hline
      
    \end{tabular}
    
    \label{tab_statistics_20GEN}
\end{table}

\section{Conclusion}
In this paper, a boosted GA method using ML is developed to mitigate the impact of non-recurrent traffic incidents under a case study network. The proposed BGA-ML model is transformed from a standard GA model by adapting the key components to traffic signal timing optimization. These components consist initialization, fitness function calculation, crossover, mutation and so on. In prior to traffic simulation with traffic incidents, the key parameters of GA such as population size and a maximum number of generations are sampled and the best setting of these key parameters are worked out by choosing the best performance with relatively short computation time. Then we boost the original GA by embedding the ML to make a new BGA-ML model. The ML model is specially trained to replace the Aimsun simulation model in order to predict the total travel time. 

At last, as a proof of concept, an experiment is designed to simulate the cases whether TMC takes action to revise traffic control plans after the appearance of an incident or not and compare the performance of applying GA and BGA-ML's in this urgent situation. The experiment results show improvement of total travel time if the TMC uses the proposed BGA-ML model to re-optimize the traffic control plan under the incident condition comparing to taking no action at all. The saving in total travel time is by $43\%$ than the regular GA and a further $25\%$ compared to the no accident conditions. BGA-ML seems even more promising than using a simple GA, and has a lower computational time with great potential of completing the task. \par
Due to a all-at-once feature selection, the ML model in this paper predicts very fuzzy outcome. Future work can be further developed to improve the way the best ML regressor learns from the traffic network and the incident response. This will mean exploring more reinforcement learning techniques online as new traffic data becomes available. 

\section{Author contribution}
The authors confirm contribution to the paper as follows: study conception, design, and validation: Dr. Mao and Dr. Mihaita; data science and model performance: Dr. Mao; draft manuscript preparation: all authors. All authors reviewed the results and approved the final version of the manuscript. The source code of this journal work can be found online at: \url{https://github.com/ft912678/BGA_ML} and is meant to be used for research and development purposes only with correct bibliographic reference to the current paper.

\appendices
\section{}
This section contains the graphs of the phase duraton convergence over different number of generations (from 1 to 59) and different size of the population (25,50,75, and 100).
\begin{figure}[H]
	\centering
	\subfloat[phase 1]{\includegraphics[width=1.5 in]{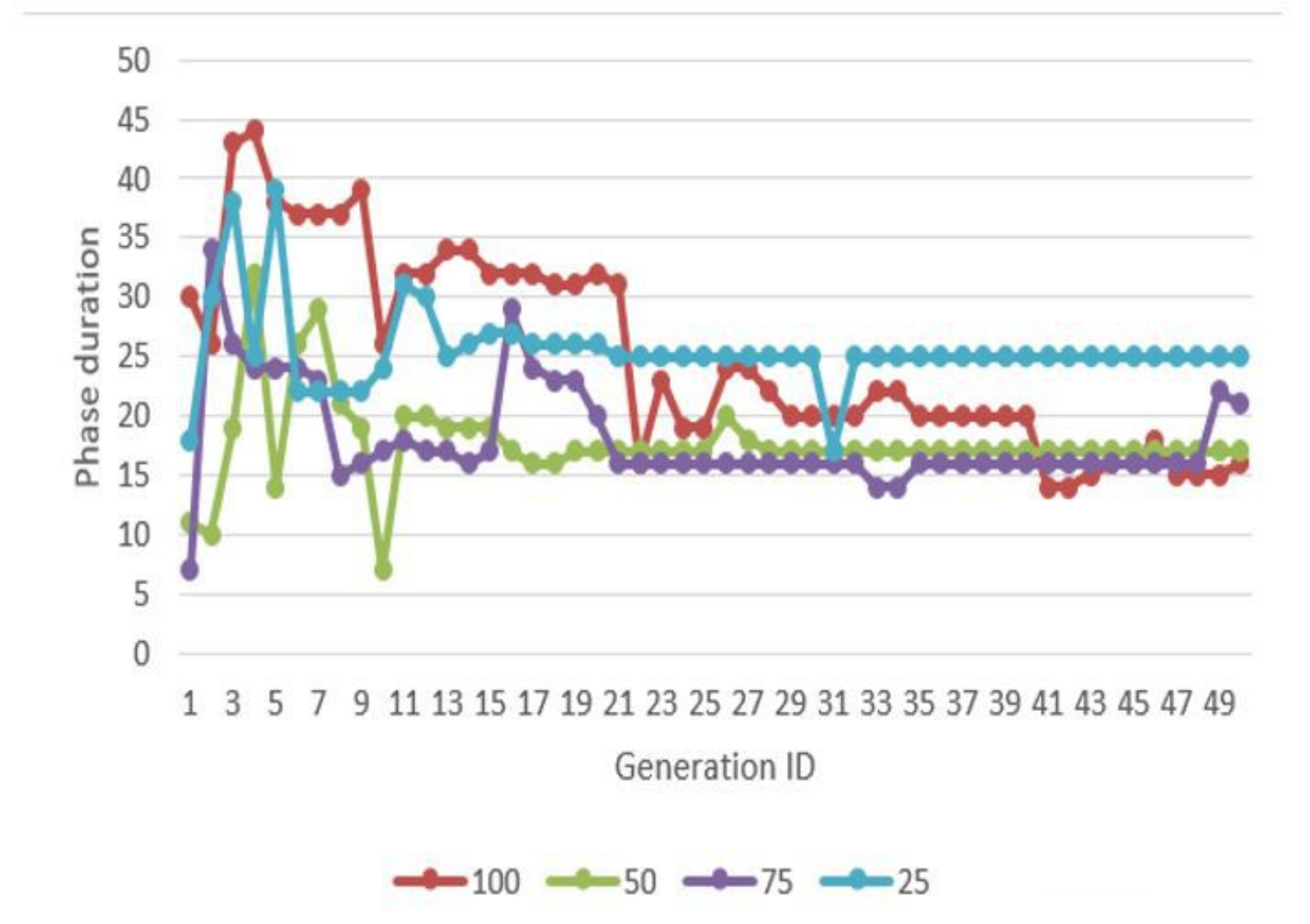}
		\label{fig_a_p21}}
	\subfloat[phase 2]{\includegraphics[width=1.5 in]{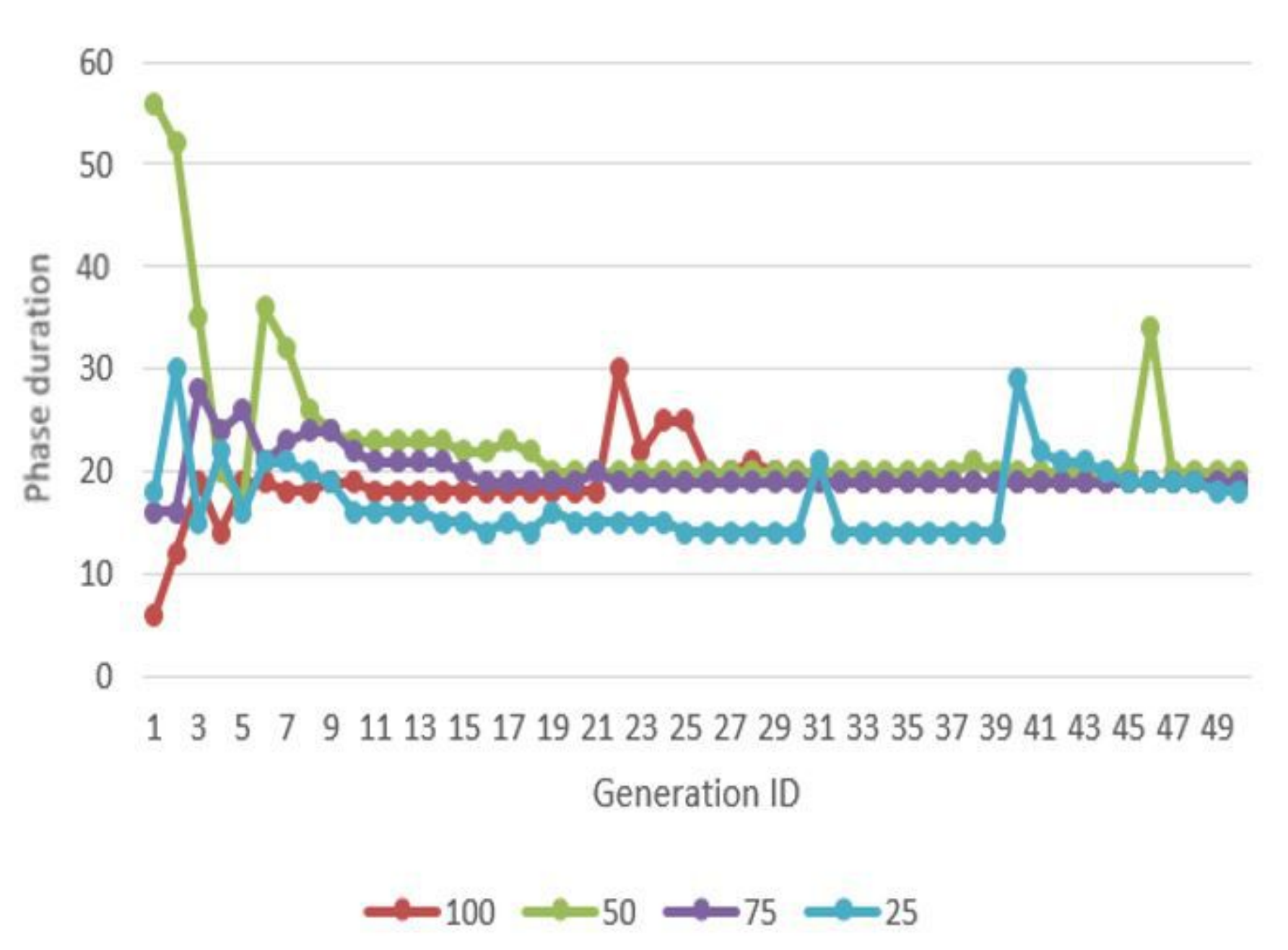}
		\label{fig_a_p22}}
	\\
	\subfloat[phase 3]{\includegraphics[width=1.5 in]{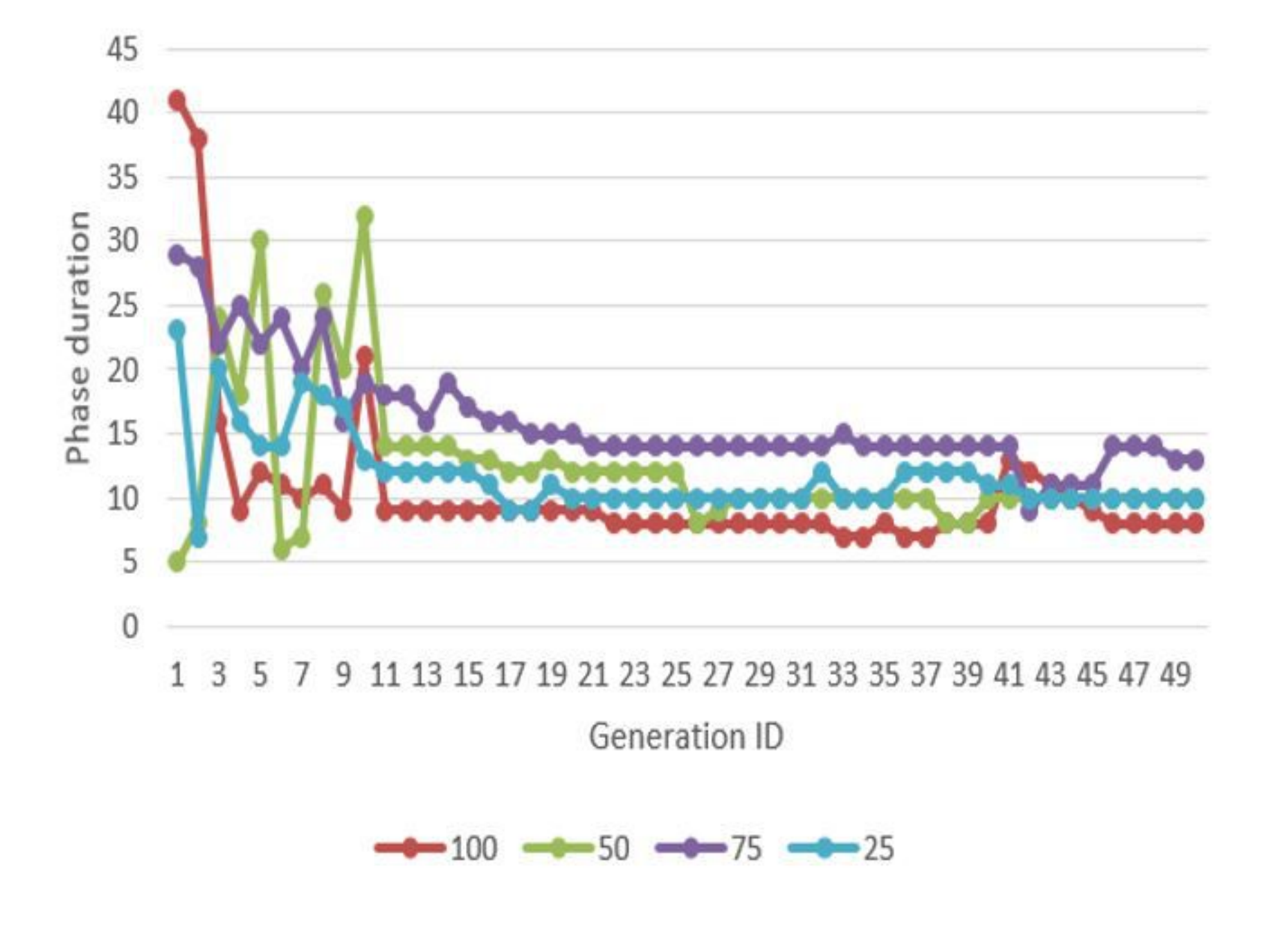}
		\label{fig_a_p23}}
	\subfloat[phase 4]{\includegraphics[width=1.5 in]{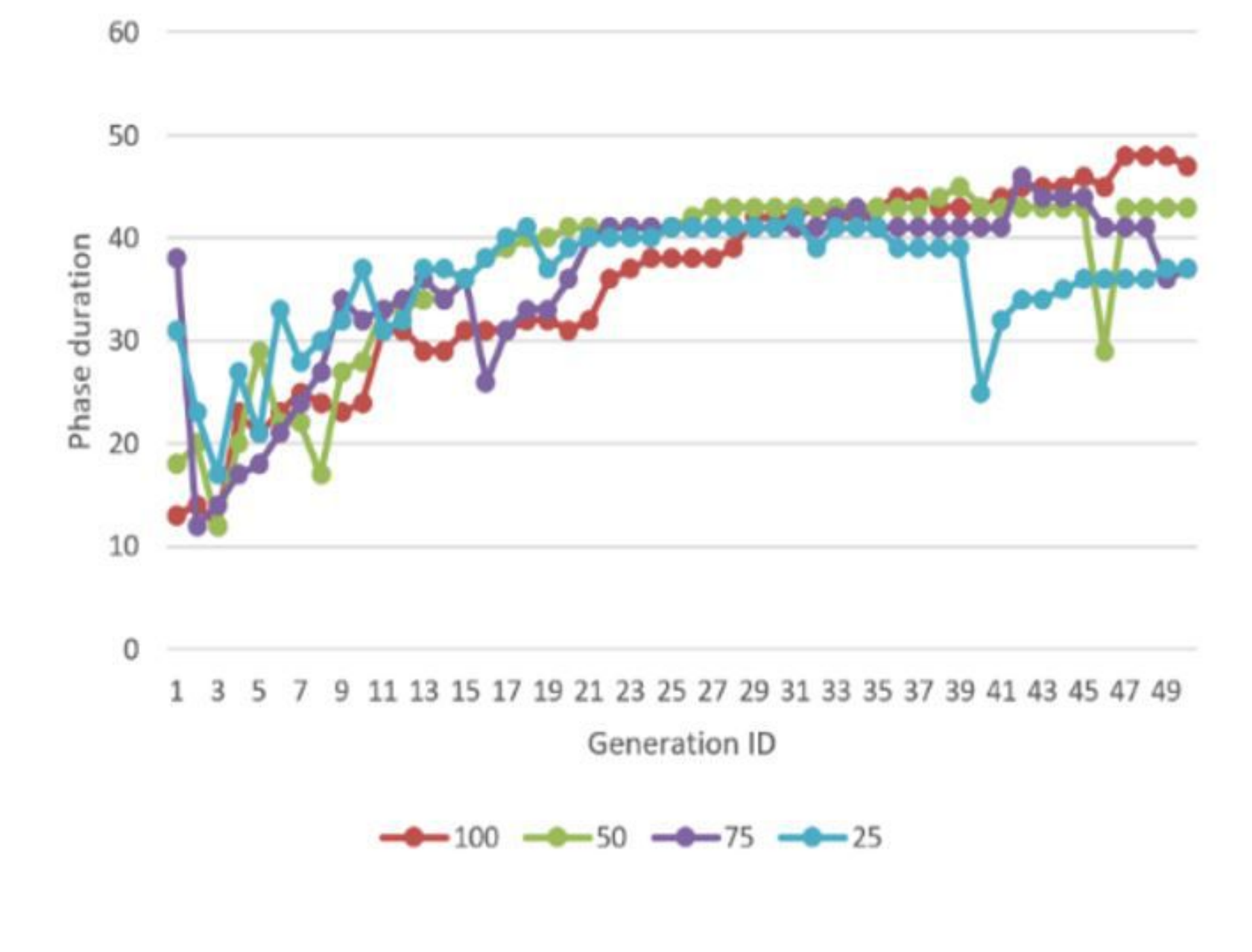}
		\label{fig_a_p24}}
	\caption{Phase duration convergence in intersection 2}
	\label{Annex_result1}
\end{figure}

\begin{figure}[H]
	\centering
	\subfloat[phase 1]{\includegraphics[width=1.5 in]{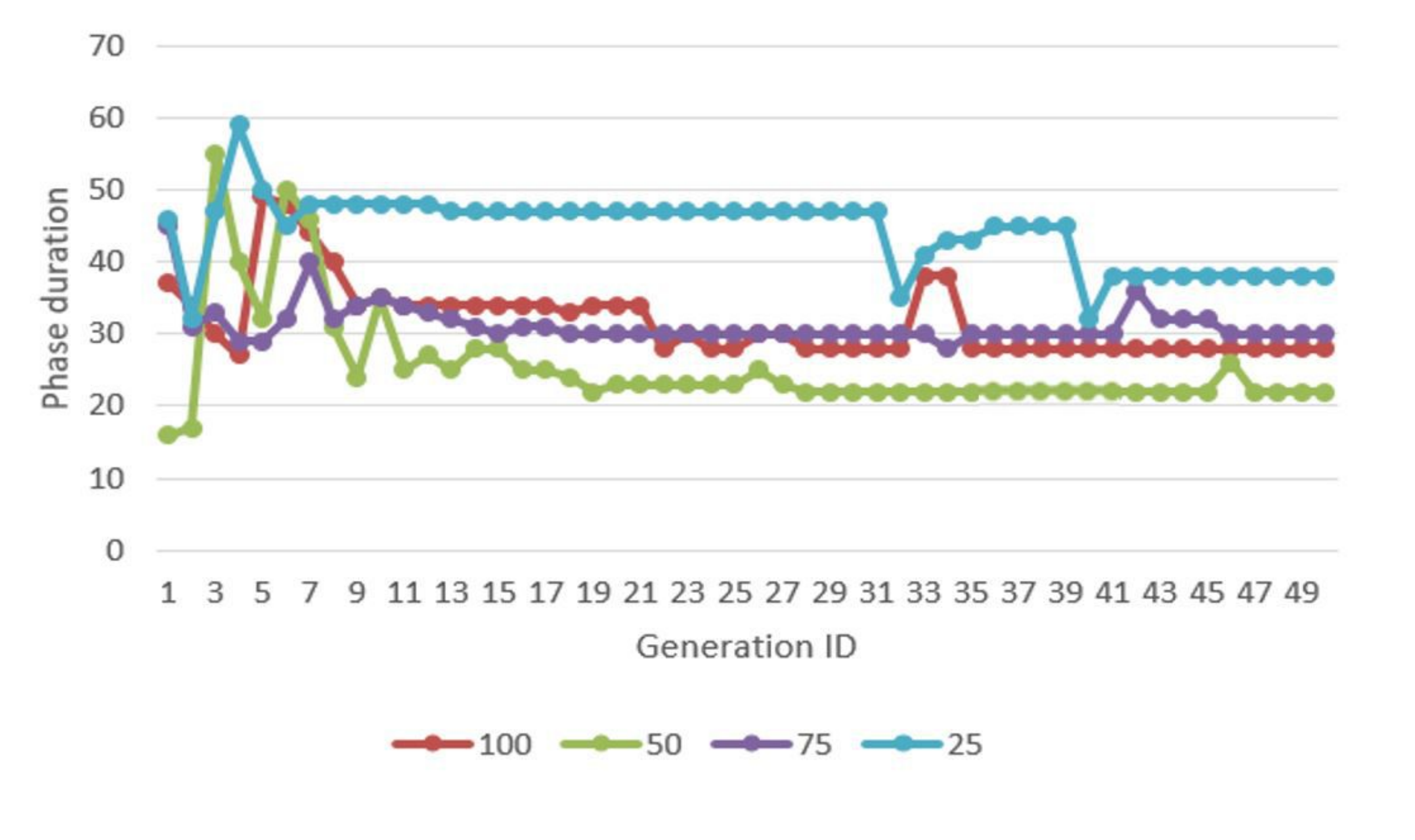}
		\label{fig_a_p41}}
	\subfloat[phase 2]{\includegraphics[width=1.5 in]{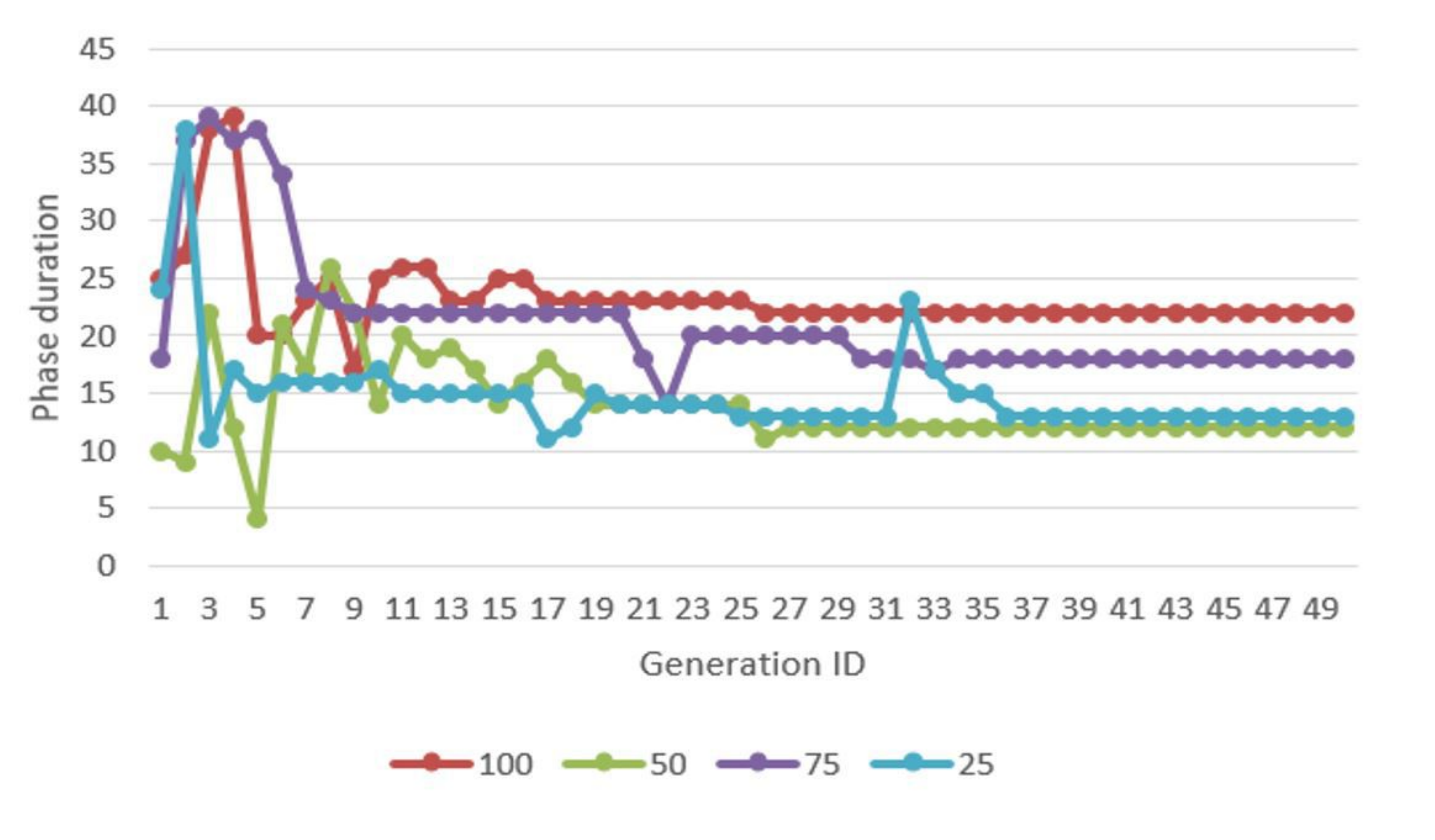}
		\label{fig_a_p42}}
	\\
	\subfloat[phase 3]{\includegraphics[width=1.5 in]{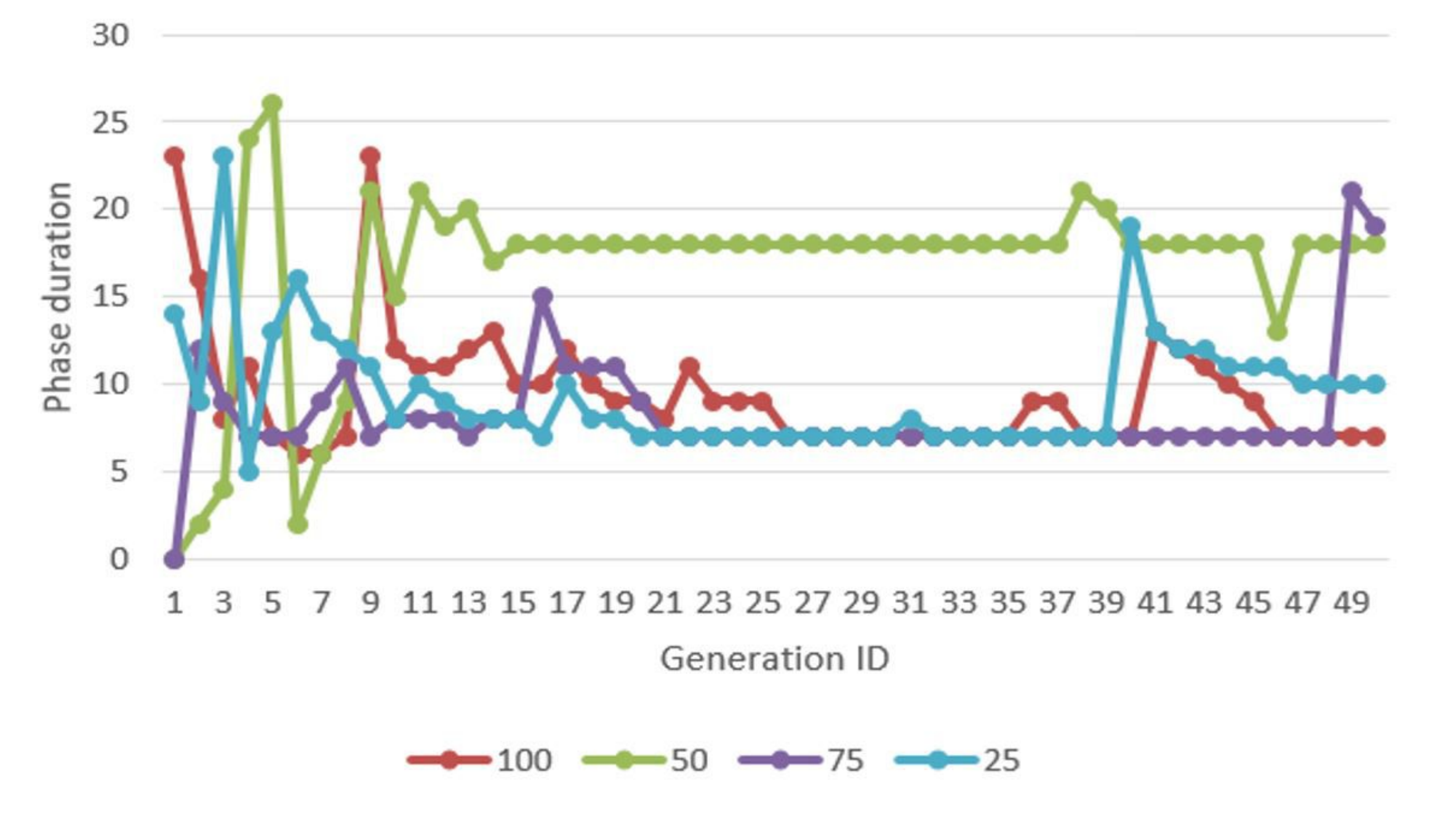}
		\label{fig_a_p43}}
	\subfloat[phase 4]{\includegraphics[width=1.5 in]{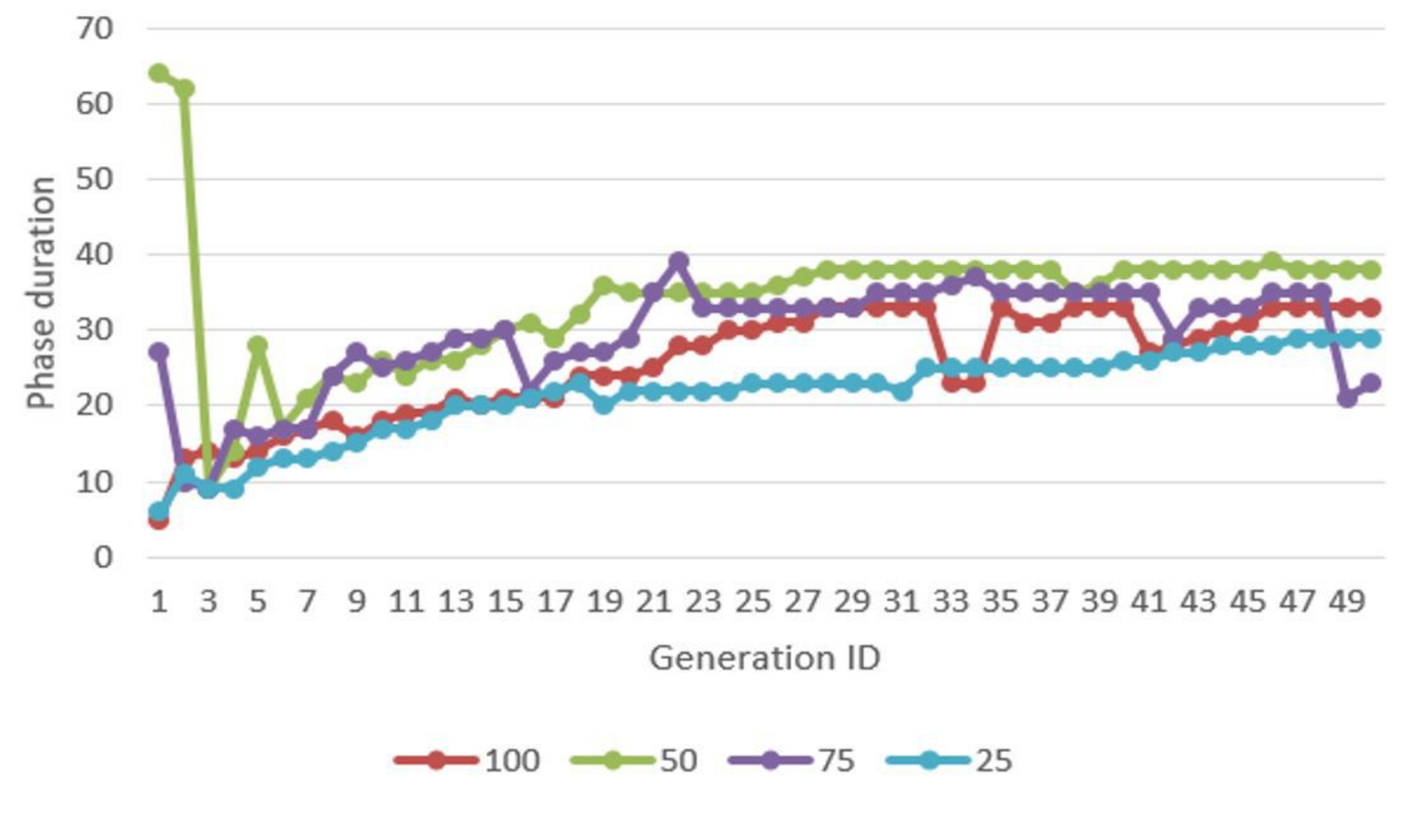}
		\label{fig_a_p44}}
	\caption{Phase duration convergence in intersection 4}
	\label{Annex_result2}
\end{figure}

\section{}
This section contains the graphs of the phase duration convergence in the scenario 1: without traffic incident. 
\begin{figure}[H]
	\centering
	\subfloat[phase 1]{\includegraphics[width=1.5 in]{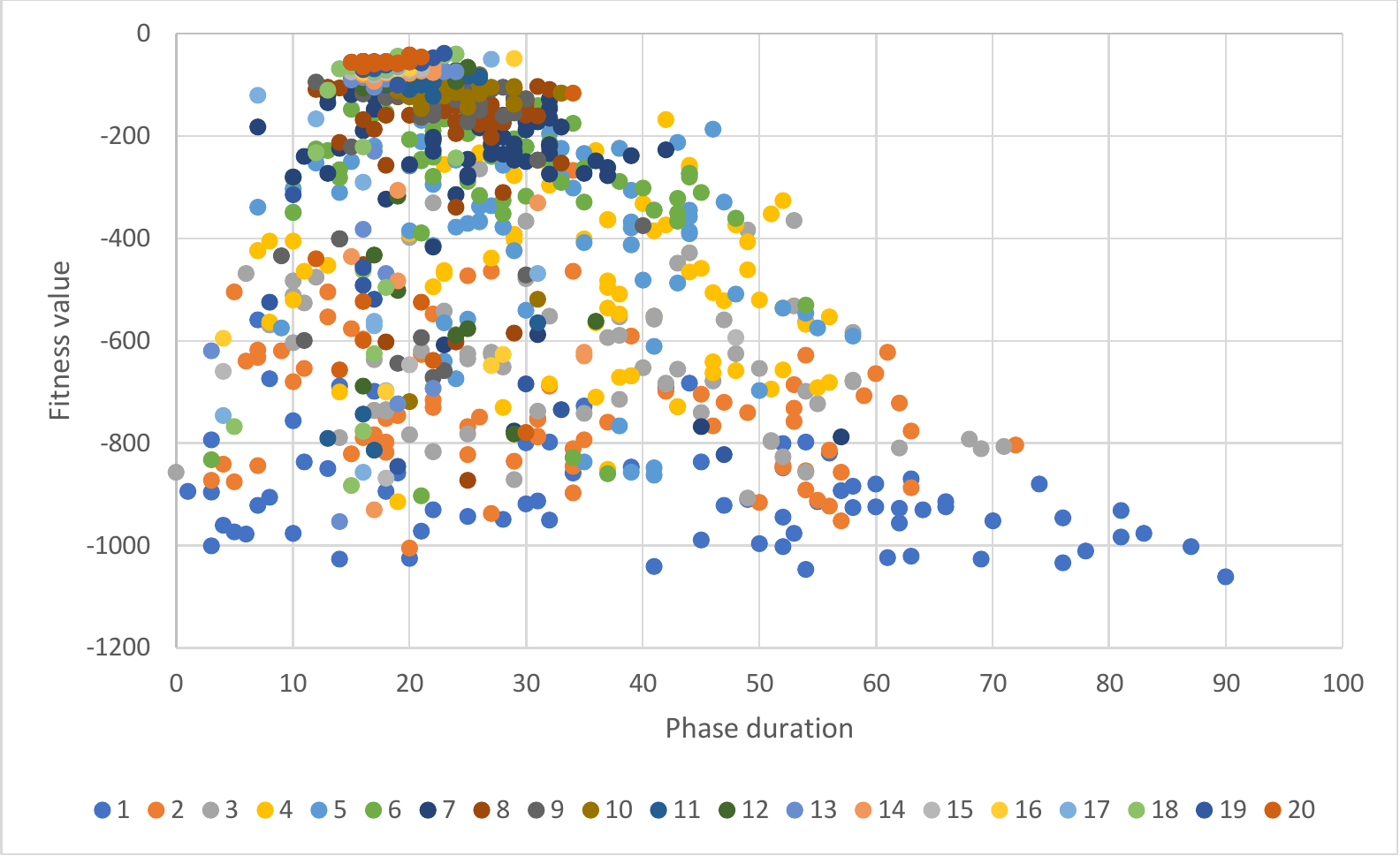}
		\label{fig_a2_p21}}
	\subfloat[phase 2]{\includegraphics[width=1.5 in]{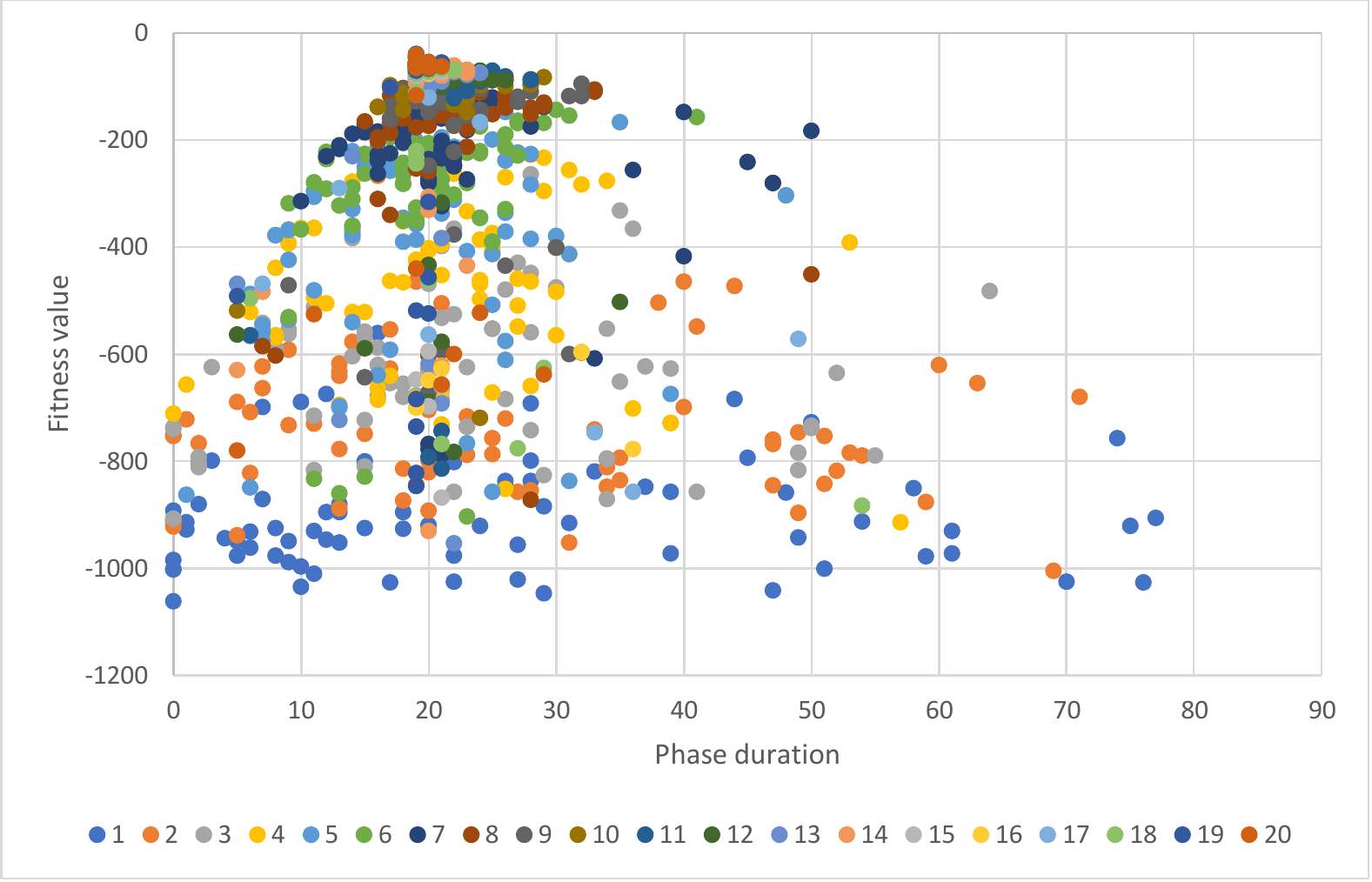}
		\label{fig_a2_p22}}
	\\
	\subfloat[phase 3]{\includegraphics[width=1.5 in]{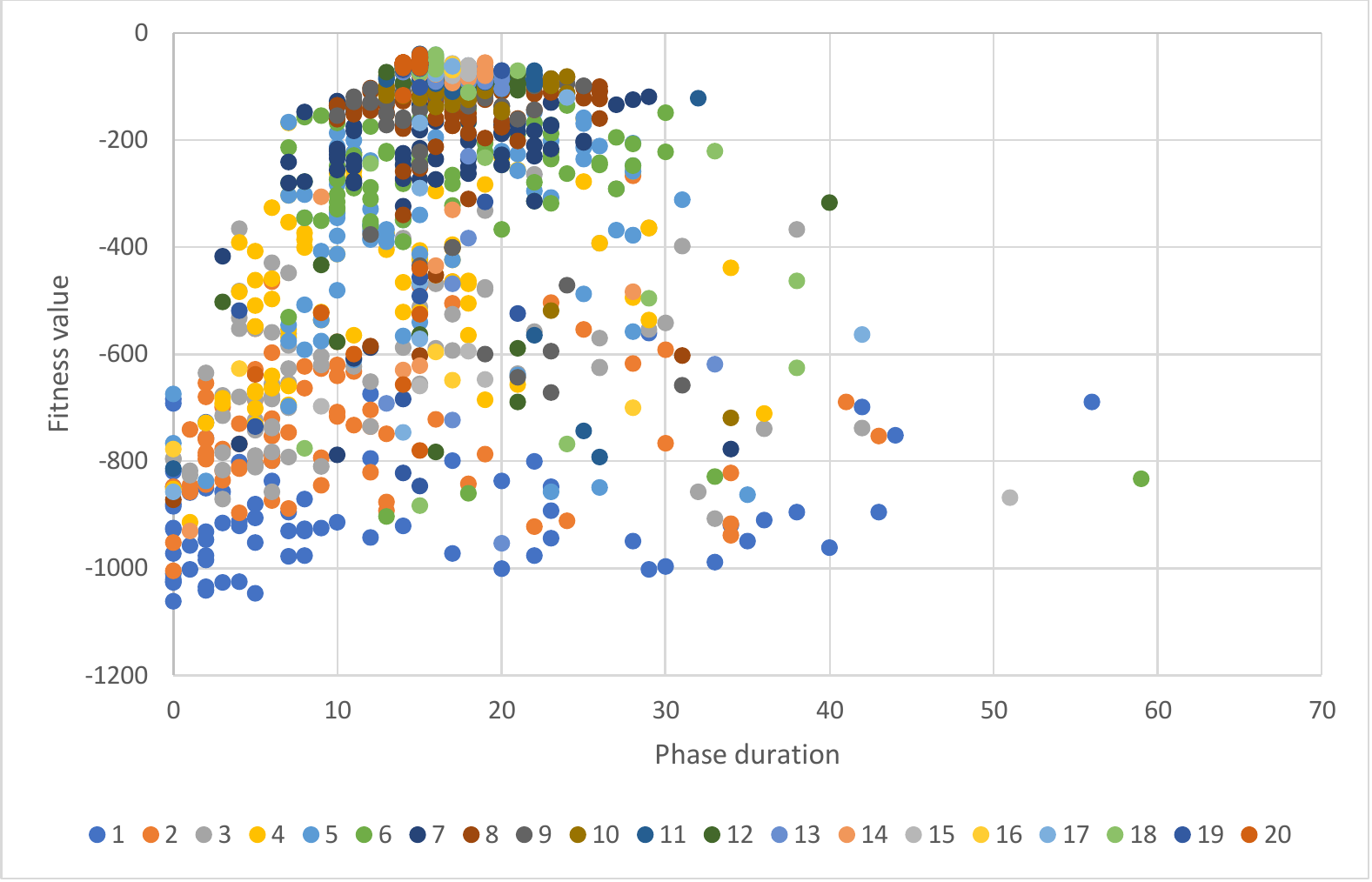}
		\label{fig_a2_p23}}
	\subfloat[phase 4]{\includegraphics[width=1.5 in]{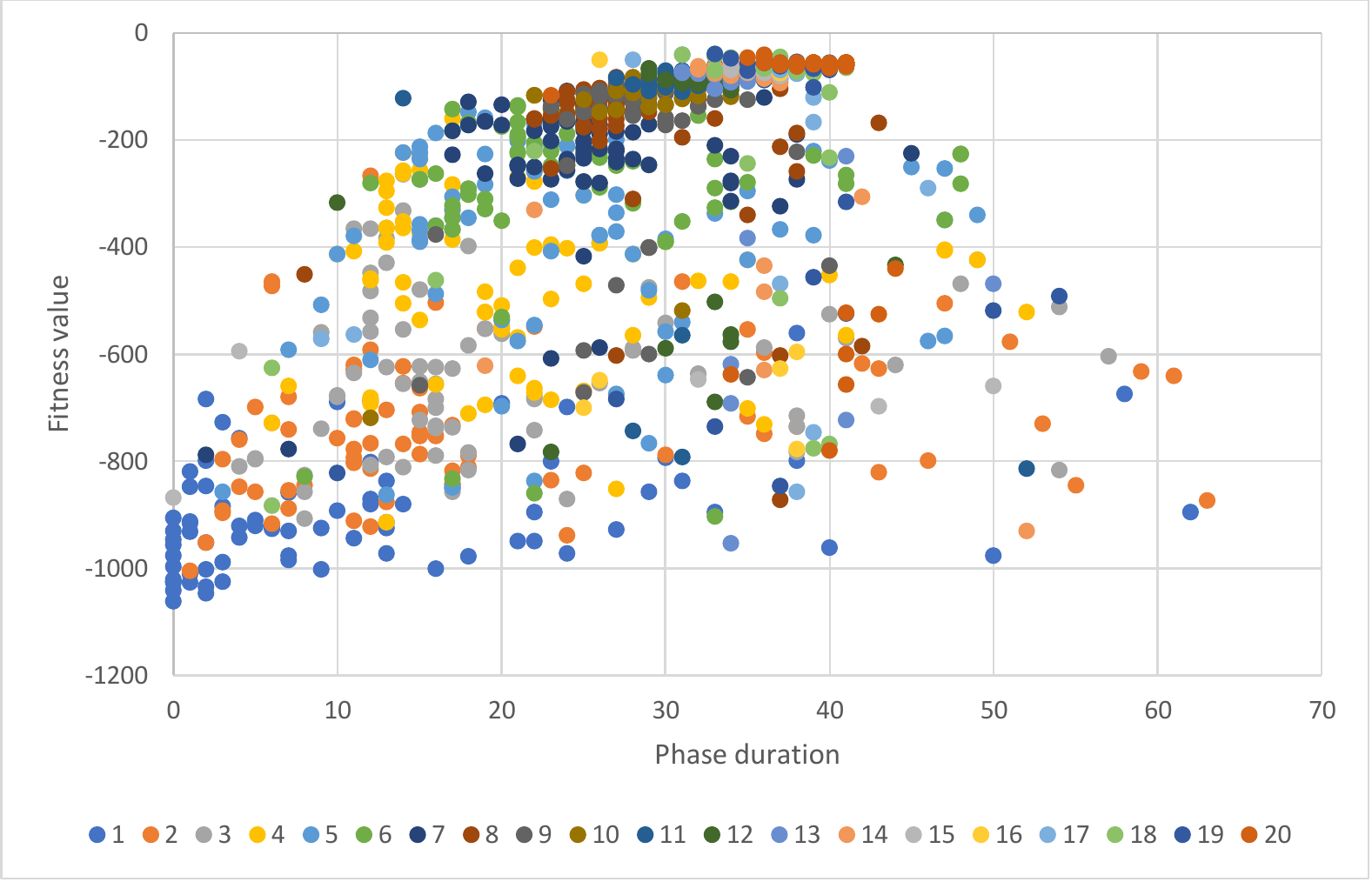}
		\label{fig_a2_p24}}
	\caption{Phase duration convergence in intersection 2}
	\label{Annex_result3}
\end{figure}

\begin{figure}[H]
	\centering
	\subfloat[phase 1]{\includegraphics[width=1.5 in]{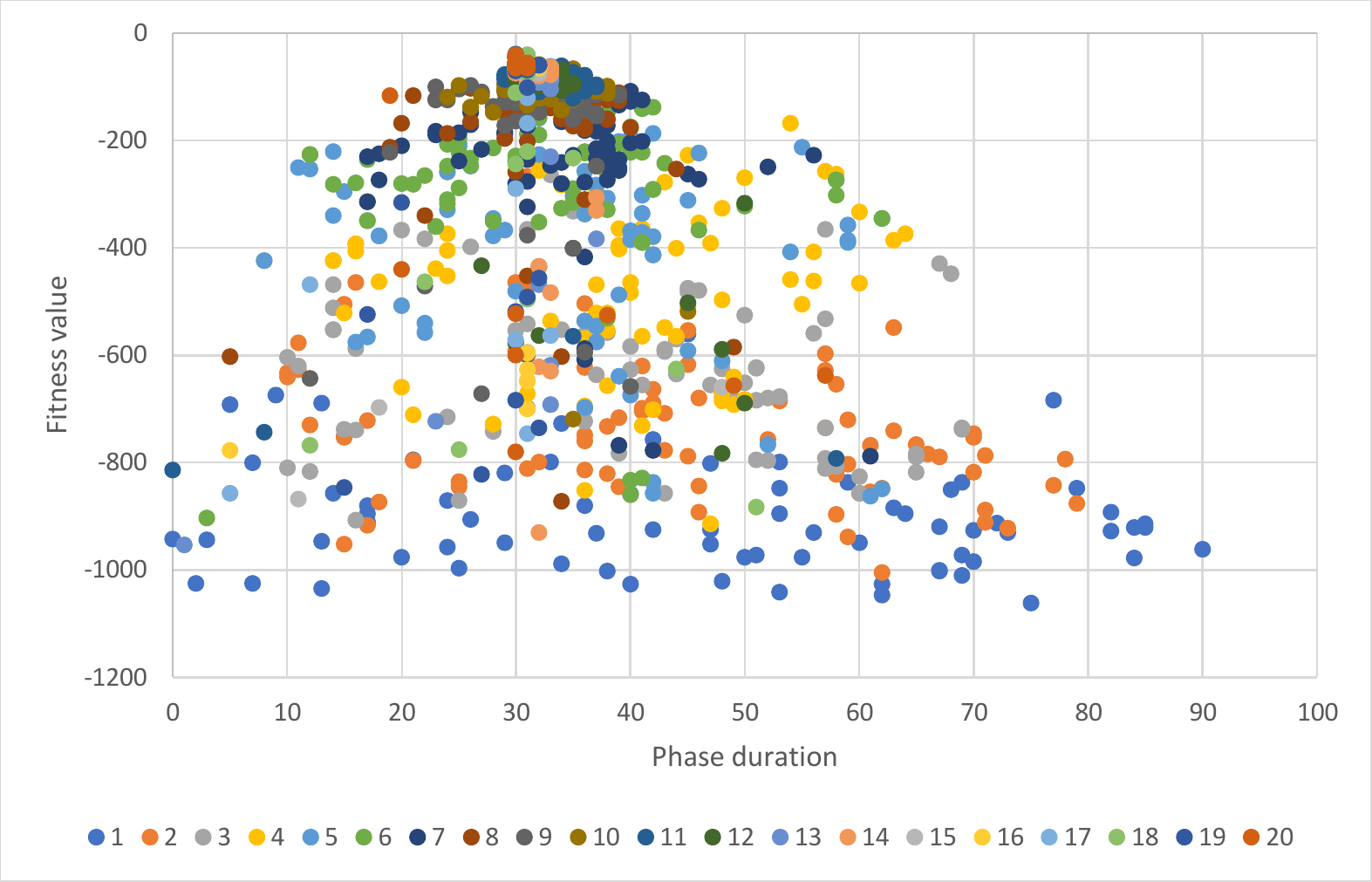}
		\label{fig_a2_p41}}
	\subfloat[phase 2]{\includegraphics[width=1.5 in]{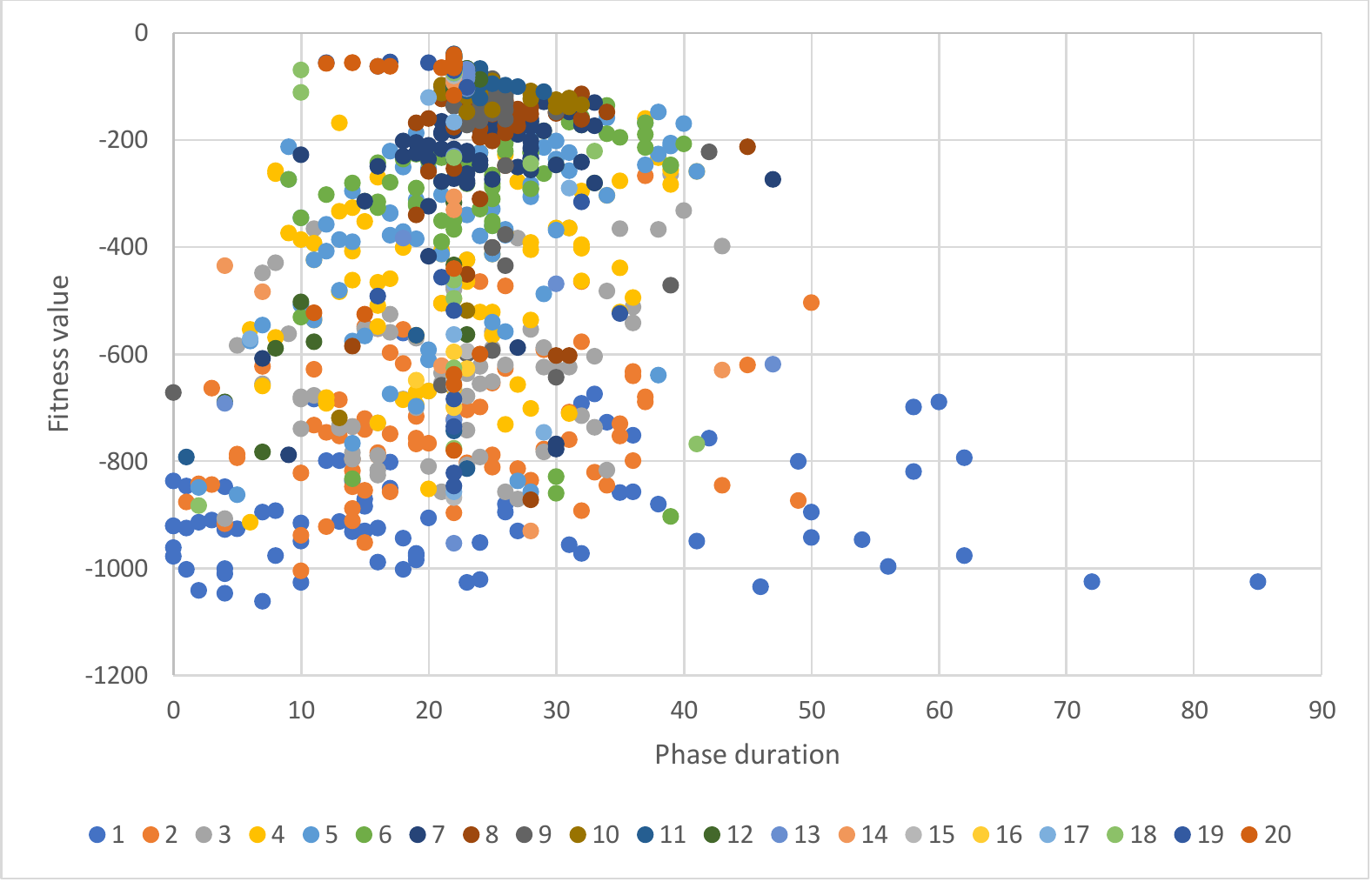}
		\label{fig_a2_p42}}
	\\
	\subfloat[phase 3]{\includegraphics[width=1.5 in]{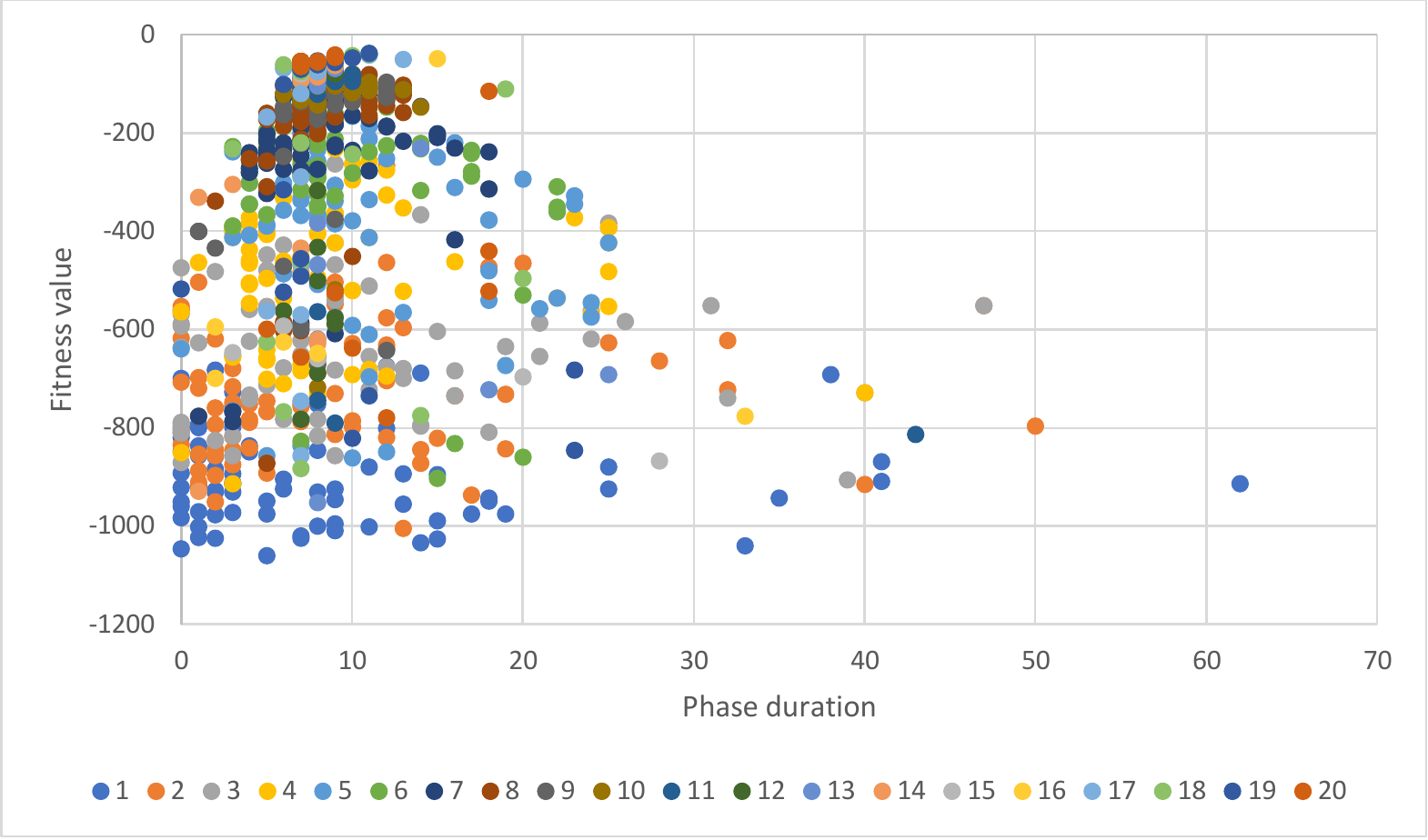}
		\label{fig_a2_p43}}
	\subfloat[phase 4]{\includegraphics[width=1.5 in]{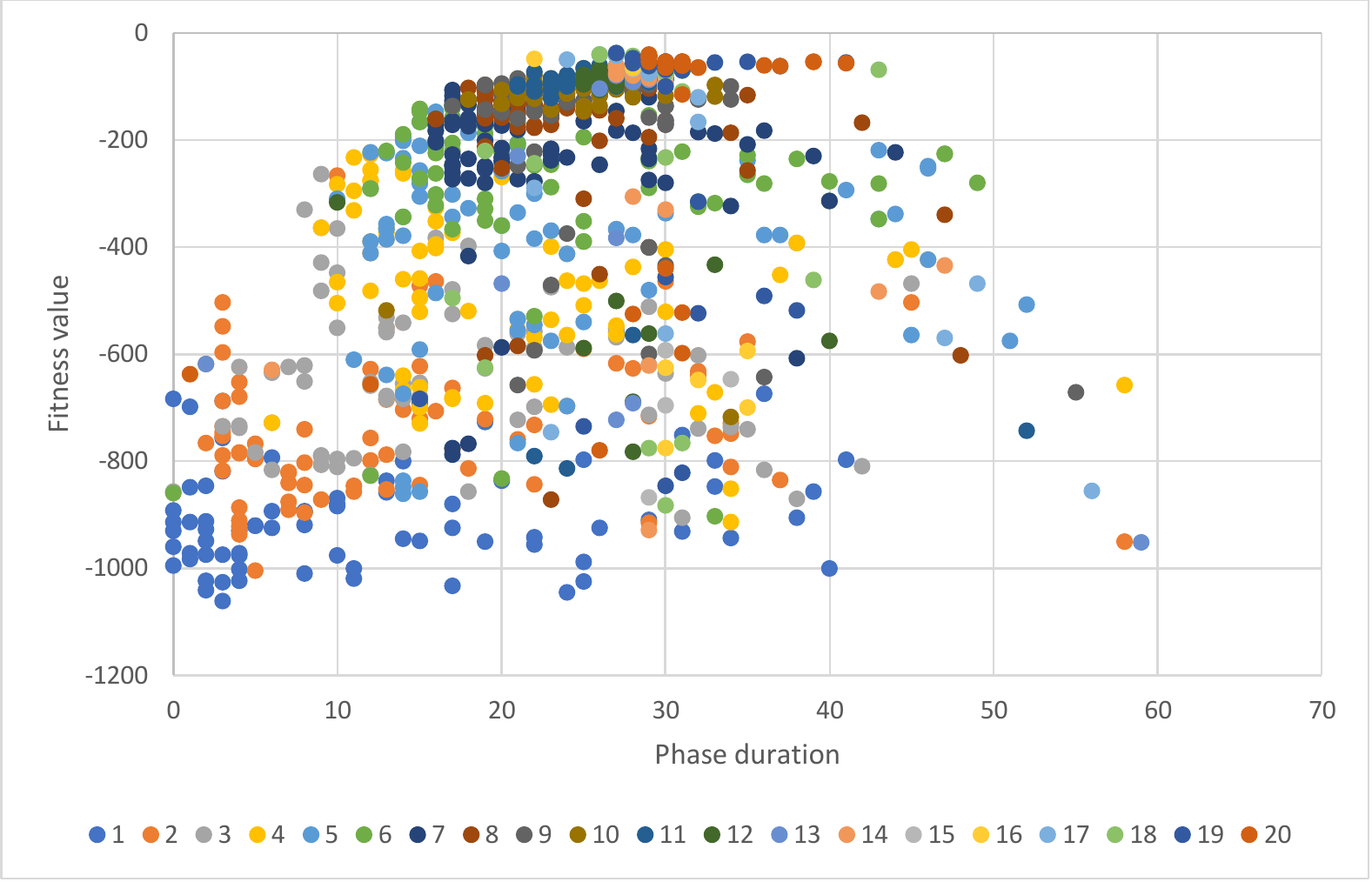}
		\label{fig_a2_p44}}
	\caption{Phase duration convergence in intersection 4}
	\label{Annex_result4}
\end{figure}
\section{}
This section contains the graphs of the phase duration convergence in the scenario 3: with traffic incident. 
\begin{figure}[H]
	\centering
	\subfloat[phase 1]{\includegraphics[width=1.5 in]{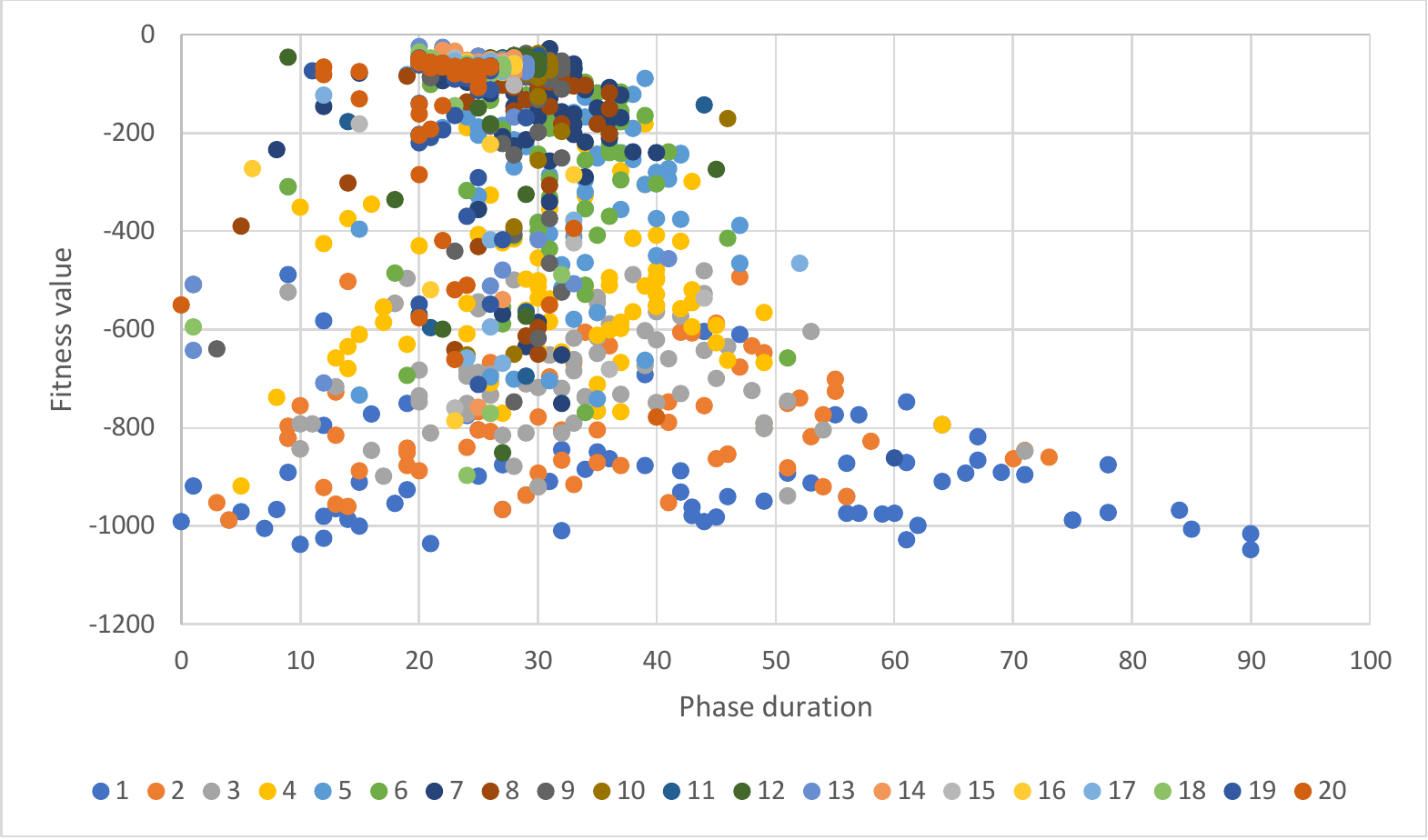}
		\label{fig_a3_p11}}
	\subfloat[phase 2]{\includegraphics[width=1.5 in]{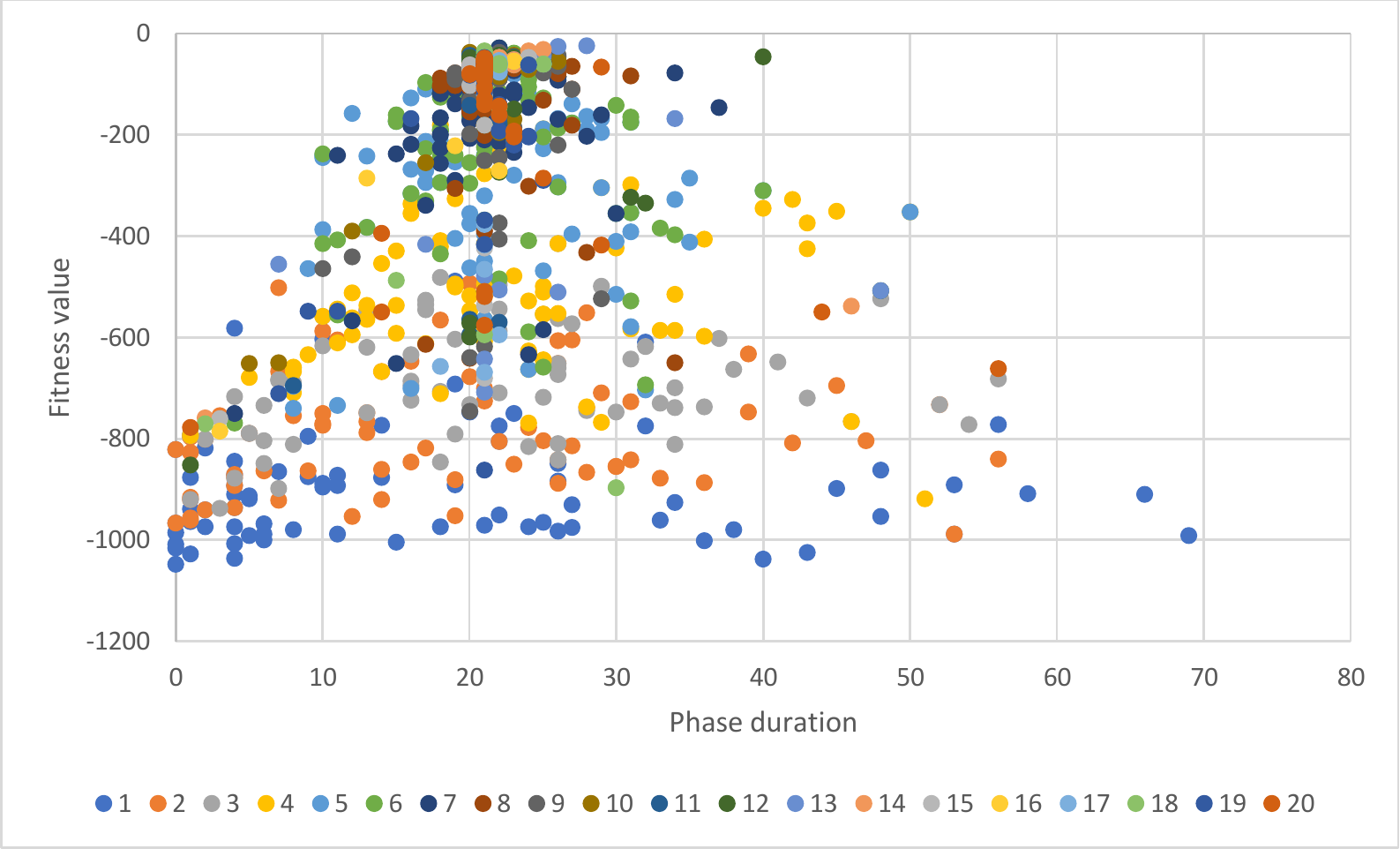}
		\label{fig_a3_p12}}
	\\
	\subfloat[phase 3]{\includegraphics[width=1.5 in]{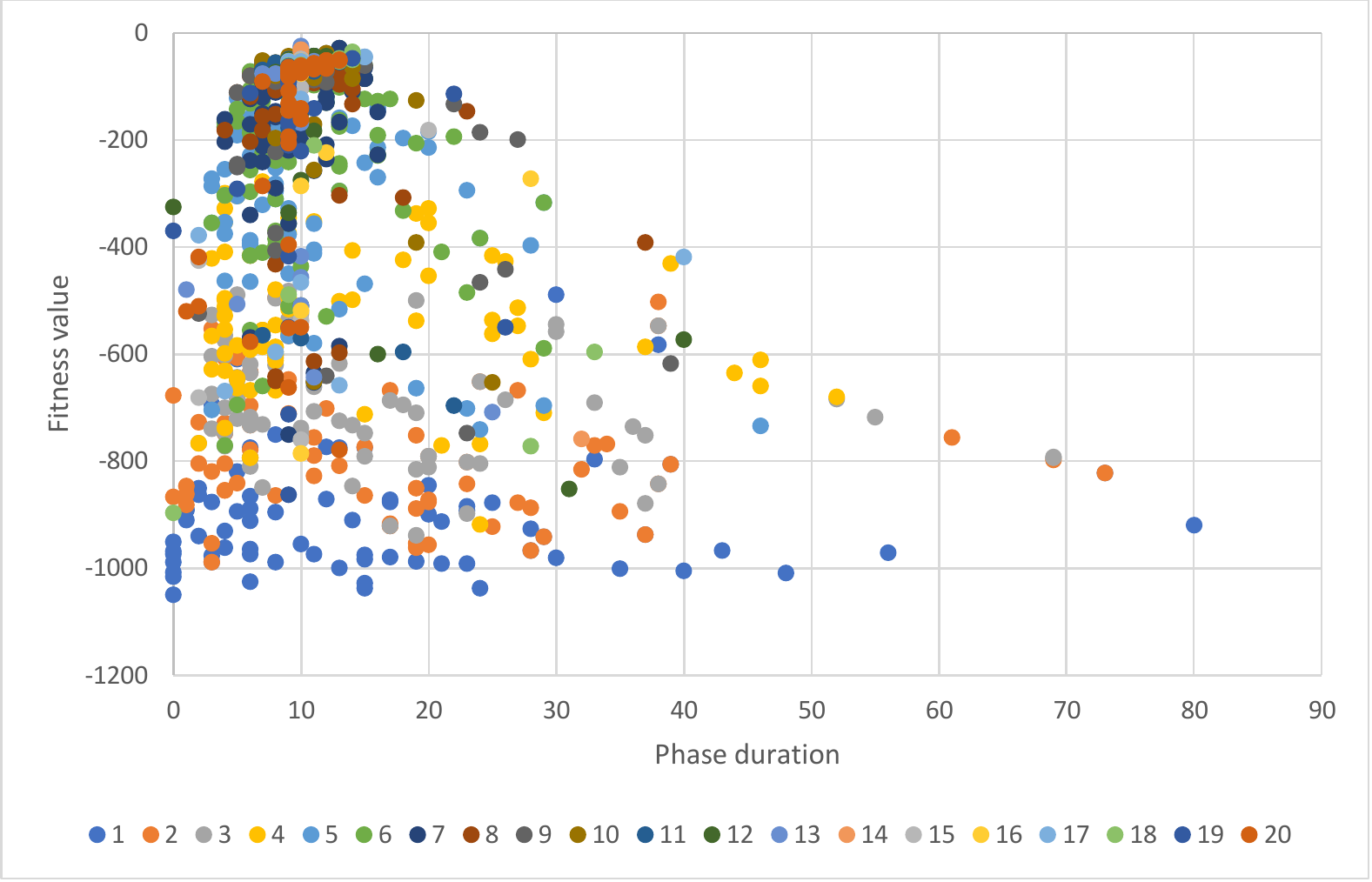}
		\label{fig_a3_p13}}
	\subfloat[phase 4]{\includegraphics[width=1.5 in]{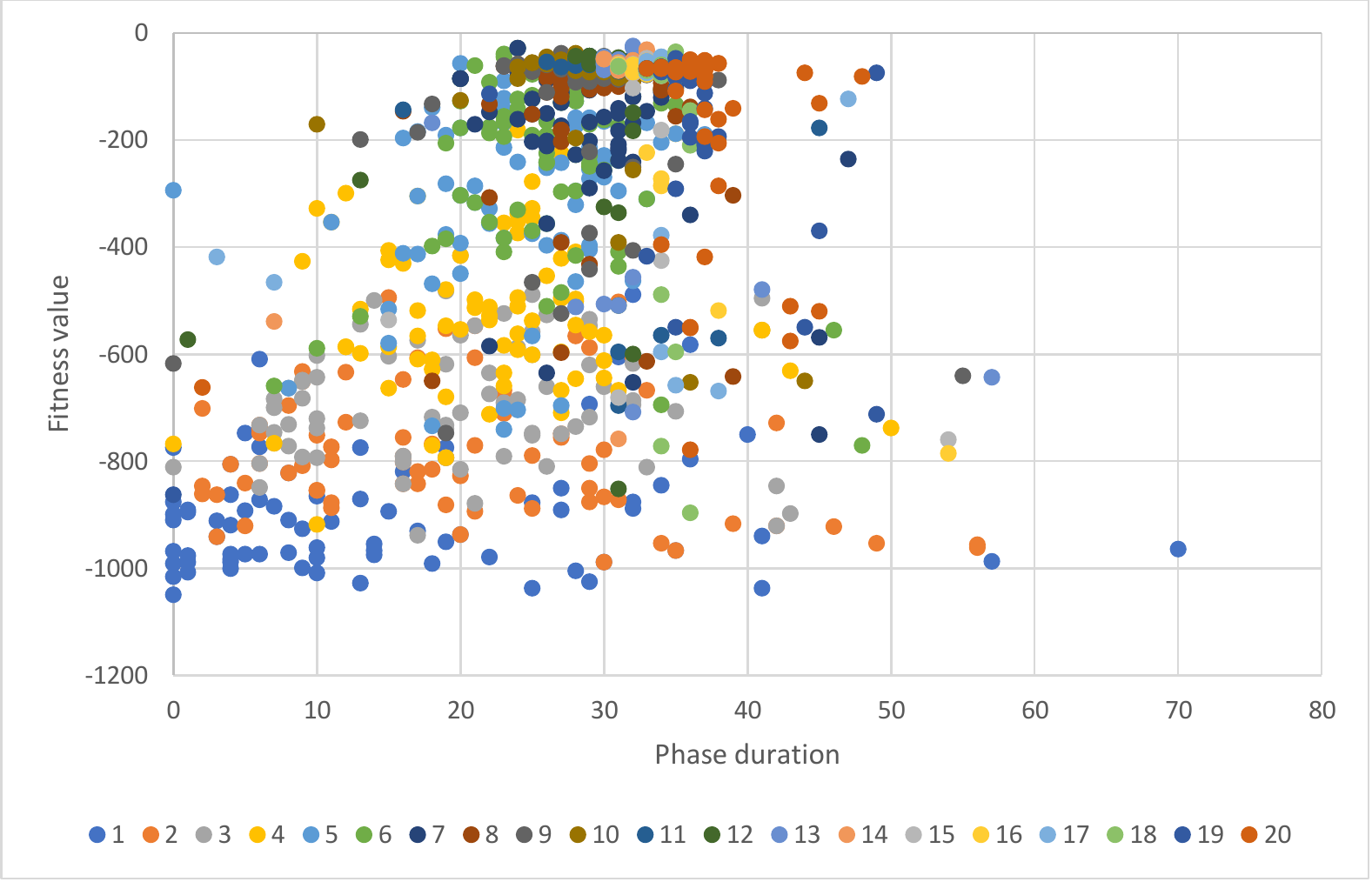}
		\label{fig_a3_p14}}
	\caption{Phase duration convergence in intersection 1}
	\label{Annex_result5}
\end{figure}
\begin{figure}[H]
	\centering
	\subfloat[phase 1]{\includegraphics[width=1.5 in]{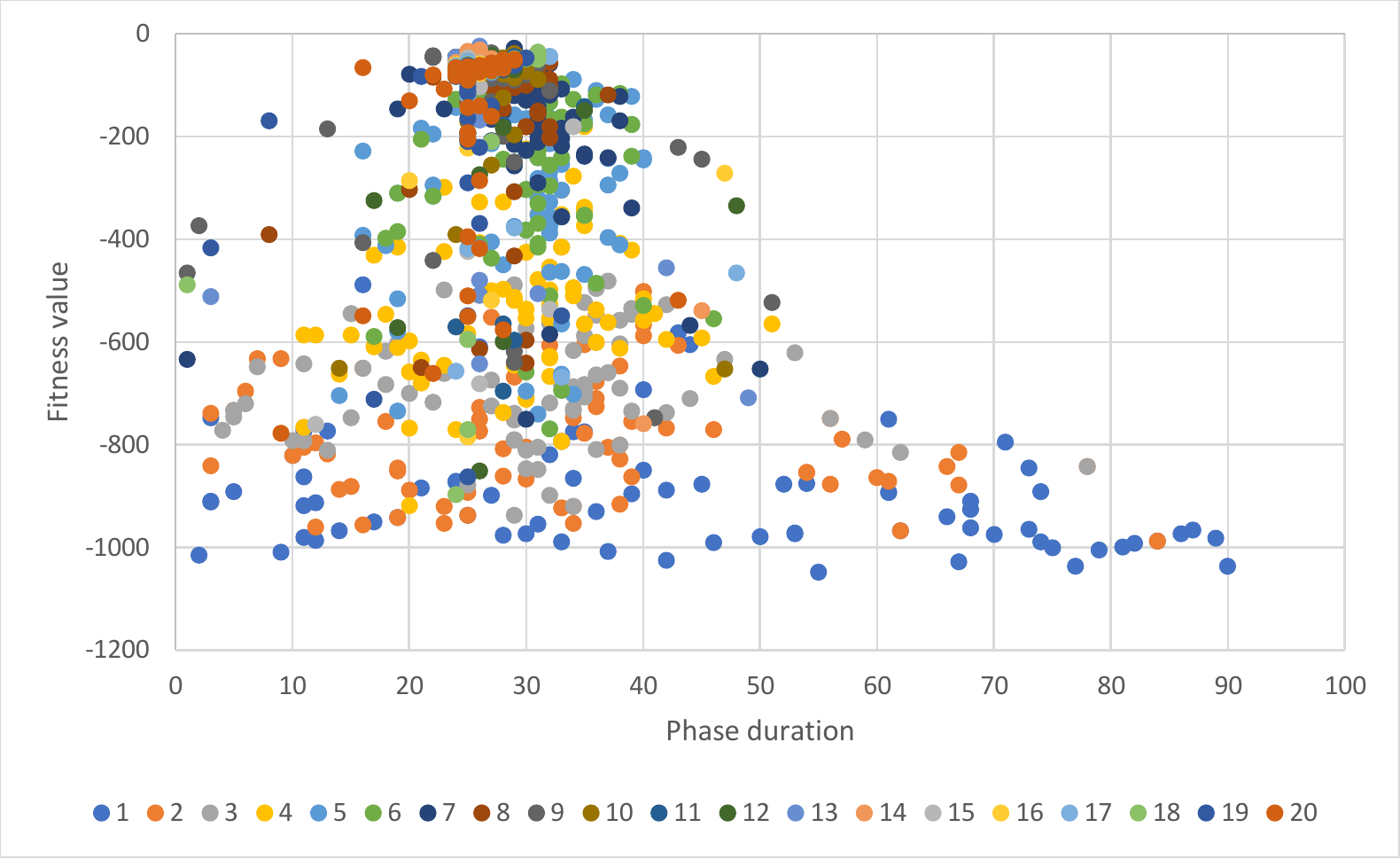}
		\label{fig_a3_p21}}
	\subfloat[phase 2]{\includegraphics[width=1.5 in]{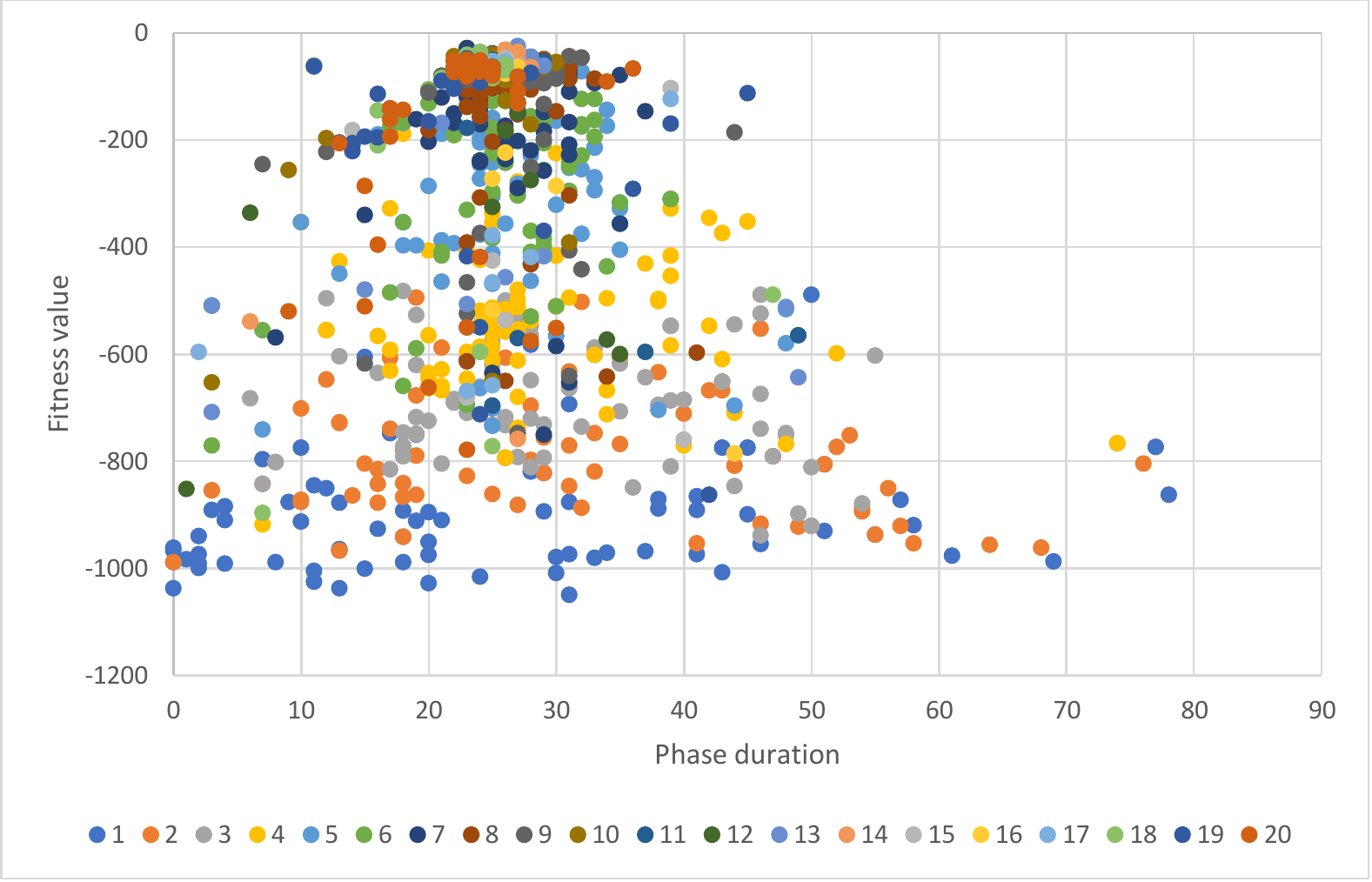}
		\label{fig_a3_p22}}
	\\
	\subfloat[phase 3]{\includegraphics[width=1.5 in]{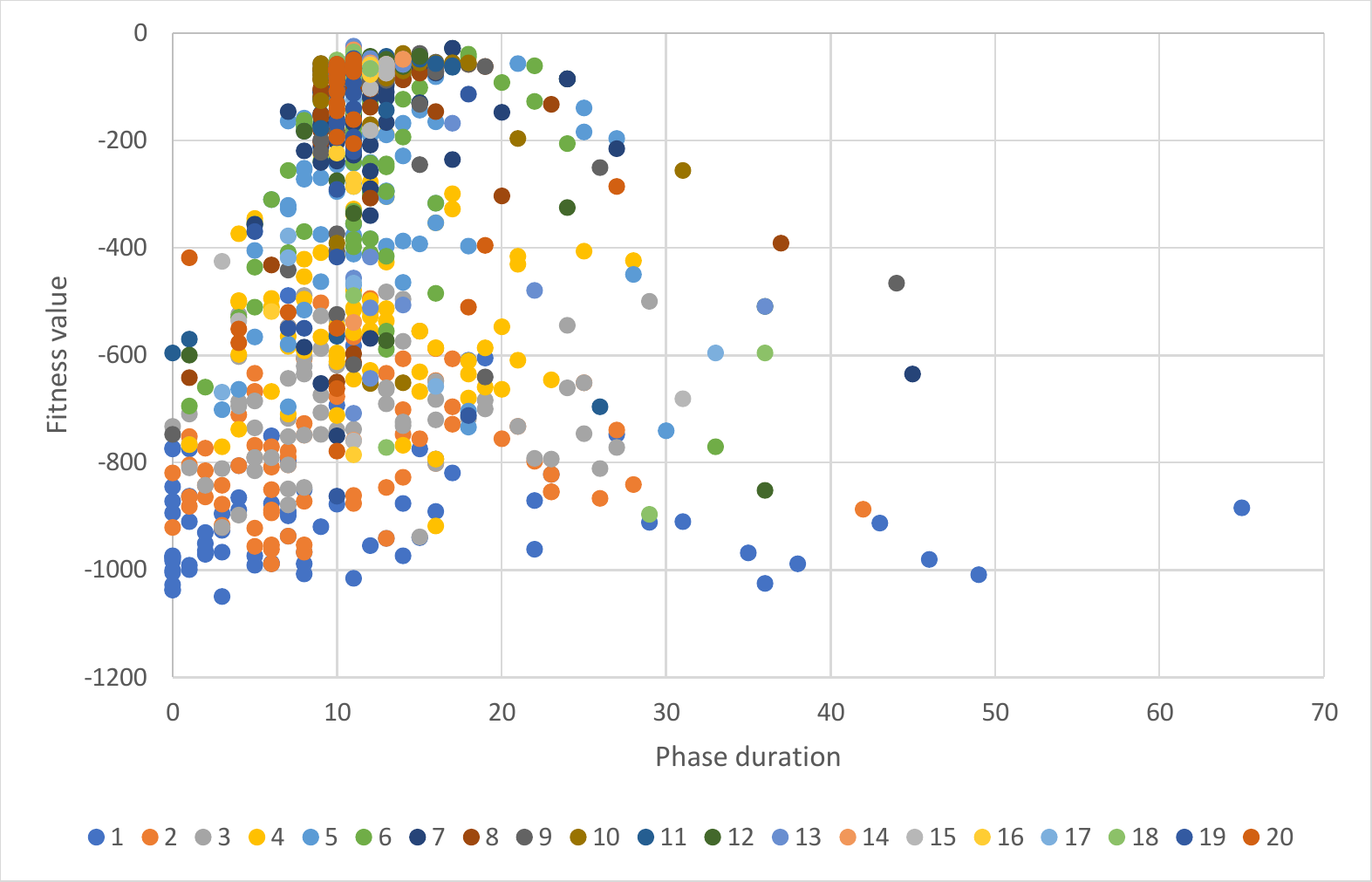}
		\label{fig_a3_p23}}
	\subfloat[phase 4]{\includegraphics[width=1.5 in]{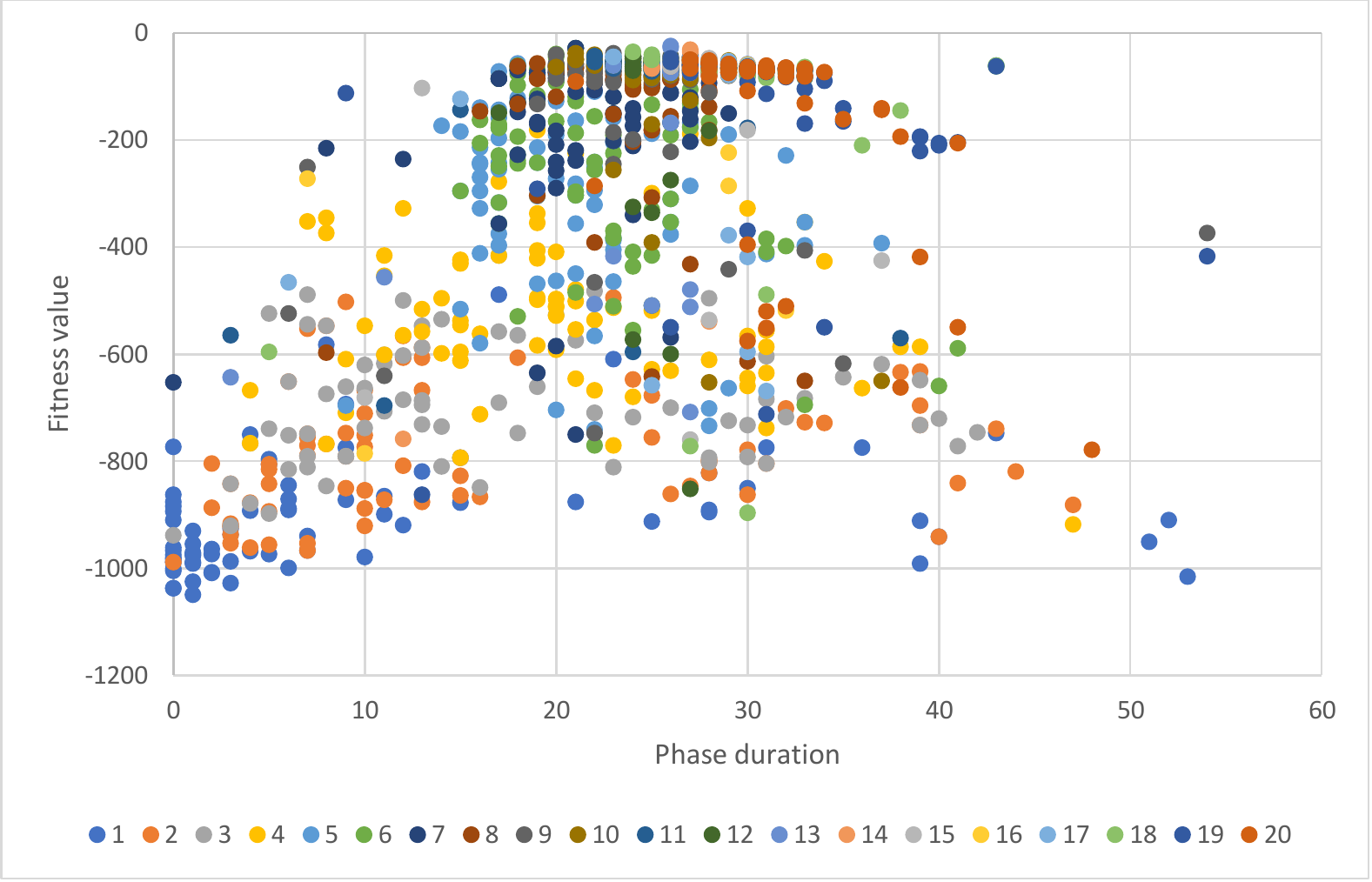}
		\label{fig_a3_p24}}
	\caption{Phase duration convergence in intersection 2}
	\label{Annex_result6}
\end{figure}
\begin{figure}[H]
	\centering
	\subfloat[phase 1]{\includegraphics[width=1.5 in]{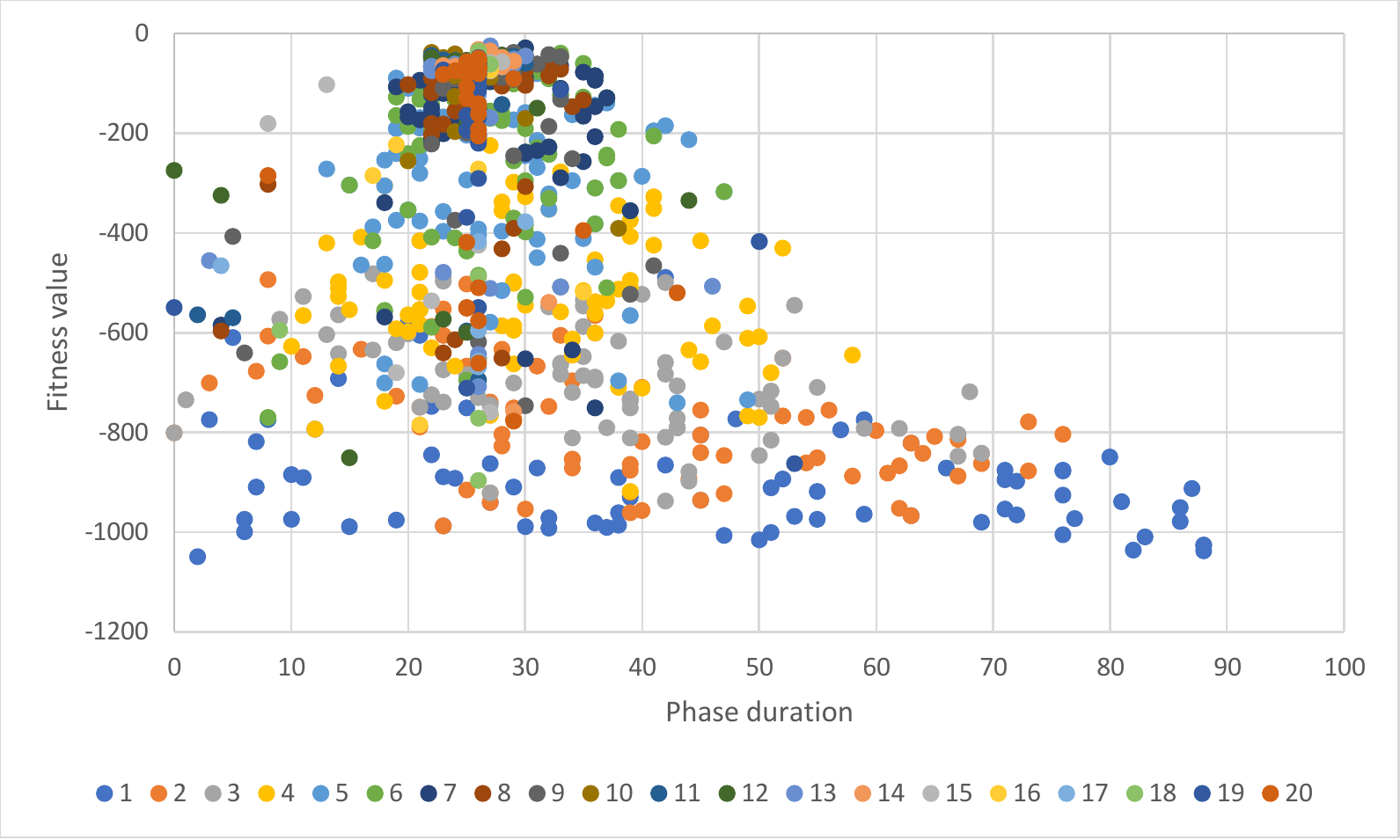}
		\label{fig_a3_p31}}
	\subfloat[phase 2]{\includegraphics[width=1.5 in]{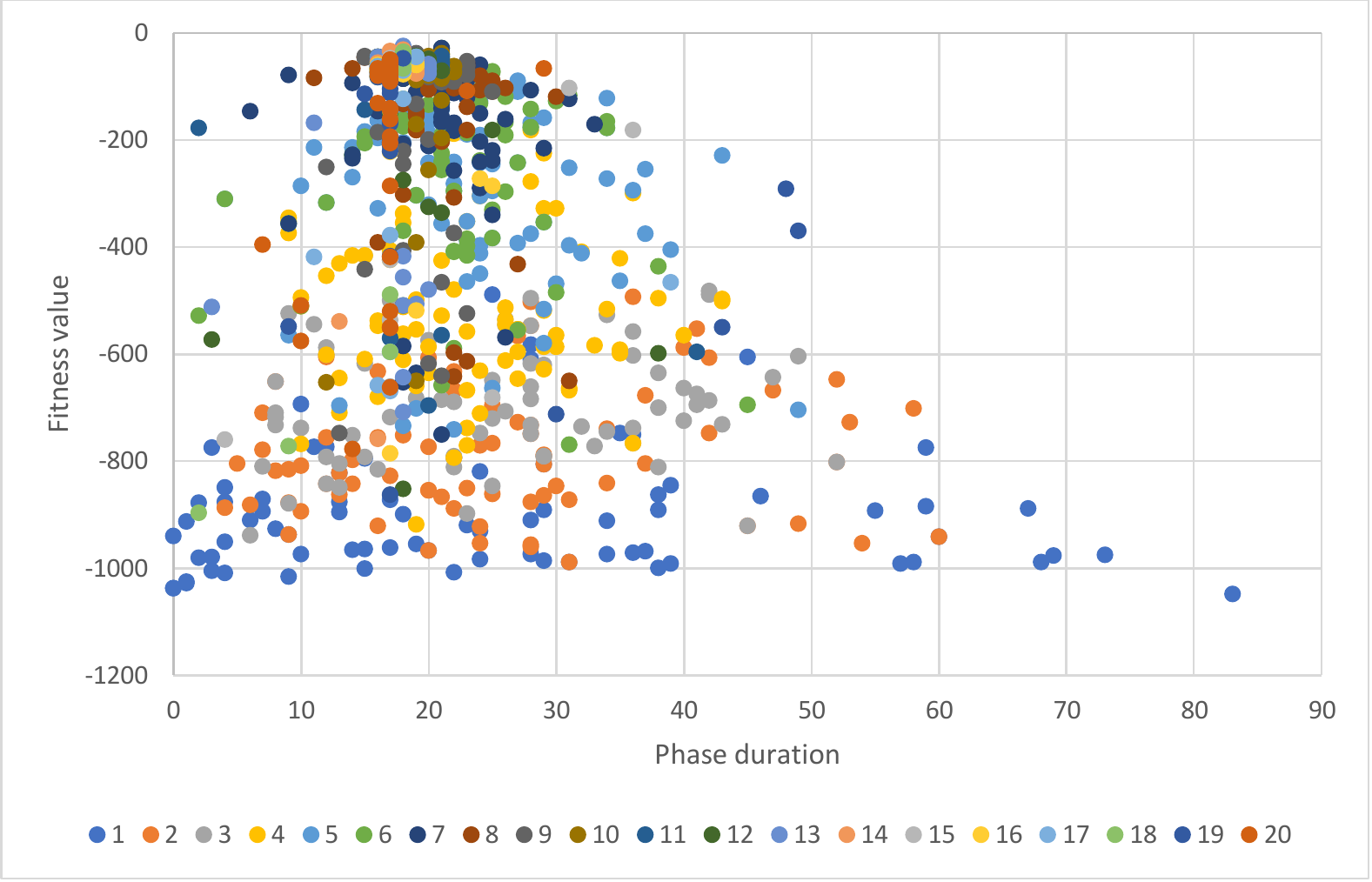}
		\label{fig_a3_p32}}
	\\
	\subfloat[phase 3]{\includegraphics[width=1.5 in]{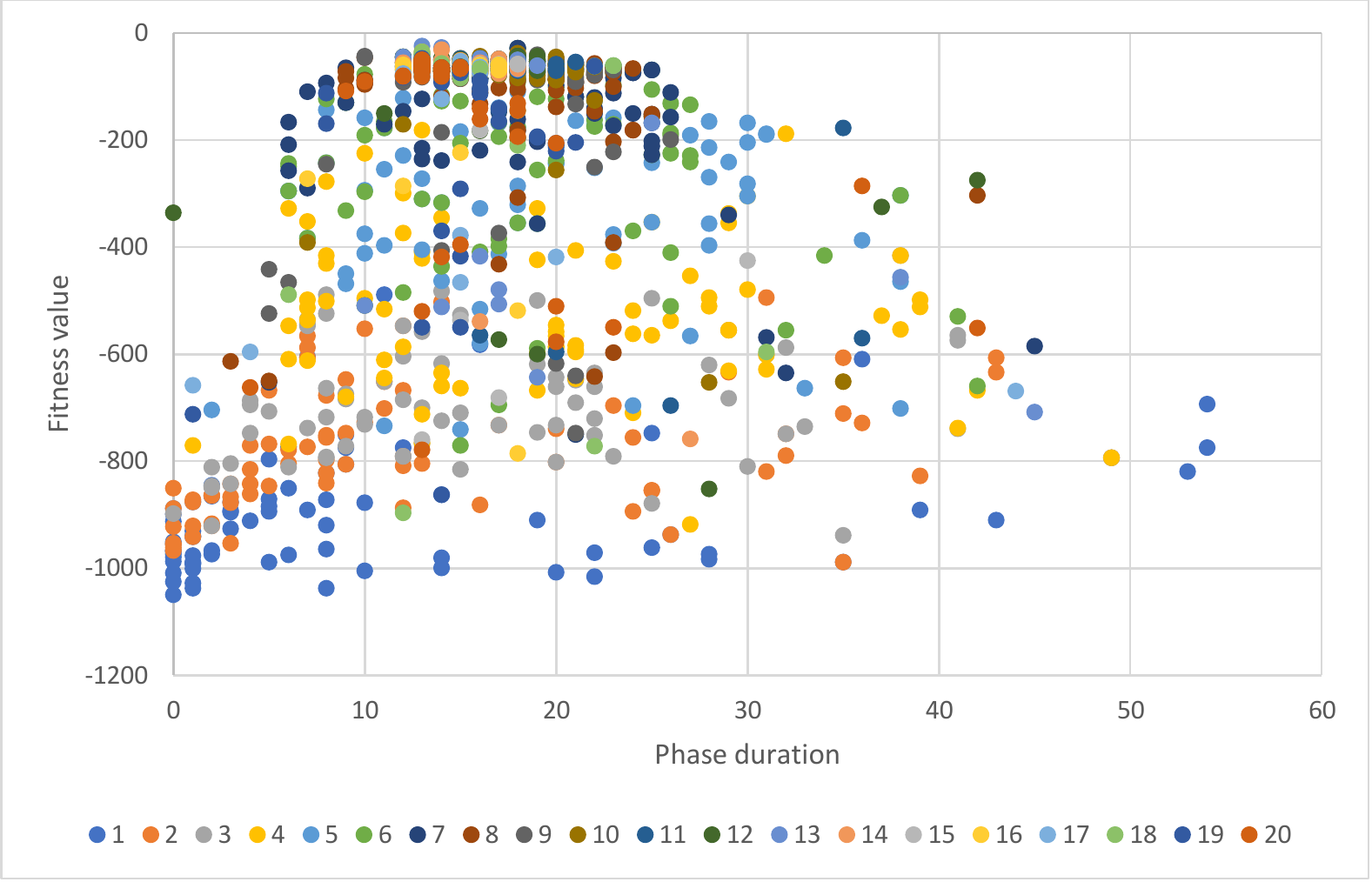}
		\label{fig_a3_p33}}
	\subfloat[phase 4]{\includegraphics[width=1.5 in]{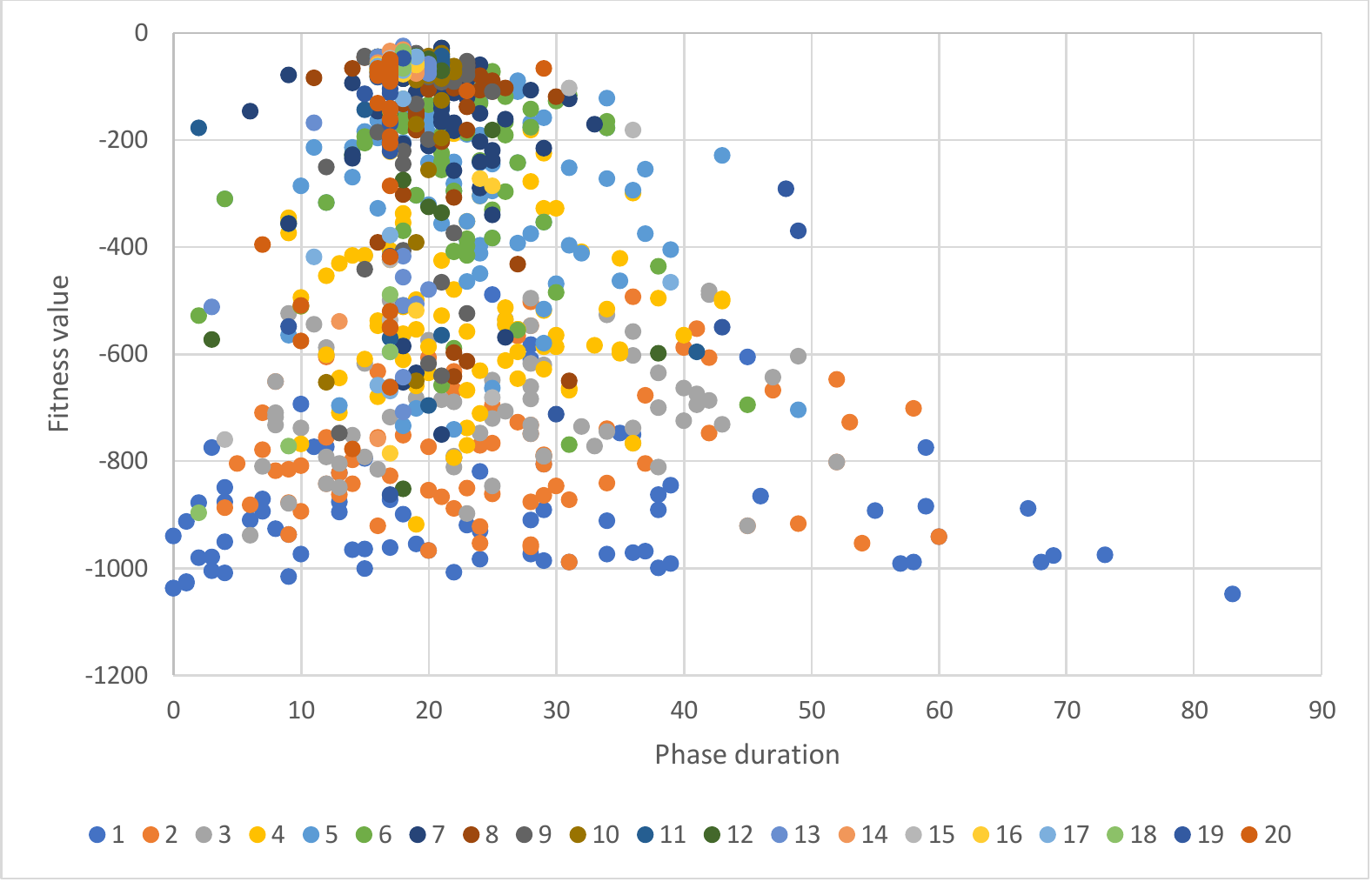}
		\label{fig_a3_p34}}
	\caption{Phase duration convergence in intersection 3}
	\label{Annex_result7}
\end{figure}
\begin{figure}[H]
	\centering
	\subfloat[phase 1]{\includegraphics[width=1.5 in]{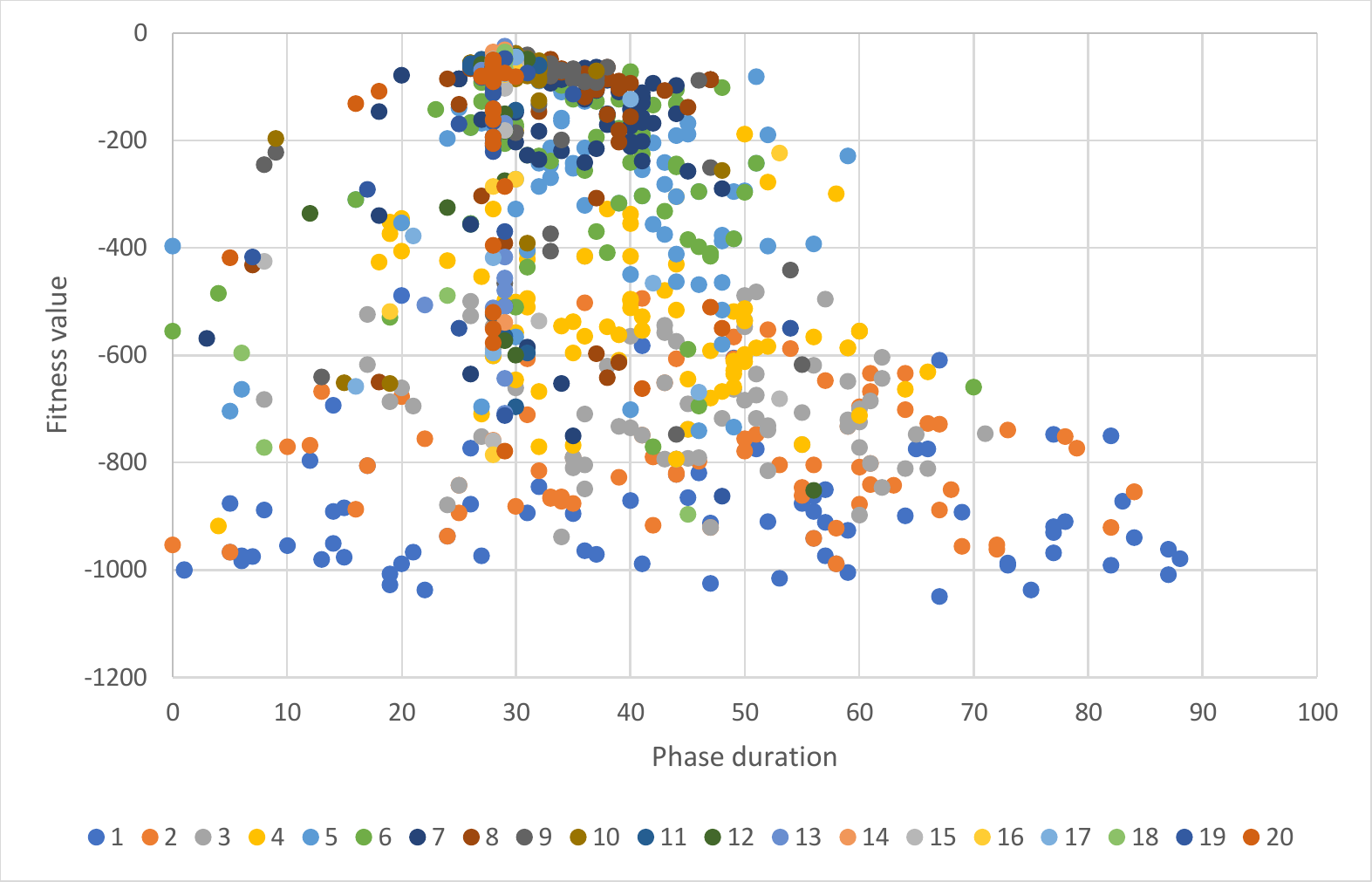}
		\label{fig_a3_p41}}
	\subfloat[phase 2]{\includegraphics[width=1.5 in]{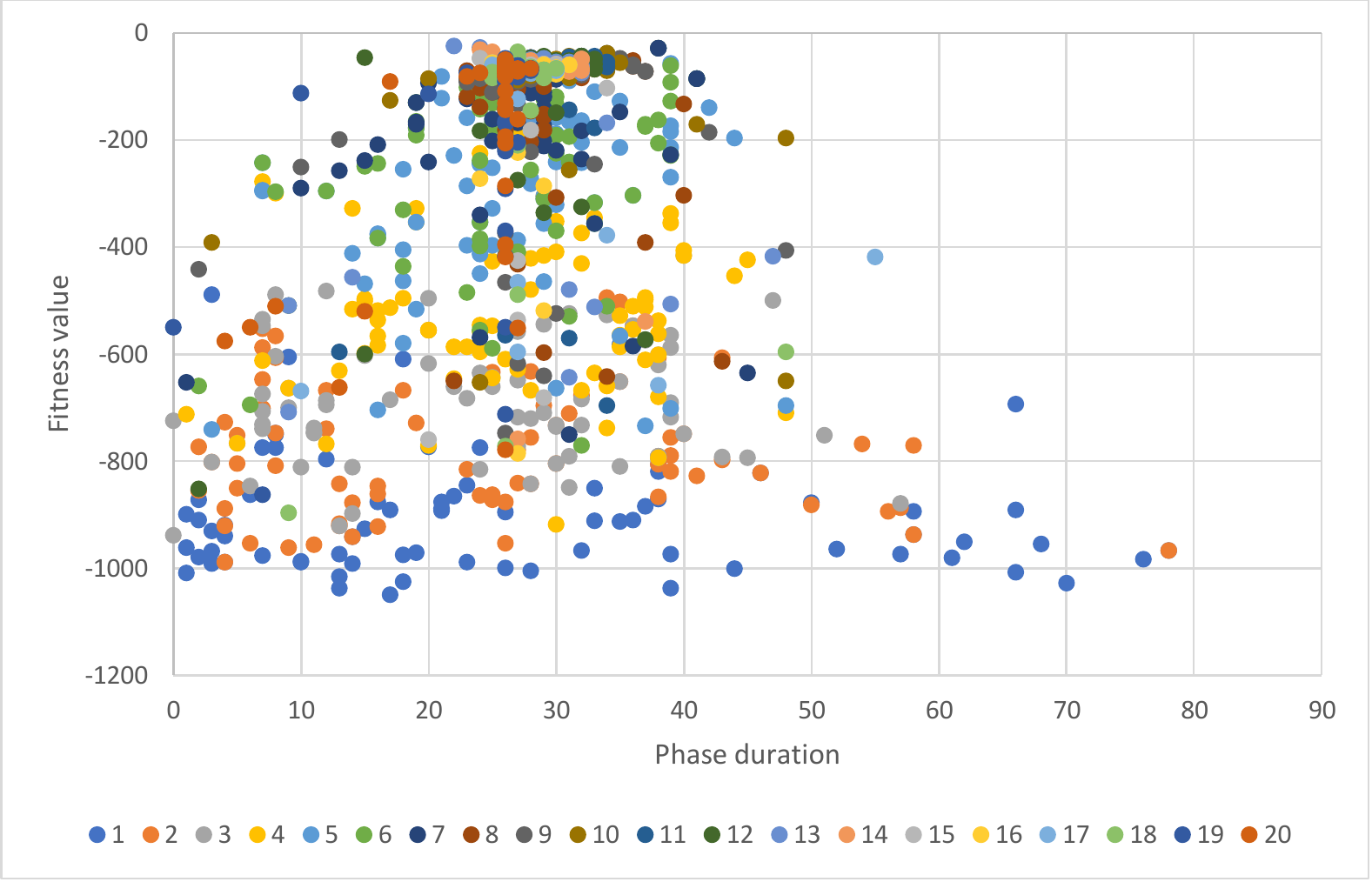}
		\label{fig_a3_p42}}
	\\
	\subfloat[phase 3]{\includegraphics[width=1.5 in]{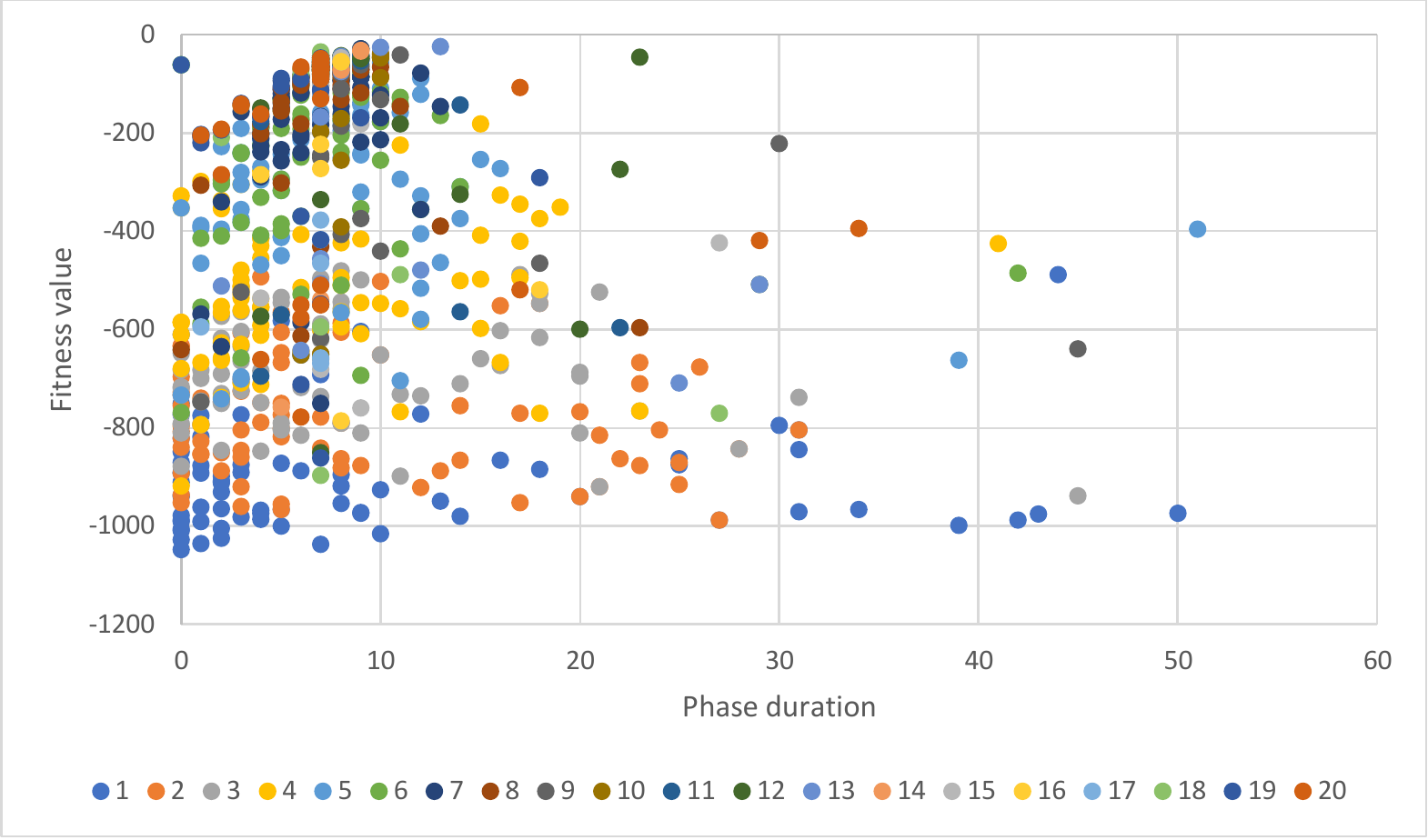}
		\label{fig_a3_p43}}
	\subfloat[phase 4]{\includegraphics[width=1.5 in]{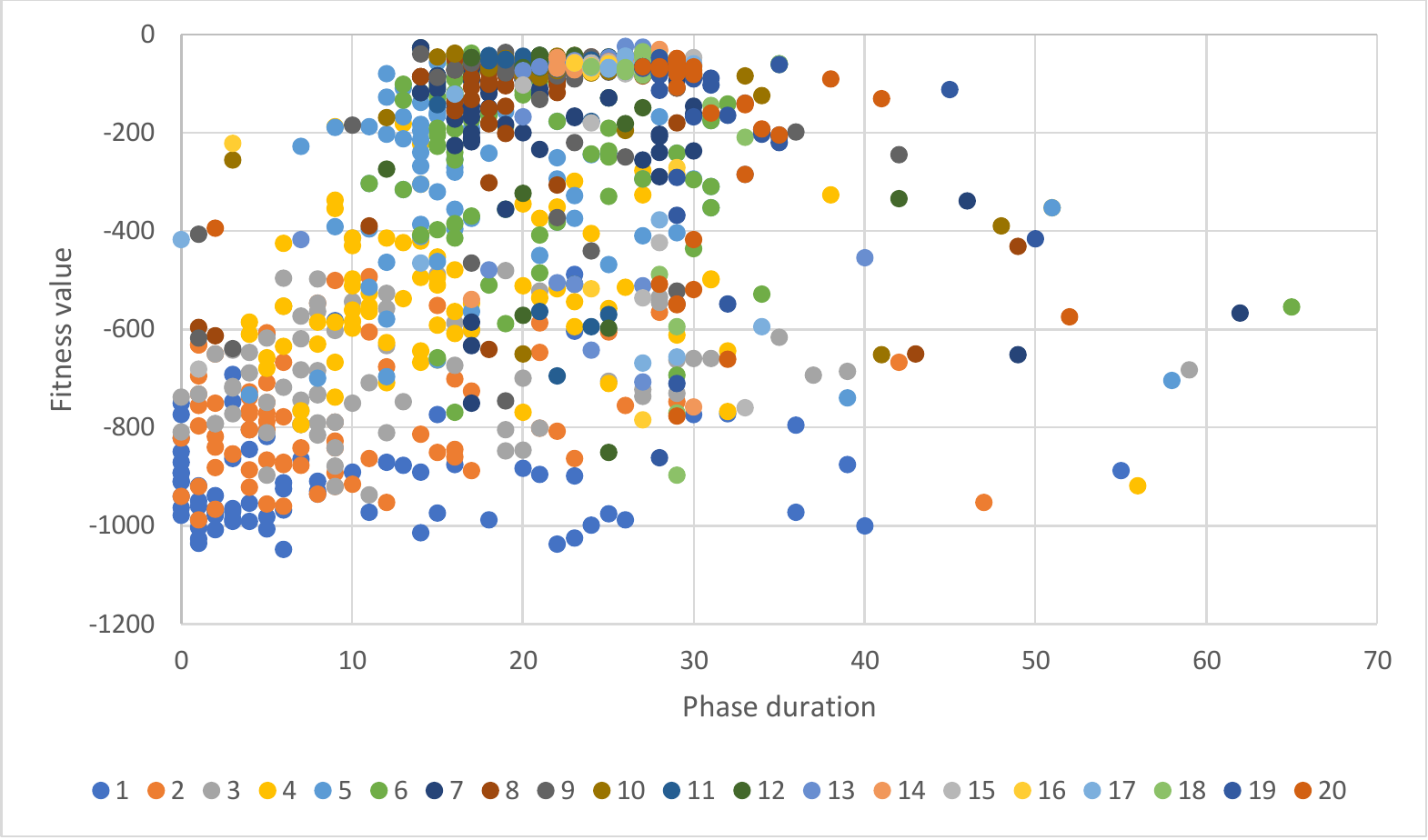}
		\label{fig_a3_p44}}
	\caption{Phase duration convergence in intersection 4}
	\label{Annex_result8}
\end{figure}

\ifCLASSOPTIONcaptionsoff
  \newpage
\fi

\printbibliography

@book{montgomery2012introduction,
  title={Introduction to linear regression analysis},
  author={Montgomery, Douglas C and Peck, Elizabeth A and Vining, G Geoffrey},
  volume={821},
  year={2012},
  publisher={John Wiley \& Sons}
}

@book{weisberg2005applied,
  title={Applied linear regression},
  author={Weisberg, Sanford},
  volume={528},
  year={2005},
  publisher={John Wiley \& Sons}
}

@article{breiman2001random,
  title={Random forests},
  author={Breiman, Leo},
  journal={Machine learning},
  volume={45},
  number={1},
  pages={5--32},
  year={2001},
  publisher={Springer}
}

@article{liaw2002classification,
  title={Classification and regression by randomForest},
  author={Liaw, Andy and Wiener, Matthew and others},
  journal={R news},
  volume={2},
  number={3},
  pages={18--22},
  year={2002}
}

@online{GA_python,
  author = {Vijini, Mallawaarachchi},
  title = {Introduction to Genetic Algorithms — Including Example Code},
  year = 2017,
  url = {https://towardsdatascience.com/introduction-to-genetic-algorithms-including-example-code-e396e98d8bf3},
  urldate = {2017-07-08}
}

@article{casas2017deep,
  title={Deep deterministic policy gradient for urban traffic light control},
  author={Casas, Noe},
  journal={arXiv preprint arXiv:1703.09035},
  year={2017}
}

@article{mousavi2017traffic,
  title={Traffic light control using deep policy-gradient and value-function-based reinforcement learning},
  author={Mousavi, Seyed Sajad and Schukat, Michael and Howley, Enda},
  journal={IET Intelligent Transport Systems},
  volume={11},
  number={7},
  pages={417--423},
  year={2017},
  publisher={IET}
}

@article{van2016deep,
  title={Deep reinforcement learning for coordination in traffic light control},
  author={van der Pol, Elise},
  journal={Master's thesis, University of Amsterdam},
  year={2016}
}

@article{mnih2015human,
  title={Human-level control through deep reinforcement learning},
  author={Mnih, Volodymyr and Kavukcuoglu, Koray and Silver, David and Rusu, Andrei A and Veness, Joel and Bellemare, Marc G and Graves, Alex and Riedmiller, Martin and Fidjeland, Andreas K and Ostrovski, Georg and others},
  journal={Nature},
  volume={518},
  number={7540},
  pages={529},
  year={2015},
  publisher={Nature Publishing Group}
}

@incollection{mannion2016experimental,
  title={An experimental review of reinforcement learning algorithms for adaptive traffic signal control},
  author={Mannion, Patrick and Duggan, Jim and Howley, Enda},
  booktitle={Autonomic Road Transport Support Systems},
  pages={47--66},
  year={2016},
  publisher={Springer}
}

@article{el2014design,
  title={Design of reinforcement learning parameters for seamless application of adaptive traffic signal control},
  author={El-Tantawy, Samah and Abdulhai, Baher and Abdelgawad, Hossam},
  journal={Journal of Intelligent Transportation Systems},
  volume={18},
  number={3},
  pages={227--245},
  year={2014},
  publisher={Taylor \& Francis}
}

@article{chin2011q,
  title={Q-learning based traffic optimization in management of signal timing plan},
  author={Chin, Yit Kwong and Bolong, Nurmin and Kiring, Aroland and Yang, Soo Siang and Teo, Kenneth Tze Kin},
  journal={International Journal of Simulation, Systems, Science \& Technology},
  volume={12},
  number={3},
  pages={29--35},
  year={2011}
}

@article{el2013multiagent,
  title={Multiagent reinforcement learning for integrated network of adaptive traffic signal controllers (MARLIN-ATSC): methodology and large-scale application on downtown Toronto},
  author={El-Tantawy, Samah and Abdulhai, Baher and Abdelgawad, Hossam},
  journal={IEEE Transactions on Intelligent Transportation Systems},
  volume={14},
  number={3},
  pages={1140--1150},
  year={2013},
  publisher={IEEE}
}

@article{balaji2010urban,
  title={Urban traffic signal control using reinforcement learning agents},
  author={Balaji, PG and German, X and Srinivasan, Dipti},
  journal={IET Intelligent Transport Systems},
  volume={4},
  number={3},
  pages={177--188},
  year={2010},
  publisher={IET}
}

@article{arel2010reinforcement,
  title={Reinforcement learning-based multi-agent system for network traffic signal control},
  author={Arel, Itamar and Liu, Cong and Urbanik, Tom and Kohls, Airton G},
  journal={IET Intelligent Transport Systems},
  volume={4},
  number={2},
  pages={128--135},
  year={2010},
  publisher={IET}
}

@article{abdoos2013holonic,
  title={Holonic multi-agent system for traffic signals control},
  author={Abdoos, Monireh and Mozayani, Nasser and Bazzan, Ana LC},
  journal={Engineering Applications of Artificial Intelligence},
  volume={26},
  number={5-6},
  pages={1575--1587},
  year={2013},
  publisher={Elsevier}
}

@techreport{thorpe1996tra,
  title={Tra c light control using sarsa with three state representations},
  author={Thorpe, Thomas L and Anderson, Charles W},
  year={1996},
  institution={Citeseer}
}

@inproceedings{wiering2000multi,
  title={Multi-agent reinforcement learning for traffic light control},
  author={Wiering, MA},
  booktitle={Machine Learning: Proceedings of the Seventeenth International Conference (ICML'2000)},
  pages={1151--1158},
  year={2000}
}

@article{brockfeld2001optimizing,
  title={Optimizing traffic lights in a cellular automaton model for city traffic},
  author={Brockfeld, Elmar and Barlovic, Robert and Schadschneider, Andreas and Schreckenberg, Michael},
  journal={Physical Review E},
  volume={64},
  number={5},
  pages={056132},
  year={2001},
  publisher={APS}
}

@article{abdulhai2003reinforcement,
  title={Reinforcement learning for true adaptive traffic signal control},
  author={Abdulhai, Baher and Pringle, Rob and Karakoulas, Grigoris J},
  journal={Journal of Transportation Engineering},
  volume={129},
  number={3},
  pages={278--285},
  year={2003},
  publisher={American Society of Civil Engineers}
}

@article{sutton1988learning,
  title={Learning to predict by the methods of temporal differences},
  author={Sutton, Richard S},
  journal={Machine learning},
  volume={3},
  number={1},
  pages={9--44},
  year={1988},
  publisher={Springer}
}

@article{guo2019model,
  title={A model and genetic algorithm for area-wide intersection signal optimization under user equilibrium traffic},
  author={Guo, Jianhua and Kong, Ye and Li, Zongzhi and Huang, Wei and Cao, Jinde and Wei, Yun},
  journal={Mathematics and Computers in Simulation},
  volume={155},
  pages={92--104},
  year={2019},
  publisher={Elsevier}
}

@article{bergstra2012random,
  title={Random search for hyper-parameter optimization},
  author={Bergstra, James and Bengio, Yoshua},
  journal={Journal of Machine Learning Research},
  volume={13},
  number={Feb},
  pages={281--305},
  year={2012}
}

@Article{Friedman2001,
  author        = {Friedman, Jerome H.},
  title         = {Greedy function approximation: a gradient boosting machine},
  journal       = {Annals of statistics},
  year          = {2001},
  pages         = {1189--1232},
  publisher     = {JSTOR},
}

@article{pedregosa2011scikit,
  title={Scikit-learn: Machine learning in Python},
  author={Pedregosa, Fabian and Varoquaux, Ga{\"e}l and Gramfort, Alexandre and Michel, Vincent and Thirion, Bertrand and Grisel, Olivier and Blondel, Mathieu and Prettenhofer, Peter and Weiss, Ron and Dubourg, Vincent and others},
  journal={Journal of machine learning research},
  volume={12},
  number={Oct},
  pages={2825--2830},
  year={2011}
}

@inproceedings{Chen:2016:XST:2939672.2939785,
 author = {Chen, Tianqi and Guestrin, Carlos},
 title = {XGBoost: A Scalable Tree Boosting System},
 booktitle = {Proceedings of the 22Nd ACM SIGKDD International Conference on Knowledge Discovery and Data Mining},
 series = {KDD '16},
 year = {2016},
 isbn = {978-1-4503-4232-2},
 location = {San Francisco, California, USA},
 pages = {785--794},
 numpages = {10},
 url = {http://doi.acm.org/10.1145/2939672.2939785},
 doi = {10.1145/2939672.2939785},
 acmid = {2939785},
 publisher = {ACM},
 address = {New York, NY, USA},
}

@Article{Freund1999,
  author        = {Freund, Yoav and Schapire, Robert and Abe, Naoki},
  title         = {A short introduction to boosting},
  journal       = {Journal-Japanese Society For Artificial Intelligence},
  year          = {1999},
  volume        = {14},
  number        = {771-780},
  pages         = {1612},
  publisher     = {JAPANESE SOC ARTIFICIAL INTELL},
}

@TechReport{Breiman1997,
  author        = {Breiman, Leo},
  title         = {Arcing the edge},
  year          = {1997},
  publisher     = {Technical Report 486, Statistics Department, University of California at …},
}

@Article{Friedman2002,
  author        = {Friedman, Jerome H.},
  title         = {Stochastic gradient boosting},
  journal       = {Computational statistics \& data analysis},
  year          = {2002},
  volume        = {38},
  number        = {4},
  pages         = {367--378},
  publisher     = {Elsevier},
}

@InProceedings{Schapire1999,
  author        = {Schapire, Robert E.},
  title         = {A brief introduction to boosting},
  booktitle     = {Ijcai},
  year          = {1999},
  volume        = {99},
  pages         = {1401--1406},
}

@Article{Aimsun2012,
  author        = {Aimsun, TSS},
  title         = {Dynamic simulators users manual},
  journal       = {Transport Simulation Systems},
  year          = {2012},
  volume        = {20},
  __markedentry = {[Tuo Mao:6]},
}

@Article{Anbaroglu2014,
  author        = {Anbaroglu, Berk and Heydecker, Benjamin and Cheng, Tao},
  title         = {Spatio-temporal clustering for non-recurrent traffic congestion detection on urban road networks},
  journal       = {Transportation Research Part C: Emerging Technologies},
  year          = {2014},
  volume        = {48},
  pages         = {47-65},
  __markedentry = {[Tuo Mao:6]},
  doi           = {10.1016/j.trc.2014.08.002},
  isbn          = {0968090X},
}

@Article{Ceylan2004,
  author        = {Ceylan, Halim and Bell, Michael GH},
  title         = {Traffic signal timing optimisation based on genetic algorithm approach, including drivers’ routing},
  journal       = {Transportation Research Part B: Methodological},
  year          = {2004},
  volume        = {38},
  number        = {4},
  pages         = {329-342},
  __markedentry = {[Tuo Mao:6]},
  isbn          = {0191-2615},
}

@Article{Diakaki2003,
  author        = {Diakaki, Christina and Dinopoulou, Vaya and Aboudolas, Kostas and Papageorgiou, Markos and Ben-Shabat, Elia and Seider, Eran and Leibov, Amit},
  title         = {Extensions and new applications of the traffic-responsive urban control strategy: Coordinated signal control for urban networks},
  journal       = {Transportation Research Record: Journal of the Transportation Research Board},
  year          = {2003},
  number        = {1856},
  pages         = {202-211},
  __markedentry = {[Tuo Mao:6]},
  isbn          = {0361-1981},
}

@Article{Foy1992,
  author        = {Foy, Mark D and Benekohal, Rahim F and Goldberg, David E},
  title         = {Signal timing determination using genetic algorithms},
  journal       = {Transportation Research Record},
  year          = {1992},
  number        = {1365},
  pages         = {108},
  __markedentry = {[Tuo Mao:6]},
}

@Article{Goldberg1988,
  author        = {Goldberg, David E and Holland, John H},
  title         = {Genetic algorithms and machine learning},
  journal       = {Machine learning},
  year          = {1988},
  volume        = {3},
  number        = {2},
  pages         = {95-99},
  __markedentry = {[Tuo Mao:6]},
  isbn          = {0885-6125},
}

@Article{Mihaita2018,
  author        = {Mihăiţă, Adriana Simona and Dupont, Laurent and Camargo, Mauricio},
  title         = {Multi-objective traffic signal optimization using 3D mesoscopic simulation and evolutionary algorithms},
  journal       = {Simulation Modelling Practice and Theory},
  year          = {2018},
  volume        = {86},
  pages         = {120-138},
  __markedentry = {[Tuo Mao:6]},
  isbn          = {1569-190X},
}

@Article{Pan2014,
  author        = {Pan, Bei and Demiryurek, Ugur and Gupta, Chetan and Shahabi, Cyrus},
  title         = {Forecasting spatiotemporal impact of traffic incidents for next-generation navigation systems},
  journal       = {Knowledge and Information Systems},
  year          = {2014},
  volume        = {45},
  number        = {1},
  pages         = {75-104},
  __markedentry = {[Tuo Mao:6]},
  doi           = {10.1007/s10115-014-0783-6},
  isbn          = {0219-1377 0219-3116},
}

@Article{Papageorgiou2003,
  author        = {Papageorgiou, Markos and Diakaki, Christina and Dinopoulou, Vaya and Kotsialos, Apostolos and Wang, Yibing},
  title         = {Review of road traffic control strategies},
  journal       = {Proceedings of the IEEE},
  year          = {2003},
  volume        = {91},
  number        = {12},
  pages         = {2043-2067},
  __markedentry = {[Tuo Mao:6]},
  isbn          = {0018-9219},
}

@Article{Ritchie1990,
  author        = {Ritchie, Stephen G},
  title         = {A knowledge-based decision support architecture for advanced traffic management},
  journal       = {Transportation Research Part A: General},
  year          = {1990},
  volume        = {24},
  number        = {1},
  pages         = {27-37},
  __markedentry = {[Tuo Mao:6]},
  isbn          = {0191-2607},
}

@Article{Skabardonis2003,
  author        = {Skabardonis, Alexander and Varaiya, Pravin and Petty, Karl},
  title         = {Measuring recurrent and nonrecurrent traffic congestion},
  journal       = {Transportation Research Record: Journal of the Transportation Research Board},
  year          = {2003},
  number        = {1856},
  pages         = {118-124},
  __markedentry = {[Tuo Mao:6]},
  isbn          = {0361-1981},
}

@Article{Smith1993,
  author        = {Smith, MJ and Van Vuren, T},
  title         = {Traffic equilibrium with responsive traffic control},
  journal       = {Transportation science},
  year          = {1993},
  volume        = {27},
  number        = {2},
  pages         = {118-132},
  __markedentry = {[Tuo Mao:6]},
  isbn          = {0041-1655},
}

@Book{Varaiya2007,
  title         = {Finding and Analyzing True Effect of Non-recurrent Congestion on Mobility and Safety},
  year          = {2007},
  author        = {Varaiya, Pravin Pratap},
  __markedentry = {[Tuo Mao:6]},
}

@Article{Yang1995,
  author        = {Yang, Hai and Yagar, Sam},
  title         = {Traffic assignment and signal control in saturated road networks},
  journal       = {Transportation Research Part A: Policy and Practice},
  year          = {1995},
  volume        = {29},
  number        = {2},
  pages         = {125-139},
  __markedentry = {[Tuo Mao:6]},
  doi           = {https://doi.org/10.1016/0965-8564(94)E0007-V},
  isbn          = {0965-8564},
  url           = {http://www.sciencedirect.com/science/article/pii/0965856494E0007V},
}

@inproceedings{Tuo2019,
  author        = {Mao, T. and Mihaita, A.S. and Cai, C.},
  title         = {Traffic Signal Control Optimisation under Severe Incident Conditions using Genetic Algorithm},
  booktitle     = {Proc. of the 26th ITS World Congress 2019},
  year          = {2019},
  }

@inproceedings{Mihaita2019,
  author        = {Mihaita, A.S. and Liu, Z. and Cai, C. and Rizoiu, M.A },
  title         = {Arterial incident duration prediction using a bi-level framework of extreme gradient-tree boosting},
  booktitle     = {Proc. of the 26th ITS World Congress 2019, 21-25 Oct},
  year          = {2019},
  }
\begin{IEEEbiography}[{\includegraphics[width=1in,height=1.25in,clip,keepaspectratio]{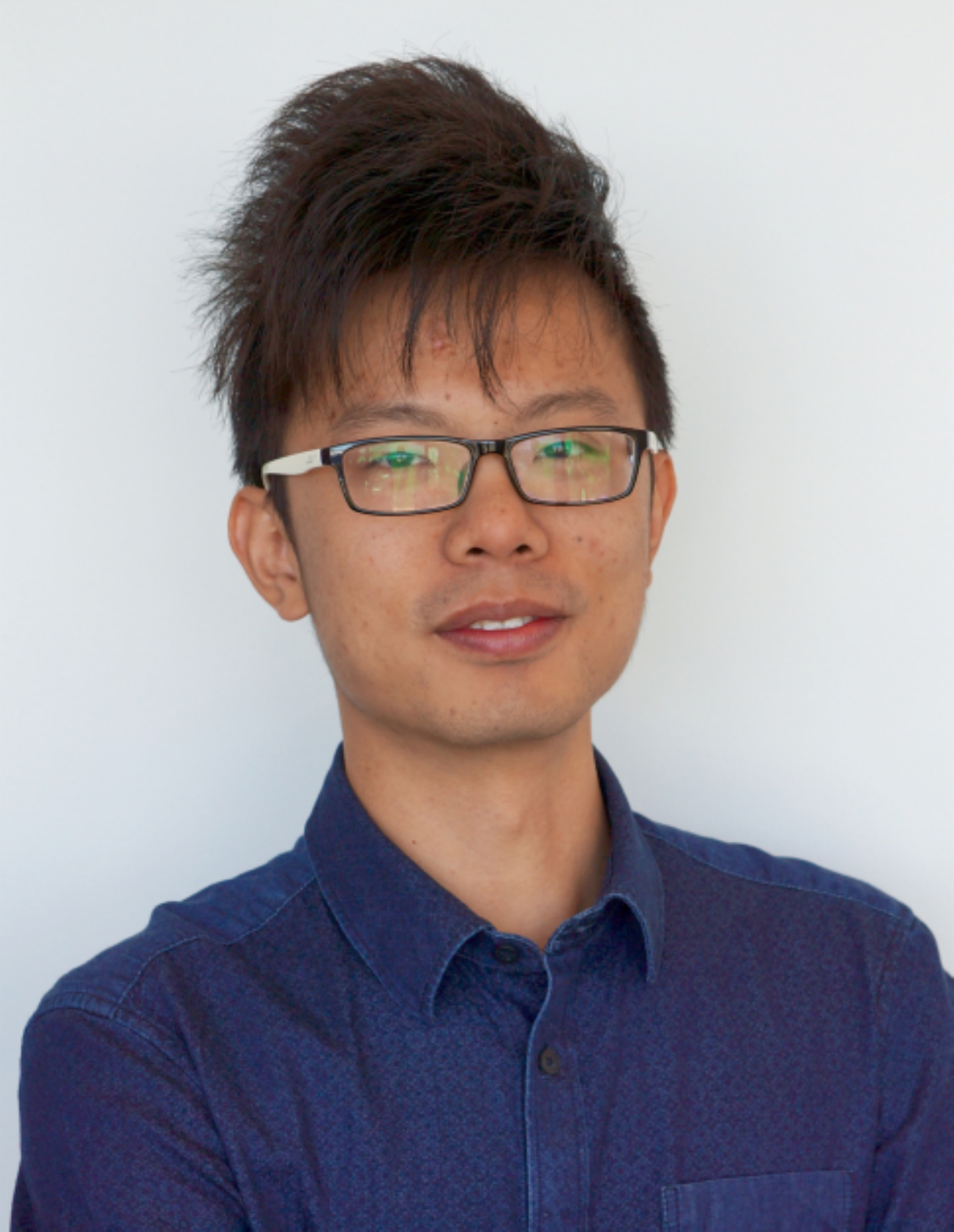}}]{Dr. Tuo Mao}
is a Ph.D. graduated from University of New South Wales (UNSW), a senior engineer in University of Technology, Sydney (UTS). He is also a visiting scientist at the Intelligent Mobility group at Data61 CSIRO. He has experience in motorway modelling and coordinated ramp metering optimization; Vehicle to infrastructure (V2I) communication wireless connection system modelling and simulation; Traffic signal control plan optimization using Genetic algorithm; General machine learning (especially reinforcement learning); Bus signal priority modelling and simulation in a corridor; Public transport assignment modelling and simulation.   
\end{IEEEbiography}
\begin{IEEEbiography}
[{\includegraphics[width=1in,height=1.25in,clip,keepaspectratio]{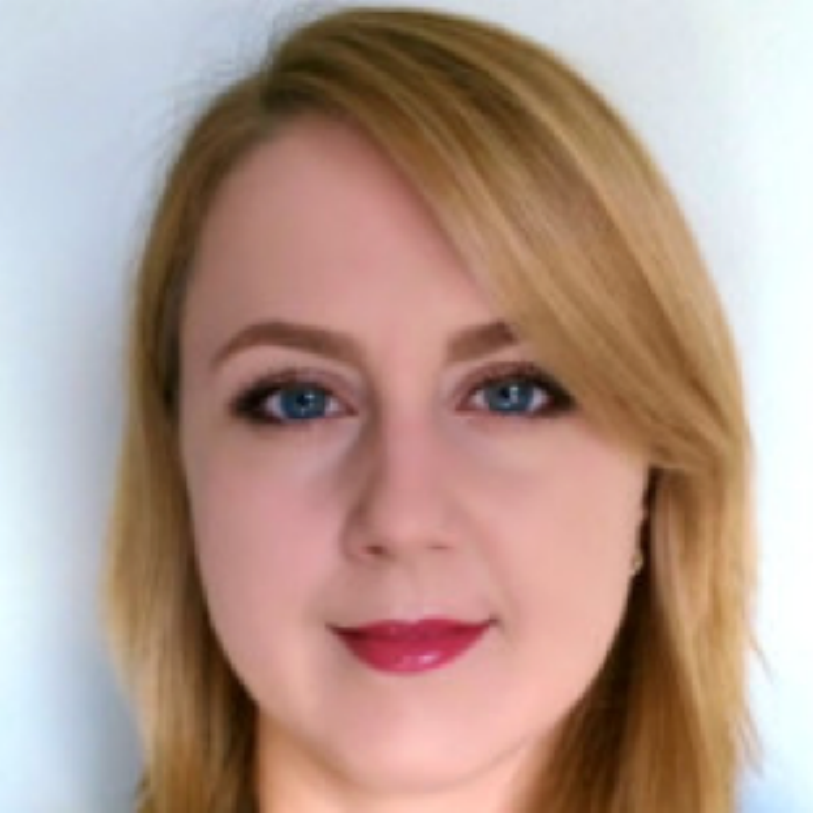}}]{Dr. Adriana Simona Mihaita} is currently a Senior Lecturer in the University of Technology in Sydney, Faculty of Engineering and IT, leading the newly created UTS Future Mobility Research lab. Before joining UTS, she was a Senior Research Scientist and team leader in the ADAIT group from NICTA (now Data61) and continues to act as an affiliated Senior Researcher.\par
Her main research focus is how to engage traffic simulation and optimization using machine learning and artificial intelligence to improve traffic congestion, predicting the duration of traffic accidents and estimating their urban impact, while also leveraging smart analytics for connected and autonomous vehicles in a smart city environment. She is highly engaged in smart city modelling and worked on traffic plan optimization inside ecological neighbourhoods using evolutionary algorithms.\par
Dr. Mihaita holds several leadership roles in various initiatives such as: currently C.I. in the ARC Linkage Project LP180100114 under the Australian-Singapore Strategic Collaboration Partnership (a \$2.4 mil program for collaborations between the two countries on solving congestion problems), and previously: transport leader and scrum master in the ``Premiere’s Innovation Initiative" (a \$3.9 mil program and sole winner of the TfNSW congestion program), the ``On-Demand Mobility” trials in Northern Beaches in partnership with Keolis Downer, as well as ``the Investigation of positioning accuracy of connected vehicles" operated by the Road Safety Centre in Transport for NSW (TfNSW). 
\end{IEEEbiography}

\begin{IEEEbiography}
[{\includegraphics[width=1in,height=1.25in,clip,keepaspectratio]{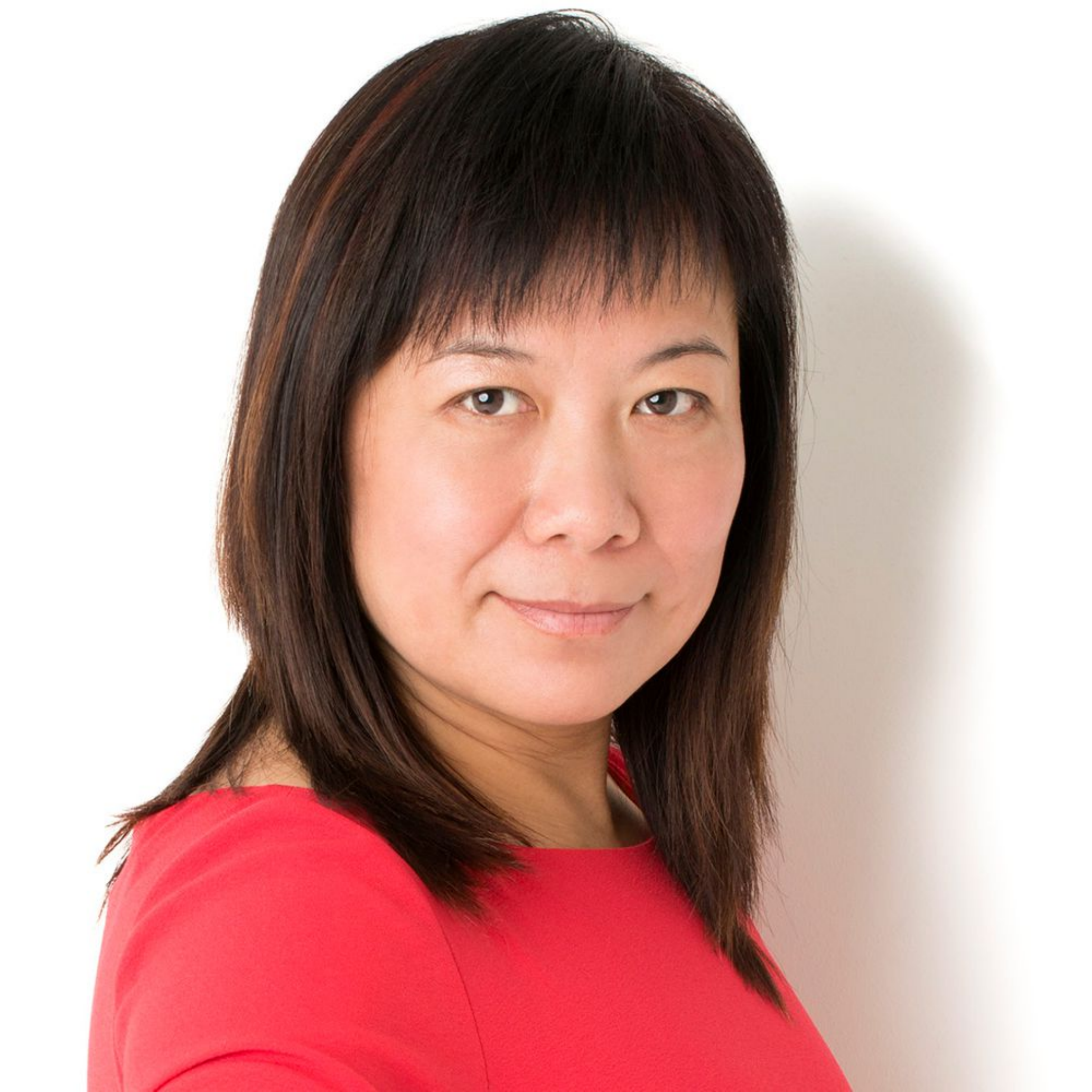}}]{Prof. Fang Chen}
is a prominent leader in AI/data science with international reputation and industrial recognition and the leader of the Data Science Institute at UTS. She is the winner the `Oscars' of Australian science, 2018 Australian Museum Eureka Prize for Excellence in Data Science.

She has created many innovative research and solutions, transforming industries that utilise AI/data science. She has helped industries worldwide advance towards excellence in increasing their productivity, innovation, profitability, and customer satisfaction. The transformations to industry with practical impact won her many industrial recognitions including being named as ``Water Professional of The Year” in 2016.

She has actively led in developing new strategies, which prioritise the organisation’s objectives, and capitalise on any growth opportunities. She has built up a career in creating research and business plans, and executing with leadership and passion.

In science and engineering, Professor Chen has 300+ refereed publications, including several books. She has filed 30+ patents in Australia, US, Canada, Europe, Japan, Korea, Mexico and China.
\end{IEEEbiography}

\begin{IEEEbiography}
[{\includegraphics[width=1in,height=1.25in,clip,keepaspectratio]{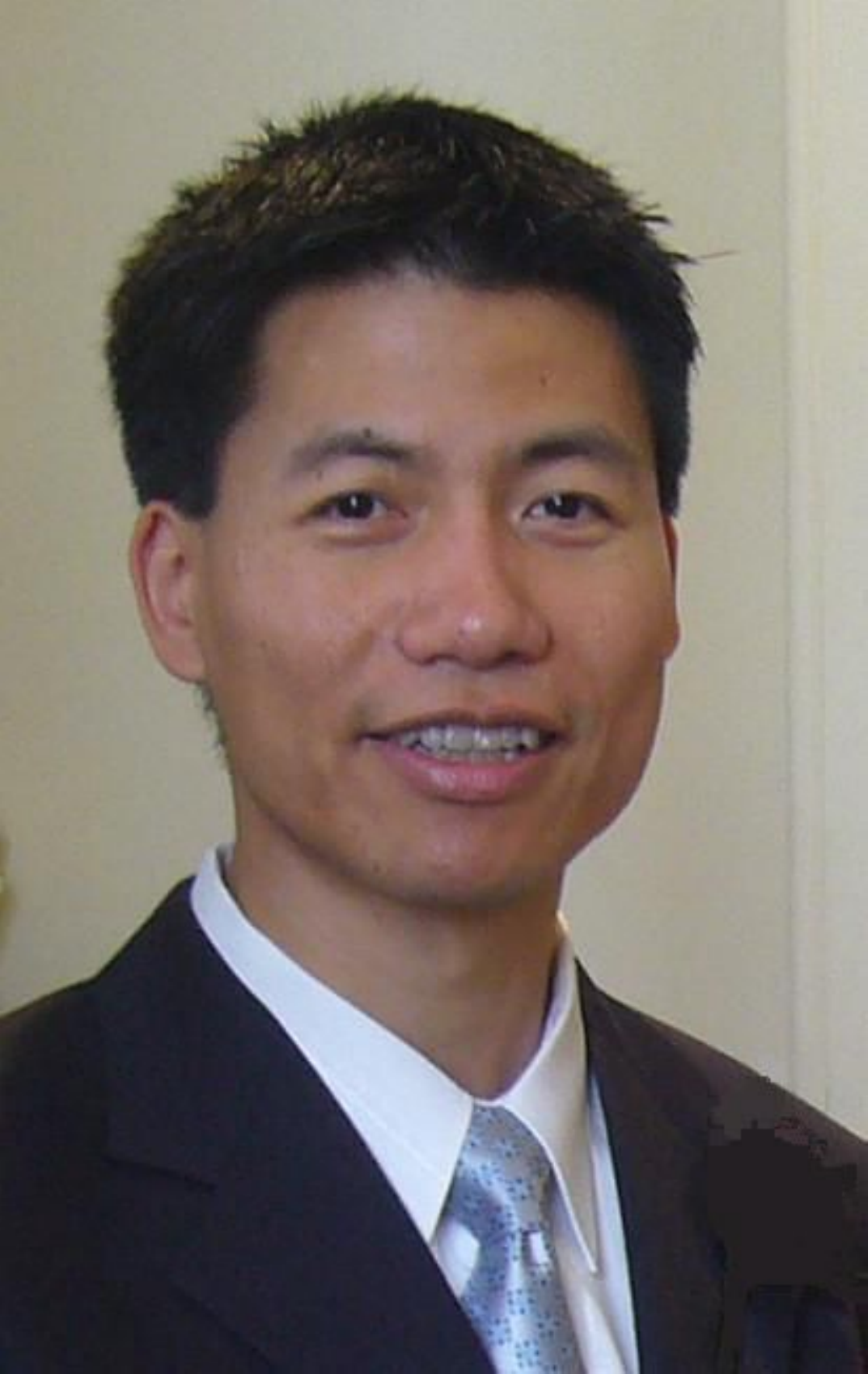}}]{Prof. Hai L. Vu}
 is a Professor of Intelligent Transport System (ITS) at the Monash Institute of Transport Studies in the Faculty of Engineering, Monash University, Australia. He is a recipient of the 2012 Australian Research Council (ARC) Future Fellowship as well as the Victoria Fellowship Award for his research and leadership in ITS. He has recently led a team at Monash in the development and validation of autonomous vehicle that won the 2019 Intelligent Transport Systems (ITS) National Research Award. Prof Vu is a world leading recognized expert with 20 years experience who has authored or coauthored over 180 scientific journals and conference papers in the ITS field. His research interests include modelling, performance analysis and design of complex networks, stochastic optimization and control with applications to connected autonomous vehicles and intelligent transportation.
\end{IEEEbiography} 
\vfill

\end{document}